\documentclass[10pt,twocolumn,letterpaper]{article}

\usepackage[pagenumbers]{cvpr} 

\usepackage{multirow} 

\usepackage[utf8]{inputenc} 
\usepackage[T1]{fontenc}    
\usepackage{url}            
\usepackage{booktabs}       
\usepackage{amsfonts}       
\usepackage{nicefrac}       
\usepackage{amsmath}
\usepackage[table]{xcolor}
\usepackage{microtype}      
\usepackage[dvipsnames]{xcolor}       
\usepackage{graphicx}
\usepackage{multirow}
\usepackage{subcaption}
\usepackage{graphics}
\usepackage{epsfig}
\usepackage{caption}
\usepackage[symbol]{footmisc}
\usepackage{titletoc}
\definecolor{cvprblue}{rgb}{0.21,0.49,0.74}
\usepackage[pagebackref,breaklinks,colorlinks,allcolors=cvprblue]{hyperref}

\newcommand{\symbolHt}{1.1em}

\newcommand{\locChar}{%
\centering
  \begingroup\normalfont
  \raisebox{-0.18\height}
  {\includegraphics[height=\symbolHt]{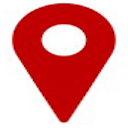}}%
  \endgroup
}

\newcommand{\textChar}{%
\centering
  \begingroup\normalfont
  \raisebox{-0.18\height}
  {\includegraphics[height=\symbolHt]{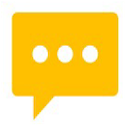}}%
  \endgroup
}

\newcommand{\satChar}{%
  \begingroup\normalfont
  \raisebox{-0.18\height}{\includegraphics[height=\symbolHt]{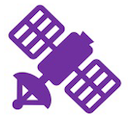}}%
  \endgroup
}
\newcommand{\plusChar}{%
  \begingroup\normalfont
  \raisebox{-0.18\height}
  {\includegraphics[height=\symbolHt]{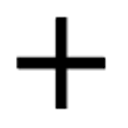}}%
  \endgroup
}
\newcommand{\arrowChar}{%
  \begingroup\normalfont
  \raisebox{-0.18\height}
  {\includegraphics[height=\symbolHt]{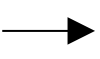}}%
  \endgroup
}
\newcommand{\imageChar}{%
\centering
  \begingroup\normalfont
  \raisebox{-0.18\height}
  {\includegraphics[height=\symbolHt]{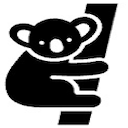}}%
  \endgroup
}
\newcommand{\audioChar}{%
  \begingroup\normalfont
  \raisebox{-0.18\height}
  {\includegraphics[height=\symbolHt]{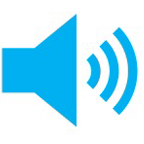}}%
  \endgroup
}

\newcommand{\envChar}{%
  \begingroup\normalfont
  \raisebox{-0.18\height}
  {\includegraphics[height=\symbolHt]{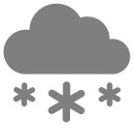}}%
  \endgroup
}

\def\paperID{6735} 
\def\confName{CVPR}
\def\confYear{2026}

\title{ProM3E: \underline{P}robabilistic \underline{M}asked \underline{M}ulti\underline{M}odal \underline{E}mbedding Model for Ecology}

\author{Srikumar Sastry, Subash Khanal, Aayush Dhakal, Jiayu Lin, Dan Cher,\\Phoenix Jarosz, Nathan Jacobs\\
Washington University in St.\ Louis\\
\{{\tt\small s.sastry, k.subash, a.dhakal, jiayu.lin, cher, jarosz, jacobsn}\}{\tt\small @wustl.edu}
}

\begin{document}
\twocolumn[{%
\renewcommand\twocolumn[1][]{#1}%
\maketitle
\begin{center}
    \centering
    \captionsetup{type=figure}
    \includegraphics[width=0.75\textwidth]{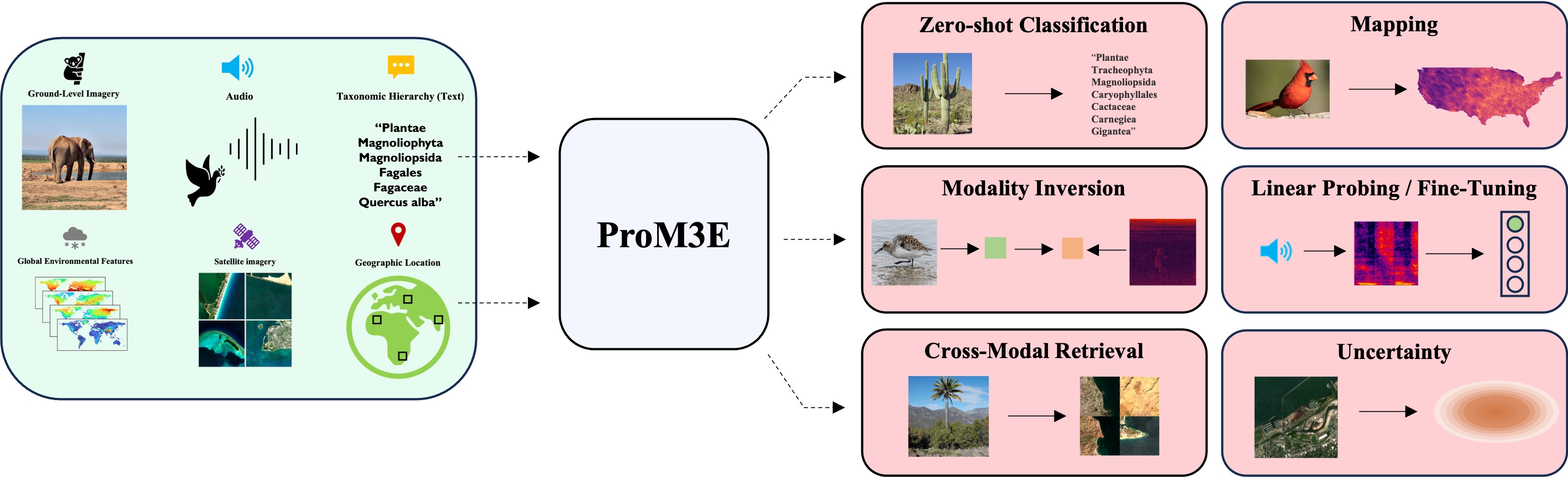}
    
    \captionof{figure}{\textbf{ProM3E Overview.} The versatile capabilities of our model with the ability to accept arbitrary input modalities.}
    \label{img:con}
\end{center}%
}]
\begin{abstract}
We introduce ProM3E, a probabilistic masked multimodal embedding model for any-to-any generation of multimodal representations for ecology. ProM3E is based on masked modality reconstruction in the embedding space, learning to infer missing modalities given a few context modalities. By design, our model supports modality inversion in the embedding space. The probabilistic nature of our model allows us to analyse the feasibility of fusing various modalities for given downstream tasks, essentially learning what to fuse. Using these features of our model, we propose a novel cross-modal retrieval approach that mixes inter-modal and intra-modal similarities to achieve superior performance across all retrieval tasks. We further leverage the hidden representation from our model to perform linear probing tasks and demonstrate the superior representation learning capability of our model. All our code, datasets and model will be released at \url{https://vishu26.github.io/prom3e}.
\end{abstract}    
\section{Introduction}
\label{sec:intro}

In the past decade, the field of multimodal\footnote[1]{With a slight abuse of notation, we use the term ``multimodal learning'' to refer to frameworks learning from more than two modalities (unless specified). This excludes bimodal models such as CLIP.} learning has undergone significant advancements, enabling the development of generalist frameworks~\cite{srivastava2024omnivec2,girdhar2023imagebind} capable of solving a wide range of tasks. This progress led to the development of domain-specific multimodal models such as in remote sensing~\cite{mall2023remote,xiong2024neural,khanal2024psm}, ecology~\cite{sastry2025taxabind,daroya2024wildsat} or medicine~\cite{sun2025causal,lawry2023multi}. Although such models are versatile, they suffer from either of these two limitations: 1) assume some/all modalities are available during inference; and 2) cannot infer missing modalities. These limitations lead to the development of "\textbf{Any-to-Any}" models~\cite{bachmann20244m,mizrahi20234m,wu2024next,shukorunival,luunified} that can generate any modality given a few modalities in the input. Recently, in remote sensing, such any-to-any models~\cite{astruc2024anysat, tseng2025galileo} have been created. These models are trained using a student-teacher framework similar to the joint-embedding predictive architecture (JEPA)~\cite{assran2023self}.

However, such any-to-any models are trained using massive amounts of "\texttt{paired}" data which are difficult to acquire with growing number of modalities. To address this,~\citet{mizrahi20234m} used powerful off-the-shelf models to synthesize paired data given RGB images. Yet, in domains like remote sensing and medicine, certain modalities, such as hyperspectral or MRI imagery, are challenging to acquire or even synthesize. This challenge becomes more difficult when dealing with multimodal data without one-to-one correspondence. For instance, in remote sensing, a single satellite image can correspond to multiple ground-level images.

In this paper, we propose ProM3E, a probabilistic masked multimodal embedding model to address the previously discussed challenges with any-to-any models. We design ProM3E as a two-stage framework with a focus on optimizing and scaling paired multimodal data required for training any-to-any models. This is done by aligning all representations before fusing them. First, we obtain modality-specific encoders using imagebind-style training on massive-scale image-paired datasets. 
Then, we train a lightweight multimodal embedding-based masked variational autoencoder (MVAE) by incorporating embeddings obtained from freezing the encoders. Since, the modalities are already aligned, we require only small-scale paired data of all modalities for training the MVAE. After our model is trained, we analyze various aspects of our model, including the uncertainty captured by our model and the modality gap present in various modalities. As our model supports modality inversion, we propose a novel retrieval strategy combining inter-modal and intra-modal similarities.

The contributions of our work are fourfold:

\begin{enumerate}
    \item \textbf{ProM3E}: We introduce a framework for any-to-any generation of representations. Our model learns a joint probability distribution over input modalities, which is then used to reconstruct the embeddings of unavailable modalities.
    \item \textbf{Modality Inversion}: We introduce a novel cross-modal retrieval strategy that combines the benefits of intra-modal and inter-modal interactions of the modalities using the modality inversion feature of our model.
    \item \textbf{Uncertainty}: We present an extensive qualitative and quantitative analysis of the uncertainty captured by our model to identify the most informative modalities and determine if combining multiple modalities reduces uncertainty in the model.
    \item \textbf{Modality Gap}: We present an analysis of the modality gap present in the modalities before and after training. Furthermore, we analyse whether modality gap is related to the uncertainty captured by our model.
\end{enumerate}
\section{Related Works}
\label{sec:related_works}

\textbf{Multimodal Learning}. Recent multimodal learning approaches aim to align diverse modalities—such as image, audio, text, and tactile signals—into shared embedding spaces using self-supervised contrastive learning and modality-specific encoders \cite{girdhar2023imagebind,wang2023one,wang2023connecting,lei2024vit,zhang2024extending,yang2024binding,feng2025anytouch,guzhov2022audioclip,akbari2021vatt}. Some frameworks leverage pre-trained encoders with learnable routers \cite{wang2024omnibind} or share transformers to enable flexible fusion under task-specific supervision \cite{srivastava2024omnivec,srivastava2024omnivec2,li2024all}. More recent approaches \cite{han2024onellm,lyu2024unibind,xu2024penetrative} utilize LLMs to unify modalities via generated textual anchors or multimodal reasoning. In parallel, multimodal frameworks for remote sensing \cite{mall2023remote,dhakal2024geobind, dhakal2024sat2cap,tseng2025galileo,wang2025towards,khanal2024psm,khanal2023learning} have emerged to integrate modalities such as satellite imagery, text or audio to learn geospatially and semantically rich representations. More recent works focus on understanding and mitigating modality gap \cite{liang2022mind}—a persistent separation between modalities in multi-modal representation spaces—attributing it to factors such as initialization \cite{liang2022mind}, contrastive loss dynamics \cite{liang2022mind,shi2023towards} and modality imbalance \cite{huo2024c2kd,schrodi2024two,mistretta2025cross}.\newline 

\noindent
\textbf{Multimodal Learning for Ecology}. Recent advances in multimodal learning for ecology have been driven by the availability of data through crowdsourced platforms~\cite{van2018inaturalist} and structured benchmarks~\cite{sastry2025taxabind, vendrow2024inquire} which provide data across multiple modalities vital for ecological tasks such as species distribution modeling (SDM) and species fine-grained visual classification (FGVC). New research has moved beyond bimodal frameworks~\cite{stevens2024bioclip, yang2024arboretum,huynh2024contrastive,zermatten2025ecowikirslearningecologicalrepresentation,dollinger2025climplicit} to richer models that integrate additional available modalities~\cite{sastry2024birdsat, daroya2024wildsat, sastry2025taxabind,zbinden2025masksdm}. All such models are provably general-purpose ecological predictors that can be utilized to address a diverse array of tasks in remote sensing and ecology.\newline

\noindent
\textbf{Masking-based Learning}. Masking the input signal and training a model to reconstruct the masked signal has proven to be an effective strategy for pretraining deep-learning models. Several works in the natural language domain mask input word tokens and learn a model to predict those missing tokens~\cite{devlin2019bert,lanalbert,liu2019roberta}. Inspired by the success of this approach, this strategy was extended to the visual domain~\cite{tong2022videomae,he2022masked,Wei_2022_CVPR,Xie_2022_CVPR}. \citet{he2022masked} was the earliest work that used a high masking ratio and train a vision transformer to predict the masked patches. Recent works~\cite{bachmann2022multimae, mizrahi20234m, wang2023image} extended masked modeling to accept multimodal inputs. These works reconstruct missing signals from other modalities using signal from one modality, creating powerful multimodal foundational models.\newline

\noindent
\textbf{Probabilistic Representation Learning}. Recent works on probabilistic multimodal representations aim to capture uncertainty and enhance alignment by projecting inputs as distributions rather than point vectors. Variational Autoencoders (VAEs) introduced foundational techniques for learning latent probabilistic spaces through variational inference and the reparameterization trick \cite{kingma2013auto}. In the vision-language domain, methods such as PCME \cite{chun2021probabilistic} and PCME++ \cite{chun2023improved,chun2024probabilistic} represent image-text pairs as Gaussian distributions and learn cross-modal similarities using Monte Carlo sampling \cite{chun2021probabilistic} or closed-form approximations \cite{chun2023improved,chun2024probabilistic}. Other approaches \cite{ji2023map, neculai2022probabilistic} extend probabilistic modeling with distribution-aware objectives for contrastive, matching, or compositional learning~\cite{upadhyay2023probvlm,park2022probabilistic}.
\begin{figure*}[!t]
  \centering
  \includegraphics[width=\linewidth]{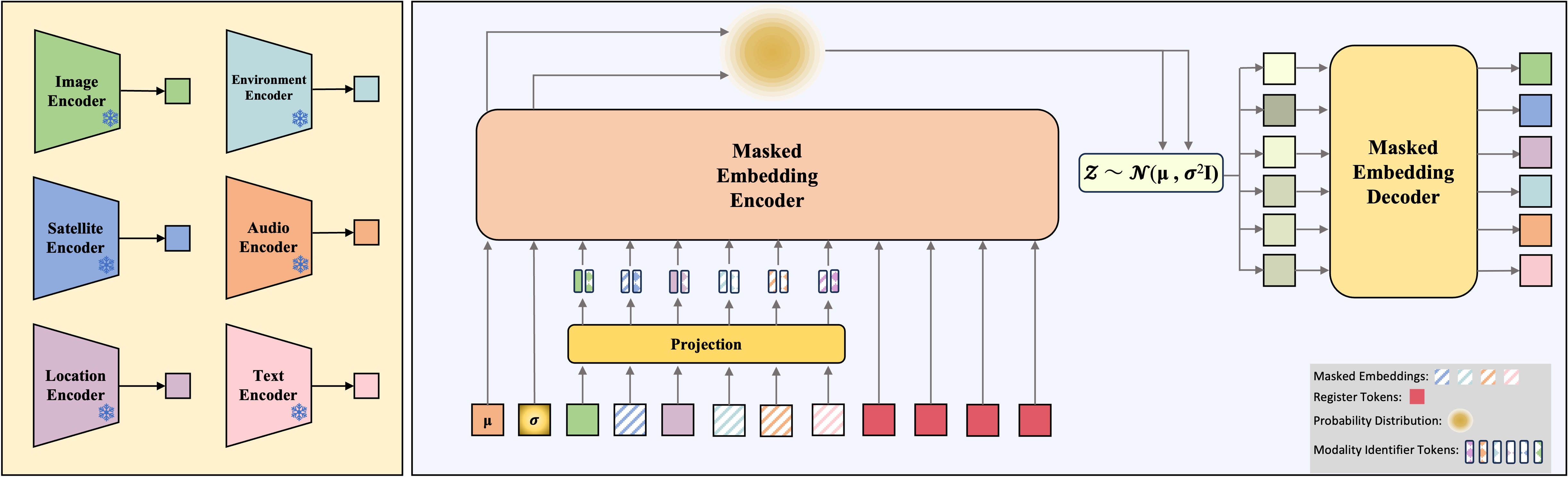}
  \caption{\textbf{ProM3E Framework.} Using embeddings obtained from aligned modality-specific encoders, we model the probability distribution of input modalities using a masked variational autoencoder framework. Subsequently, we utilize the predicted variational distribution of input modalities to reconstruct the embeddings of masked modalities.}
  \label{framework}
\end{figure*}
\section{Method}
Our overall framework is depicted in Figure~\ref{framework}. We design ProM3E as a two stage framework to make it data efficient, scalable and flexible in terms of modalities it can consume. Our method is based on reconstructing global embeddings of modalities since there does not exist one-to-one correspondence between the observations of the modalities. Below we describe our framework with additional details in the subsequent sections.\newline

\noindent
\textbf{Multimodal Alignment}. The first stage of our framework entails aligning all modalities and projecting them into a unified embedding space. This is accomplished using ImageBind or TaxaBind training recipe. After alignment, modality-specific encoders project their respective modalities into the unified embedding space. This simple stage allows training on massive-scale image-paired datasets. It works on global alignment, representing all observations of each modality using a single global embedding. For domains with a one-to-one pixel correspondence, such as in \citet{mizrahi20234m}, patch-wise contrastive methods can be trained to obtain patch-wise embeddings.\newline

\noindent
\textbf{Masked Modality Training}. The second stage involves an SSL-based training for the objective of reconstruction of masked global embeddings. We use a Masked VAE (MVAE) based approach to simultaneously model the joint distribution of modalities and capture uncertainty in the modalities. This stage employs aligned embeddings of various modalities, thus requiring much less paired data.\newline

\noindent
\textbf{Modalities}. In this work, we consider six modalities: ground-level images of species, satellite images, geographic location, species audio, taxonomic text, and environmental covariates. Modalities pertaining to species observations are naturally occurring together and are easily accessible through open citizen science platforms such as iNaturalist~\cite{van2018inaturalist}. Modalities such as satellite imagery and environmental covariates are available through various remote sensing platforms.

\subsection{Multimodal Alignment}
Let $\mathcal{M} = \{m_1, m_2 \ldots, m_6\}$ be the set of modalities in consideration. Consider modality-specific encoders: $\mathcal{H} = \{h_1, h_2 \ldots, h_6\}$. We employ these modality-specific encoders to transform each modality into a single global normalized embedding: $f_i^j=h_i(m_i^j)$, where $m_i^j$ is the $j^{\text{th}}$ observation for the modality $m_i$. For ground-level images, satellite images, audio, and taxonomic texts, we utilize transformer-based models. In contrast, for geographic locations, we employ a Random Fourier Feature-based network, while for environmental covariates, we use a feedforward network.  We utilize TaxaBind's~\cite{sastry2025taxabind} training recipe, which includes multimodal patching to align each modality. We use frozen image and text encoders and project all other modalities into the image-text embedding space. We use image-paired datasets to align all other modalities to the ground-level image modality. We use symmetric SupConLoss~\cite{khosla2020supervised} to align each modality with the image modality. Each alignment training is done independently. We then patch each encoder using the multimodal patching technique in TaxaBind. In the end, we have modality-specific encoders that are in a unified embedding space.

\subsection{Masked Modality Training}
We employ global normalized embeddings from modality-specific encoders. We freeze these encoders during training. We use a transformer-based encoder-decoder architecture to train for masked embedding reconstruction. Our encoder learns a joint probability distribution over arbitrary input modalities, parameterized as a Gaussian distribution. We draw sample embeddings from this distribution for each modality and feed them to the decoder to reconstruct the masked embeddings. This probabilistic model handles the many-to-many correspondence problem between modalities. For this stage we require an all paired dataset of the modalities. Below we describe the training process in detail.\newline

\noindent
\textbf{Encoding.} We treat each embedding as a distinct token for our transformer encoder. Modality-specific projectors transform the embeddings into a compatible dimension with the encoder. Each projector has a two-layer feedforward network with GELU activation. We add modality identifier tokens as positional encoding for the encoder. We introduce two tokens, \textit{[$\mu$]} and \textit{[$\sigma$]}, to learn the joint probability distribution’s mean and standard deviation. In practice, \textit{[$\sigma$]} captures the log of variance. We also incorporate four register tokens~\cite{darcetvision}, useful for eliminating noise and memorizing distribution of modalities. All tokens are concatenated and passed to our transformer encoder, which has stacked self-attention blocks similar to MAE~\cite{he2022masked}. We extract \textit{[$\mu$]} and \textit{[$\sigma$]} from the encoder output to determine the encoded Gaussian distribution’s mean and standard deviation. Let $\mathcal{E}$ be our encoder and $\mathcal{F}=\{f_1, f_2 \ldots, f_6\}$ be a mini-batch of embeddings. Then, the encoding function is given as follows:
\begin{equation}
    \mu_\mathcal{G},\: log\:\sigma^2_\mathcal{G} = \mathcal{E}(\mathcal{G})
\end{equation}
where, $\mathcal{G}\subseteq\mathcal{F}$, which represents the set of input modalities. Our encoder learns a joint gaussian distribution represented as $\mathcal{Z}_i \sim p_\mathcal{E}(\mathcal{Z|\mathcal{G}})$.\newline

\noindent
\textbf{Masking.} For pre-training our VAE, we use a masking strategy similar to MultiMAE~\cite{bachmann2022multimae}. During training, we randomly select one or two visible modalities as input to the encoder. As we will show, the model learns to incorporate additional modalities effectively during inference. The remaining masked modalities are dropped and not encoded. The encoder predicts the joint probability distribution of only the unmasked modalities. This approach aligns with real-world scenarios where most modalities are missing.\newline

\noindent
\textbf{Decoding.} Once we obtain the encoded \textit{[$\mu$]} and \textit{[$\sigma$]} tokens, we employ the reparameterization trick~\cite{kingma2013auto} to generate embedding tokens for each modality. Each sampled token is fed to our decoder for reconstruction. Our decoder learns to generate marginals of each modality from the joint distribution learned by the encoder. Our decoder comprises modality-specific decoders, one for each modality. Each modality-specific decoder is a two-layer FeedForward network with GELU activation. Let $\mathcal{D}=\{\mathcal{D}_1, \mathcal{D}_2\ldots, \mathcal{D}_6\}$ be the set of modality-specific decoders. Then the decoding function is given by the following expressions:
\begin{align}
    \mathcal{Z}_i(\mathcal{G}) &= \mu_\mathcal{G} + \sigma_\mathcal{G}.\epsilon_i\\
    \hat{f_i}(\mathcal{G}) &= \mathcal{D}_i(\mathcal{Z}_i(\mathcal{G}))
\end{align}
where $\epsilon_i\sim\mathcal{N}(0, 1)$ denotes noise used for sampling the latent embedding $\mathcal{Z}_i(\mathcal{G})$.\newline

\noindent
\textbf{Objective Function.} Our training objective is similar to the traditional VAE objective with a modification to its reconstruction loss term. We first calculate the reconstruction loss between the predicted and ground-truth embeddings. This is simply the Euclidean distance between the embeddings. Let $\hat{f_i^j}(\mathcal{G})$ and $f_i^j$ represent the $j^{\text{th}}$ predicted and the ground-truth embedding. Then their distance is calculated as $d_i^\mathcal{G}(j,j) = ||\hat{f_i^j}(\mathcal{G})-f_i^j||_2$. We then use this distance to compute an InfoNCE-based contrastive loss. By employing a contrastive-based loss, we ensure that the model effectively learns the intra-modal distribution of embeddings without merely predicting its centroid. After obtaining the distances, we scale and shift them. This operation acts similarly to the temperature parameter in the InfoNCE loss. The contrastive objective is given as follows:
\begin{equation}
\label{infonce}
    \mathcal{L}_{\text{recon}}(m_i) = \frac{1}{N}\sum_{j=1}^{N}\frac{e^{[\alpha.d_i^\mathcal{G}(j,j)+\beta]}}{\sum_{p=1}^{N} e^{[\alpha.d_i^\mathcal{G}(j,p)+\beta]}}
\end{equation}
where, $\alpha$ and $\beta$ are scaling and shifting parameters respectively. $N$ is the size of the mini-batch used in training. We use the variational information bottleneck (VIB) loss~\cite{chun2023improved,chun2024probabilistic} to regularize our training and prevent the $\sigma$ term from collapsing to zero. The loss is formulated as the KL-divergence between the variational distribution predicted by the model and the Gaussian normal distribution. This loss is given by the closed form as follows:
\begin{equation}
\label{vib}
\mathcal{L}_{VIB} = -\frac{1}{2}(1+log\:\sigma_\mathcal{G}^2-\mu_\mathcal{G}^2-\sigma_\mathcal{G}^2)    
\end{equation}
The final loss is a combination of Equations~\ref{infonce} and~\ref{vib}:
\begin{align}
    \mathcal{L}(m_i) &= \mathcal{L}_{\text{recon}}(m_i) + \lambda \mathcal{L}_{VIB}\\
    \mathcal{L} &= \frac{1}{|\mathcal{M}|}\sum_{i=1}^{|\mathcal{M}|}\mathcal{L}(m_i) + \lambda \mathcal{L}_{VIB}
\end{align}
\subsection{Training Datasets}
\label{train_datasets}
We primarily rely on species observation data from iNaturalist~\cite{van2018inaturalist} for training. For multimodal alignment, we use the same training datasets as TaxaBind, including iSatNat and iSoundNat. In the second stage, we compile an all-paired dataset called \textbf{MultiNat}. We download all species observations from iNaturalist with ground-level images and sound, then filter out observations in the TaxaBench-8k dataset, since this dataset will be used for evaluation. For each observation, we retrieve 256x256 Sentinel-2 imagery from the sentinel-2-cloudless platform and extract bioclimatic variables from the WorldClim-2 dataset. Our dataset contains 79,317 samples. For more details, refer to the appendix.

\begin{table*}[!t]
  \centering
  \begin{center}
  \resizebox{\linewidth}{!}{
  \begin{tabular}{lccccccc}
    \toprule
    Model & Modality &Birds525& CUB-200-2011& BioCLIP-Rare& Inquire (iNat-2024) &iNat-2021&TaxaBench-8k\\
    \midrule
    BioCLIP~\cite{stevens2024bioclip}&\imageChar&82.92&77.51&34.52&62.36&68.24&32.88\\
   ArborCLIP~\cite{yang2024arboretum}&\imageChar&65.84&82.41&27.58&53.02&68.00&31.34\\
   \midrule
   TaxaBind~\cite{sastry2025taxabind}&\imageChar&83.74&78.22&35.84&64.66&70.09&34.45\\
   ProM3E&\imageChar&\textbf{86.89}&\textbf{82.85}&\textbf{37.49}&\textbf{66.96}&\textbf{75.83}&\textbf{39.45}\\
    \midrule
    \multirow{4}{*}{ImageBind~\cite{girdhar2023imagebind}} &\imageChar\plusChar\locChar&-&-&-&-&71.02&36.40\\
    &\imageChar\plusChar\satChar&-&-&-&-&72.62&36.30\\
    &\imageChar\plusChar\envChar&-&-&-&-&71.96&36.59\\
    &\imageChar\plusChar\audioChar&-&-&-&-&-&35.91\\
    \midrule
    \multirow{4}{*}{TaxaBind~\cite{sastry2025taxabind}}&\imageChar\plusChar\locChar&-&-&-&-&72.73&36.59\\
    &\imageChar\plusChar\satChar&-&-&-&-&73.20&37.54\\
    &\imageChar\plusChar\envChar&-&-&-&-&72.02&36.51\\
    &\imageChar\plusChar\audioChar&-&-&-&-&-&36.27\\
    \midrule
    \multirow{4}{*}
    {ProM3E}&\imageChar\plusChar\locChar&-&-&-&-&\textbf{78.27}&\textbf{47.00}\\
    &\imageChar\plusChar\satChar&-&-&-&-&\textbf{77.63}&\textbf{47.05}\\
    &\imageChar\plusChar\envChar&-&-&-&-&\textbf{78.36}&\textbf{46.42}\\
    &\imageChar\plusChar\audioChar&-&-&-&-&-&\textbf{44.84}\\
    \bottomrule
  \end{tabular}
  }
  \caption{Zero-shot classification performance on various fine-grained species classification datasets using the taxonomic description of species.}
  \label{tab:zero-shot-image}
  \end{center}
\end{table*}

\subsection{Multimodal Embeddings for Downstream Transfer}
Our trained model can handle various downstream tasks. Two crucial tasks are linear probing and cross-modal retrieval, which require careful embedding selection, especially in multimodal settings. We provide detailed information on embedding generation for these tasks.\newline

\noindent
\textbf{Linear Probing.} Probing the learned representation of our model is crucial to assess the effectiveness of our pretraining strategy. Several design choices exist for generating embeddings for linear probing. We find that using the hidden representations of our model outperforms using the reconstructed representations. Furthermore, incorporating all encoded tokens, including the register tokens, yields superior performance. For a detailed comparison of linear probing performance across various design choices, please refer to the appendix.

\noindent
\textbf{Cross-Modal Retrieval.} Cross-modal retrieval requires robust representations. Many design choices exist for query and target modality embeddings. Since our model supports modality inversion, we combine the input query embedding with the reconstructed target embedding. This merges inter-modal and intra-modal interactions for retrieval. Let $m_q$ and $m_t$ denote the query and target modalities, respectively. Let $f_q$ and $f_t$ represent the query and target embeddings, respectively. Suppose $\mathcal{G}=\{f_q\}$ denotes the input embeddings. When $\mathcal{G}$ is input, the model reconstructs the target embeddings as $\hat{f_t}(\mathcal{G})$. Finally, we combine the input query embedding with the reconstructed target embedding to generate the final query embedding as follows:
\begin{equation}
    f_q = (1-\delta).f_q + \delta.\hat{f_t}(\mathcal{G})
\end{equation}
where, $\delta$ is a mixing coefficient found using optimal performance on our validation split of MultiNat dataset. We then use these embeddings to compute the cosine similarity with $f_t$ to perform the retrieval.
\section{Experiments}
We assess the effectiveness of our trained model on retrieval and linear probing tasks spanning all modalities. For ease of learning, we initialize our  modality-specific encoders using pretrained TaxaBind~\cite{sastry2025taxabind} encoders. We then train our MVAE with 27M parameters on the MultiNat dataset on a single NVIDIA H-100 GPU with a batch size of 1024. Our MVAE model takes merely 2.5 GPU hours to train, proving to be time and cost effective. For exact implementation details and hyperparameters used, please refer to the appendix. Below we present results on three distinct tasks. We also analyze the uncertainty captured by our model and the modality gap observed. Please see appendix for additional experimental results.

\begin{table}[!t]
    \centering
\begin{minipage}[t]{0.48\linewidth}\centering
\caption*{}
\resizebox{\linewidth}{!}{
\begin{tabular}{lcccc}
    \toprule
    Method & Modality & R@1 & R@5 & R@10\\
    \midrule
    \textit{Random Baseline}&-&0.01&0.05&0.11\\
    \midrule
     \multirow{4}{*}{ImageBind~\cite{girdhar2023imagebind}}&\satChar{}\arrowChar{}\locChar&8.79&22.72&30.84\\
     &\locChar{}\arrowChar{}\satChar{}&9.32&24.16&32.24\\
     &\satChar{}\arrowChar{}\audioChar{}&1.94&6.68&10.56\\
     &\audioChar{}\arrowChar{}\satChar{}&1.86&5.33&9.05\\
     \midrule
     \multirow{4}{*}{TaxaBind~\cite{sastry2025taxabind}}&\satChar{}\arrowChar{}\locChar&8.43&21.72&30.42\\
    &\locChar{}\arrowChar{}\satChar{}&9.62&24.60&33.42\\
     &\satChar{}\arrowChar{}\audioChar{}&2.05&7.03&11.05\\
     &\audioChar{}\arrowChar{}\satChar{}&2.36&5.96&9.50\\
    \midrule
    \multirow{4}{*}{ProM3E}&\satChar{}\arrowChar{}\locChar&\textbf{17.87}&\textbf{43.16}&\textbf{54.53}\\
    &\locChar{}\arrowChar{}\satChar{}&\textbf{13.18}&\textbf{32.29}&\textbf{42.74}\\
     &\satChar{}\arrowChar{}\audioChar{}&\textbf{3.71}&\textbf{10.56}&\textbf{15.22}\\
     &\audioChar{}\arrowChar{}\satChar{}&2.24&\textbf{7.48}&\textbf{12.02}\\
     \bottomrule
  \end{tabular}}
\end{minipage}
\begin{minipage}[t]{0.48\linewidth}\centering
\caption*{}
\label{tab:The parameters 2 }
\resizebox{\linewidth}{!}{
\begin{tabular}{lcccc}
    \toprule
    Method & Modality & R@1 & R@5 & R@10\\
    \midrule
    \textit{Random Baseline}&-&0.01&0.05&0.11\\
    \midrule
     \multirow{4}{*}{ImageBind~\cite{girdhar2023imagebind}}&\imageChar{}\arrowChar{}\locChar&1.09&4.34&7.32\\
     &\imageChar{}\arrowChar{}\satChar{}&1.27&5.24&8.87\\
     &\imageChar{}\arrowChar{}\audioChar{}&11.92&26.81&35.30\\
     &\imageChar{}\arrowChar{}\envChar{}&0.05&1.12&1.57\\
     \midrule
     \multirow{4}{*}{TaxaBind~\cite{sastry2025taxabind}}&\imageChar{}\arrowChar{}\locChar&1.03&4.30&7.59\\
    &\imageChar{}\arrowChar{}\satChar{}&1.34&5.42&9.26\\
     &\imageChar{}\arrowChar{}\audioChar{}&12.27&27.63&36.13\\
     &\imageChar{}\arrowChar{}\envChar{}&0.06&1.12&1.57\\
    \midrule
    \multirow{4}{*}{ProM3E}&\imageChar{}\arrowChar{}\locChar&\textbf{2.26}&\textbf{8.21}&\textbf{13.72}\\
    &\imageChar{}\arrowChar{}\satChar{}&\textbf{2.51}&\textbf{9.00}&\textbf{14.89}\\
     &\imageChar{}\arrowChar{}\audioChar{}&\textbf{14.25}&\textbf{29.42}&\textbf{37.65}\\
     &\imageChar{}\arrowChar{}\envChar{}&\textbf{0.07}&\textbf{2.01}&\textbf{2.21}\\
     \bottomrule
  \end{tabular}}
\end{minipage}
\caption{\textbf{Cross-Modal Retrieval.} We present cross-modal retrieval performance of our model on the TaxaBench-8k dataset with comparison to ImageBind and TaxaBind.}
\label{cm_ret}
\end{table}
\begin{table}[!t]
  \centering
  \resizebox{\linewidth}{!}{
  \begin{tabular}{lcccc}
    \toprule
    Model & Modality & BirdCLEF'22 (\%)&BirdCLEF'23 (\%)&BirdCLEF'24 (\%)\\
    \midrule
    CLAP~\cite{elizalde2023clap}&\audioChar&42.33&32.85&39.72\\
    MGACLAP~\cite{li2024advancing}&\audioChar&56.05&44.03&47.36\\
    ImageBind~\cite{girdhar2023imagebind}&\audioChar&47.11&37.46&45.04\\
    TaxaBind~\cite{sastry2025taxabind}&\audioChar&52.60&42.19&49.31\\
    \midrule
    ProM3E \textbf{(ours)}&\audioChar&\textbf{58.94}&\textbf{52.30}&\textbf{56.66}\\
    \midrule
    \midrule
    ImageBind&\audioChar\plusChar\locChar&60.22&44.04&51.64\\
    TaxaBind&\audioChar\plusChar\locChar&65.07&46.97&56.24\\
    \midrule
    ProM3E 
 \textbf{(ours)}&\audioChar\plusChar\locChar&\textbf{71.56}&\textbf{59.06}&\textbf{62.43}\\
    \bottomrule
  \end{tabular}
  }
  \caption{Top-1 linear probing results on the task of bird species audio classification.}
  \label{tab:audiocls}
\end{table}

\textbf{Image Classification.} In Table~\ref{tab:zero-shot-image}, we present the zero-shot species image classification performance of our model across various datasets. We use full taxonomic text for classification encompassing labels from the kingdom level upto the species level. We compare our model against BioCLIP~\cite{stevens2024bioclip}, ArborCLIP~\cite{yang2024arboretum}, ImageBind~\cite{girdhar2023imagebind}, and TaxaBind~\cite{sastry2025taxabind}. Notably, we observe significant improvements of up to 5\% in the unimodal setting and 10\% in the multimodal setting compared to TaxaBind. ProM3E emerges as the superior model, outperforming all others in all six datasets.

\textbf{Cross-modal Retrieval.} We evaluate the performance of our models' cross-modal alignment on the task of cross-modal retrieval. We use the all paired TaxaBench-8k dataset and perform retrieval given various input and target modalities as shown in Table~\ref{cm_ret}. Our model outperforms TaxaBind and ImageBind in all settings. This demonstrates our model can better understand the interactions between various modalities than the other models in consideration.

\textbf{Audio Classification.} One challenging ecological task is the fine-grained classification of audio of species. We evaluate our model's linear probing capability to predict bird species given audio recording from three geographically distinct datasets. Table~\ref{tab:audiocls} showcases the superior performance of our model in this task with gains upto $\sim$12\%. Our model outperforms other state-of-the-art audio encoders in both unimodal and multimodal setting.

\subsection{What is captured by $\sigma$?}
\begin{figure*}[!t]
\centering
      \begin{minipage}[t]{0.161\linewidth}
  \centering
  \includegraphics[width=\linewidth]{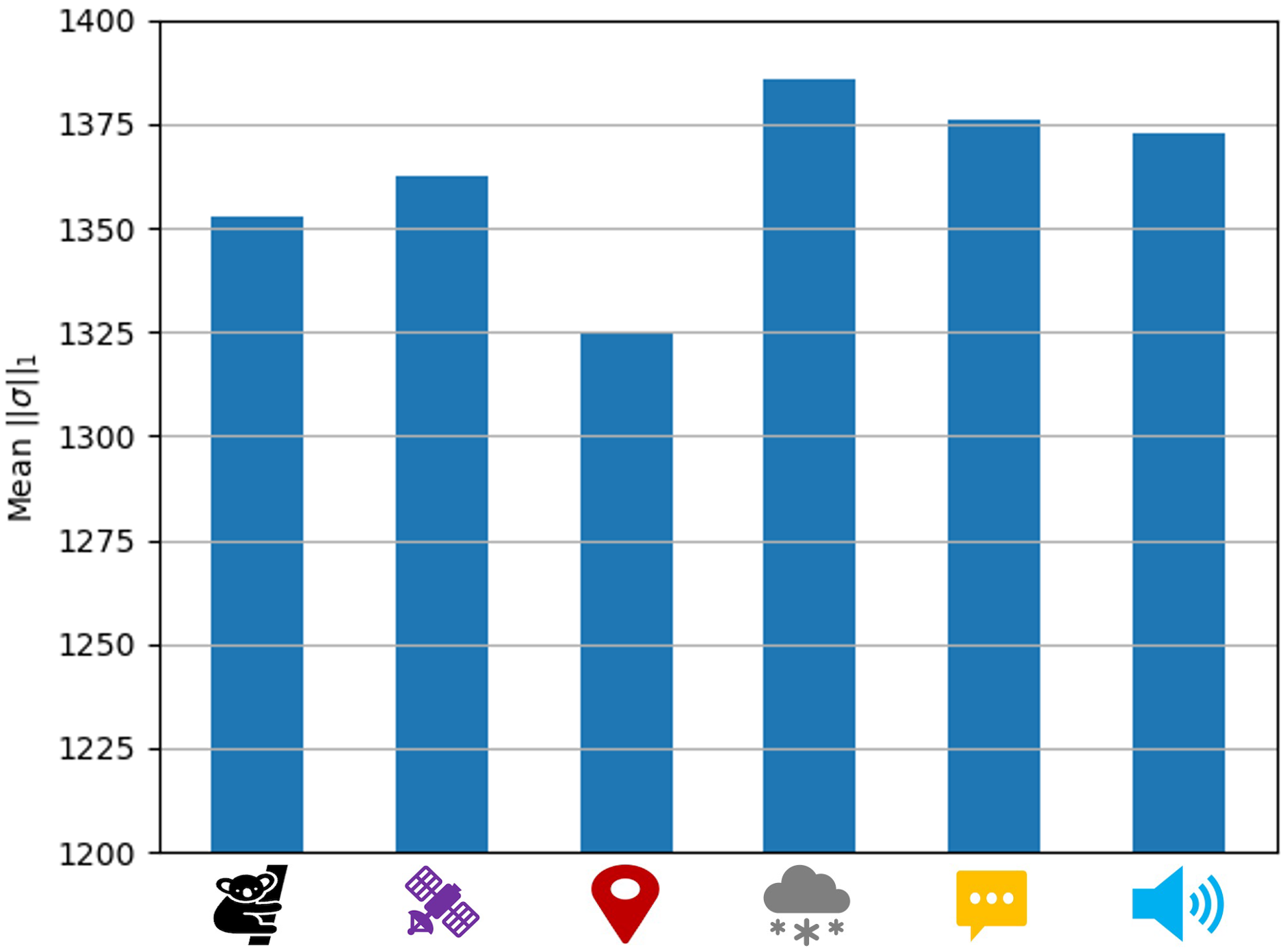}
  \caption*{\scalebox{0.7}{\small(a) Uncertainty in each modality}}
  \end{minipage}
  \begin{minipage}[t]{0.158\linewidth}
  \centering
  \includegraphics[width=\linewidth]{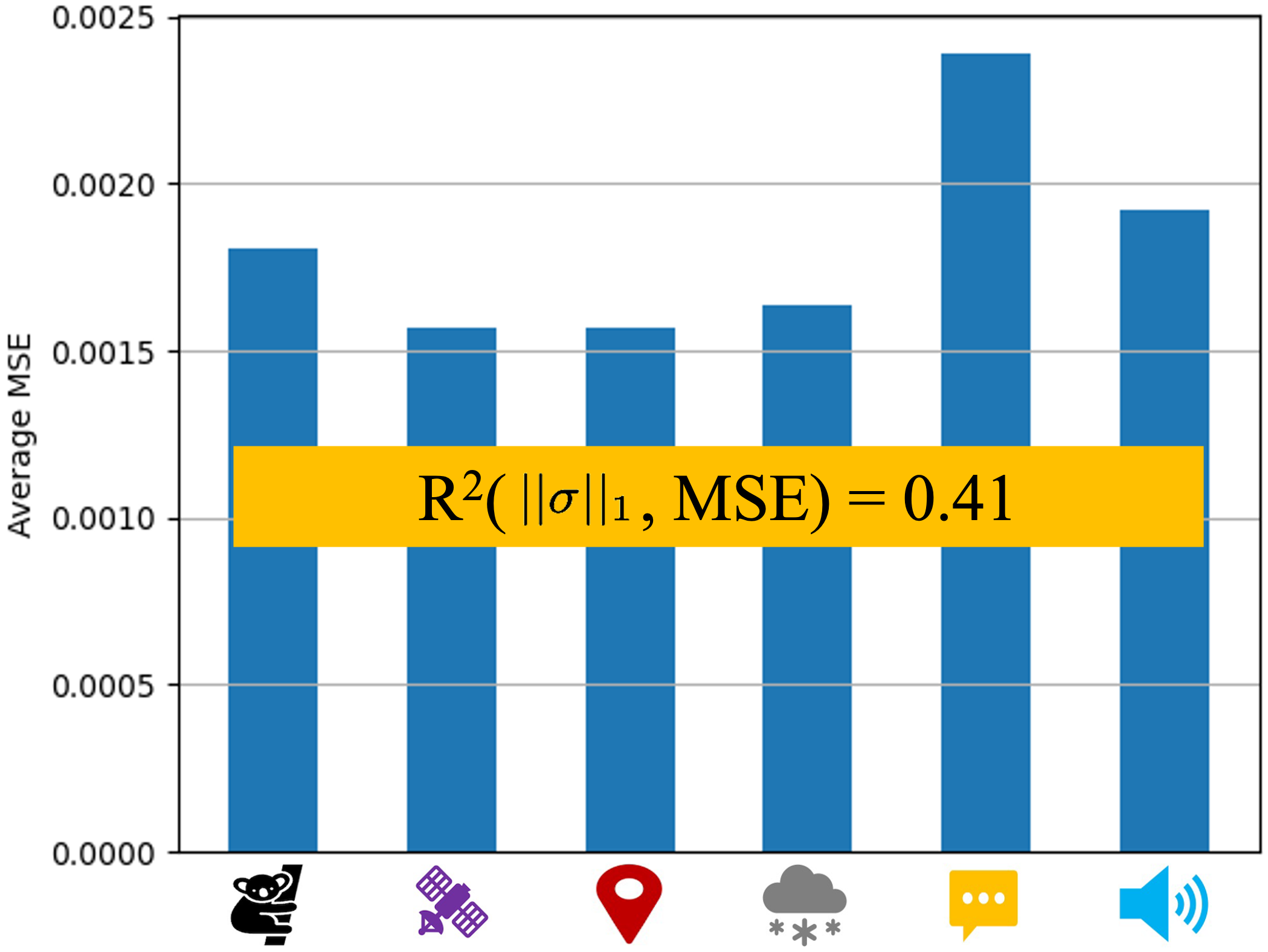}
  \caption*{\scalebox{0.7}{\small(b) MSE Objective}}
  \end{minipage}
  \begin{minipage}[t]{0.1525\linewidth}
  \centering
  \includegraphics[width=\linewidth]{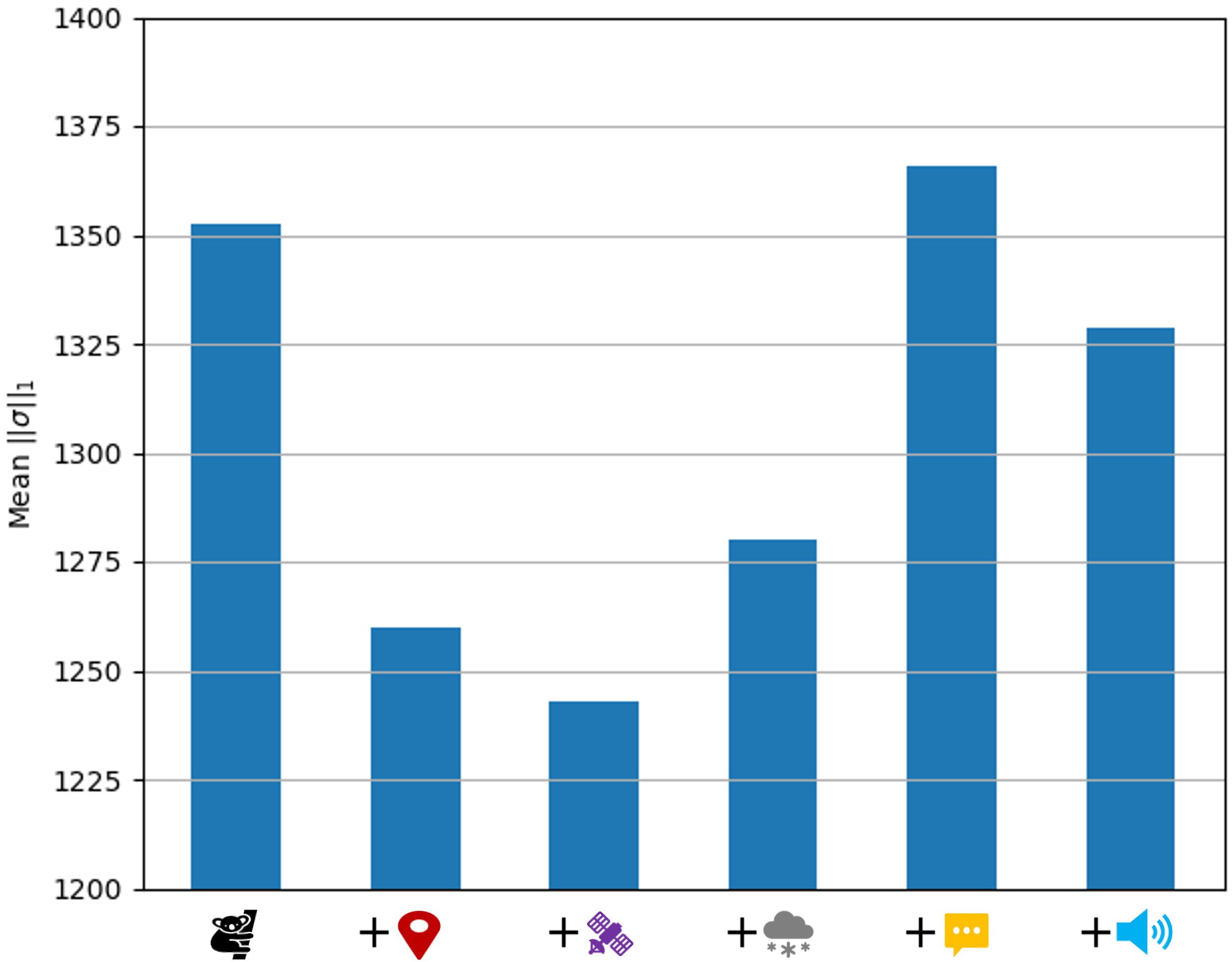}
  \caption*{\scalebox{0.7}{\small(c) Adding single modality}}
  \end{minipage}
  \begin{minipage}[t]{0.158\linewidth}
  \centering
  \includegraphics[width=\linewidth]{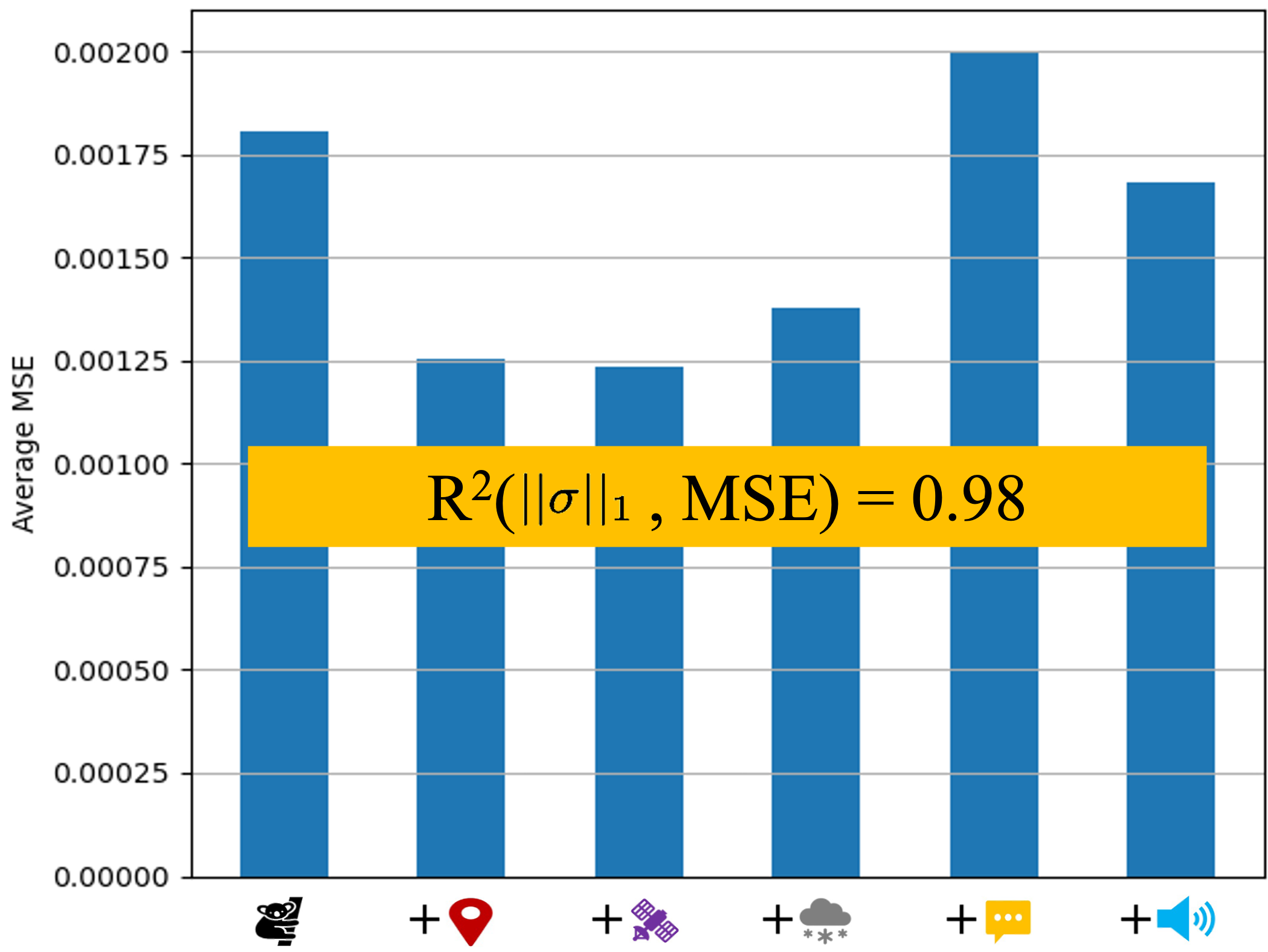}
  \caption*{\scalebox{0.7}{\small(d) MSE Objective}}
  \end{minipage}
  \begin{minipage}[t]{0.1535\linewidth}
  \centering
  \includegraphics[width=\linewidth]{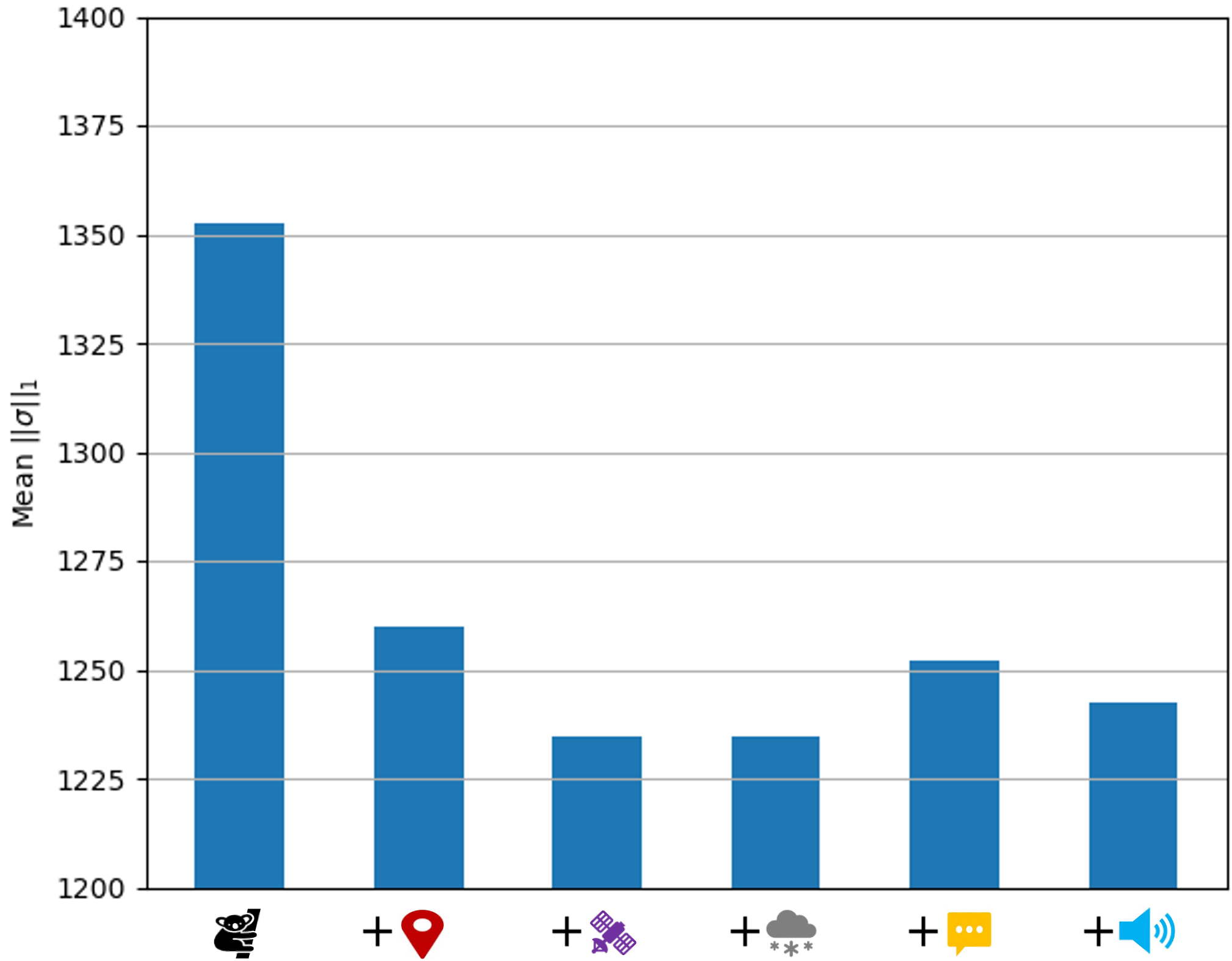}
  \caption*{\scalebox{0.7}{\small(e) Progressively adding modalities}}
  \end{minipage}
  \begin{minipage}[t]{0.152\linewidth}
  \centering
  \includegraphics[width=\linewidth]{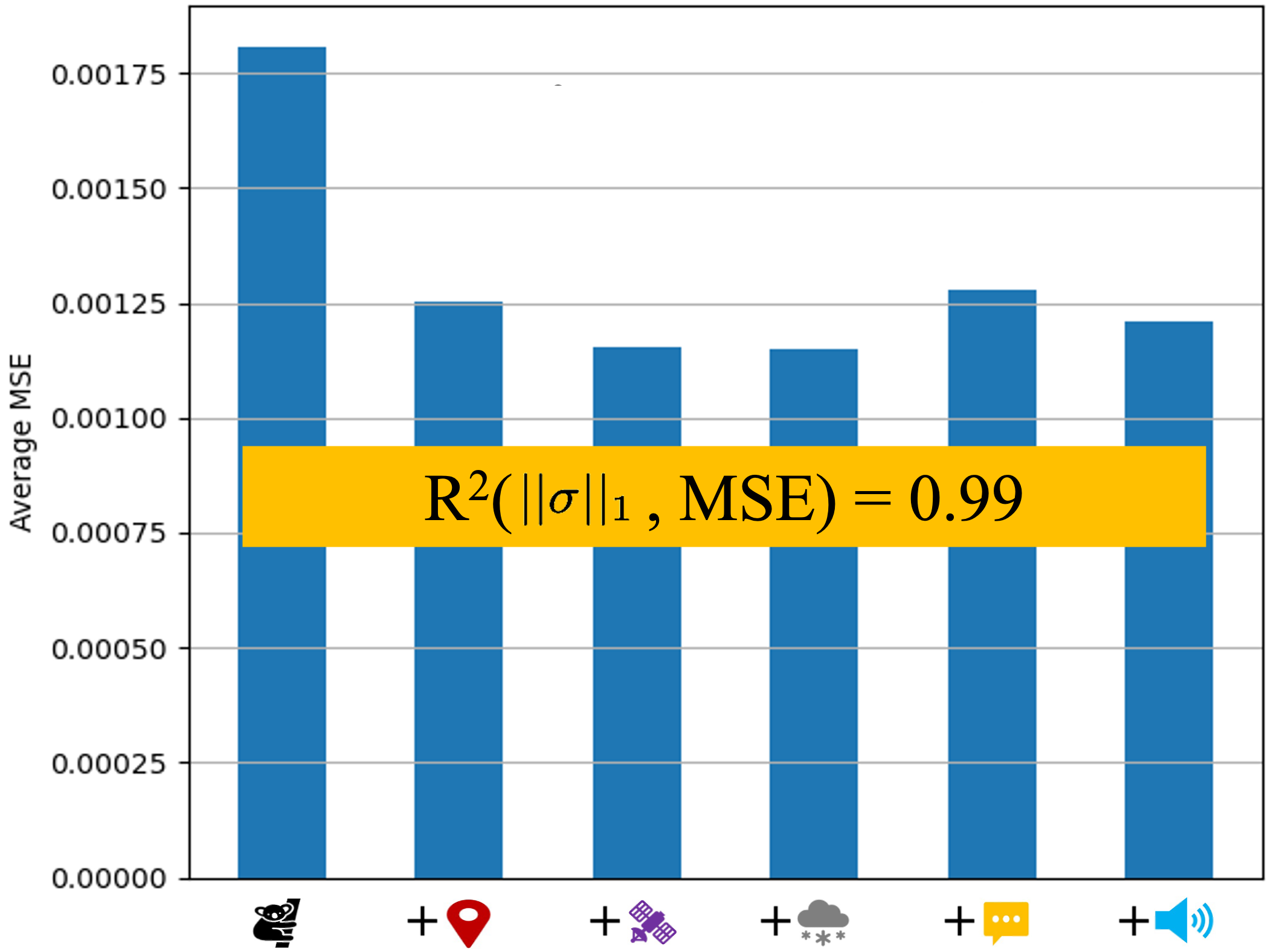}
  \caption*{\scalebox{0.7}{\small(e) MSE Objective}}
  \end{minipage}
\caption{\textbf{Effect on $||\sigma||_1$ values when adding modalities}. We show mean $||\sigma||_1$ predicted and compare it to reconstruction loss of our model. This figure demonstrates a correlation between uncertainty and predictive power of input modalities. In general, adding additional modalities improves reconstruction and the corresponding loss is positively correlated with $||\sigma||_1$.}
\label{fig:sigma_add}
\end{figure*}

We answer this question by analyzing $||\sigma||_1$ values upon adding modalities together and comparing them to reconstruction loss of our model. Essentially, $||\sigma||_1$ depends on the pretraining strategy used. Our pretraining strategy involves predicting masked modalities from a few input modalities. Therefore, $||\sigma||_1$ captures the informativeness of input modalities in predicting the masked modalities. Lower $||\sigma||_1$ values depict high informativeness. We compute mean $||\sigma||_1$ values on the TaxaBench-8k dataset for all the modalities. Figure~\ref{fig:sigma_add}(a) shows the mean $||\sigma||_1$ values of each modality. It depicts that geographic location and ground-level imagery are the most informative. Figure~\ref{fig:sigma_add}(b) shows correlation between MSE objective and $||\sigma||_1$ values in Figure~\ref{fig:sigma_add}(a). In Figure~\ref{fig:sigma_add}(c), we add a single modality at a time to the ground-level image modality. In Figure~\ref{fig:sigma_add}(d) we progressively add modalities from left to right. As we combine multiple modalities, the mean $||\sigma||_1$ values decrease, as shown in Figures~\ref{fig:sigma_add} (d) and (f) (with the exception of text). The correlation between $||\sigma||_1$ and MSE increases as modalities are added together. We find that combining text with other modalities does not add new information. Overall, the figure suggests that the task of inferring masked modalities becomes easier with more input modalities.

\begin{figure}[!t]
    \begin{center}
    \includegraphics[width=\linewidth]{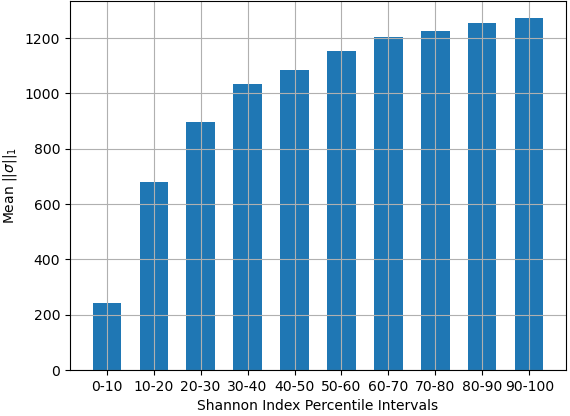}
    \end{center}
  \caption{Mean $||\sigma||_1$ values of geographic locations at various percentile intervals of the Shannon Diversity Index derived from the iNaturalist dataset.
  }\label{fig:biod_interval_shannon}
\end{figure}

\begin{figure*}[!t]
\centering
  \begin{minipage}[t]{0.328\linewidth}
  \centering
  \includegraphics[width=\linewidth]
  {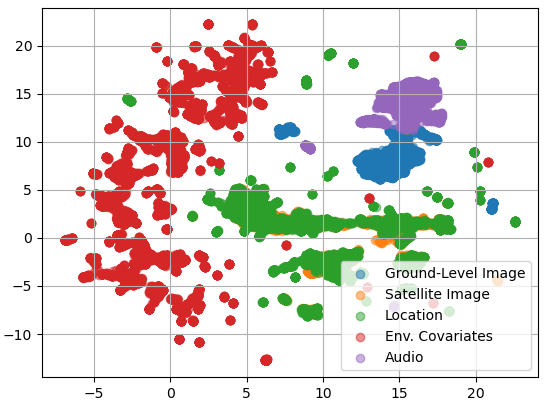}
  \caption*{\small(a) Input TaxaBind Representations}
  \end{minipage}
  \begin{minipage}[t]{0.328\linewidth}
  \centering
  \includegraphics[width=\linewidth]{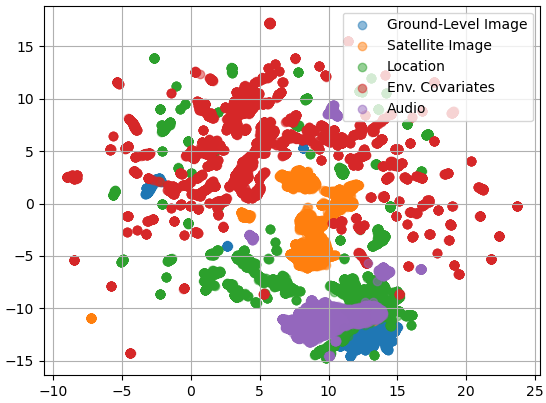}
  \caption*{\small(b) Input to ProM3E Encoder}
  \end{minipage}
  \begin{minipage}[t]{0.322\linewidth}
  \centering
  \includegraphics[width=\linewidth]{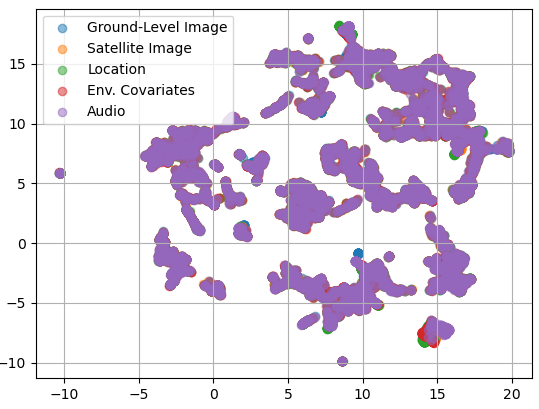}
  \caption*{\small(c) Hidden Representations}
  \end{minipage}
\caption{\textbf{Crossing the Modality Gap}. Our training strategy minimizes the existing modality gap in the hidden representation space of the masked embedding model. This allows our model to predict masked modalities in the input.}
\label{fig:mod_gap_pic}
\end{figure*}

\begin{figure}[!t]
\centering
  \begin{minipage}[t]{0.484\linewidth}
  \centering
  \includegraphics[width=\linewidth]{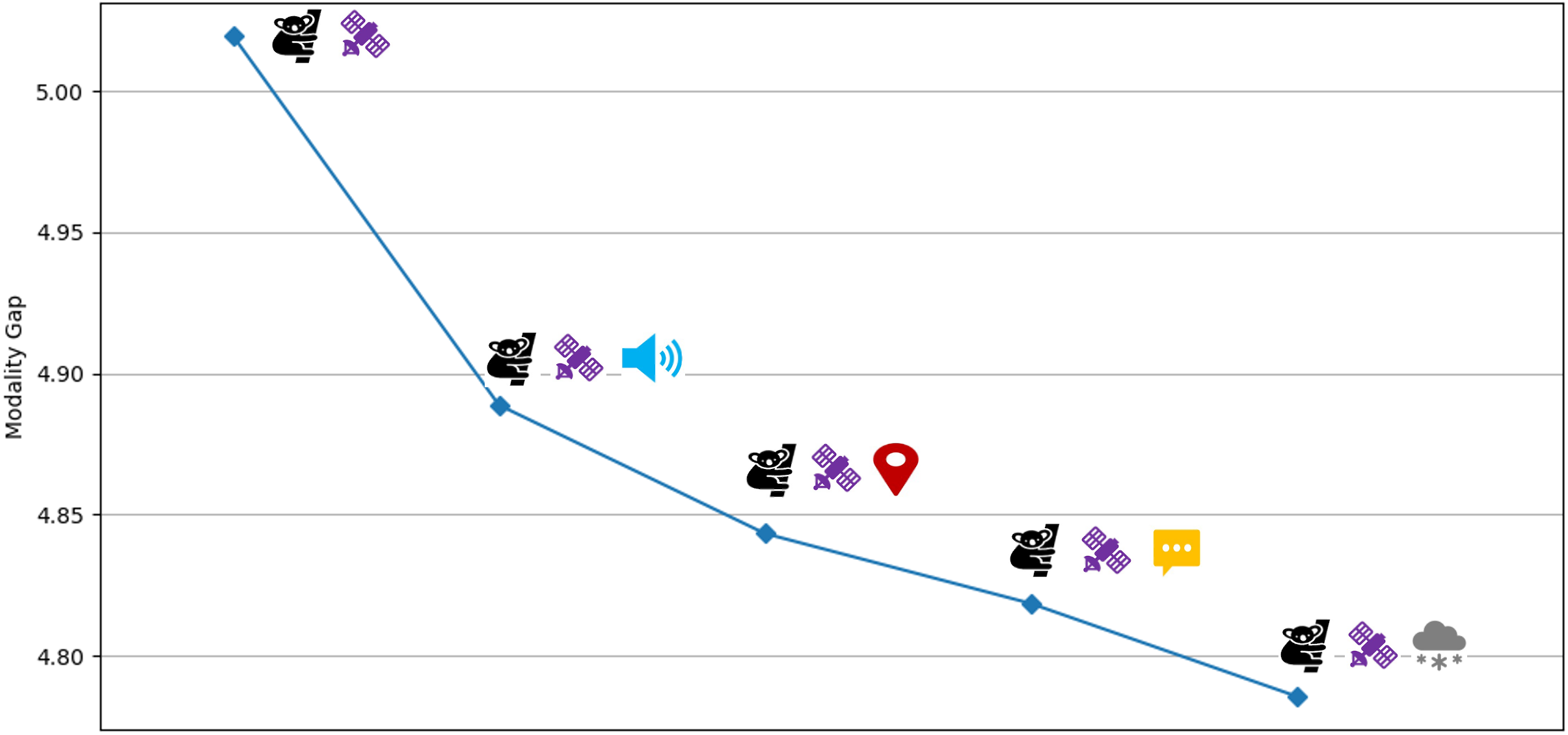}
  \caption*{\small(a) Adding single modality}
  \end{minipage}
  \begin{minipage}[t]{0.496\linewidth}
  \centering
  \includegraphics[width=\linewidth]{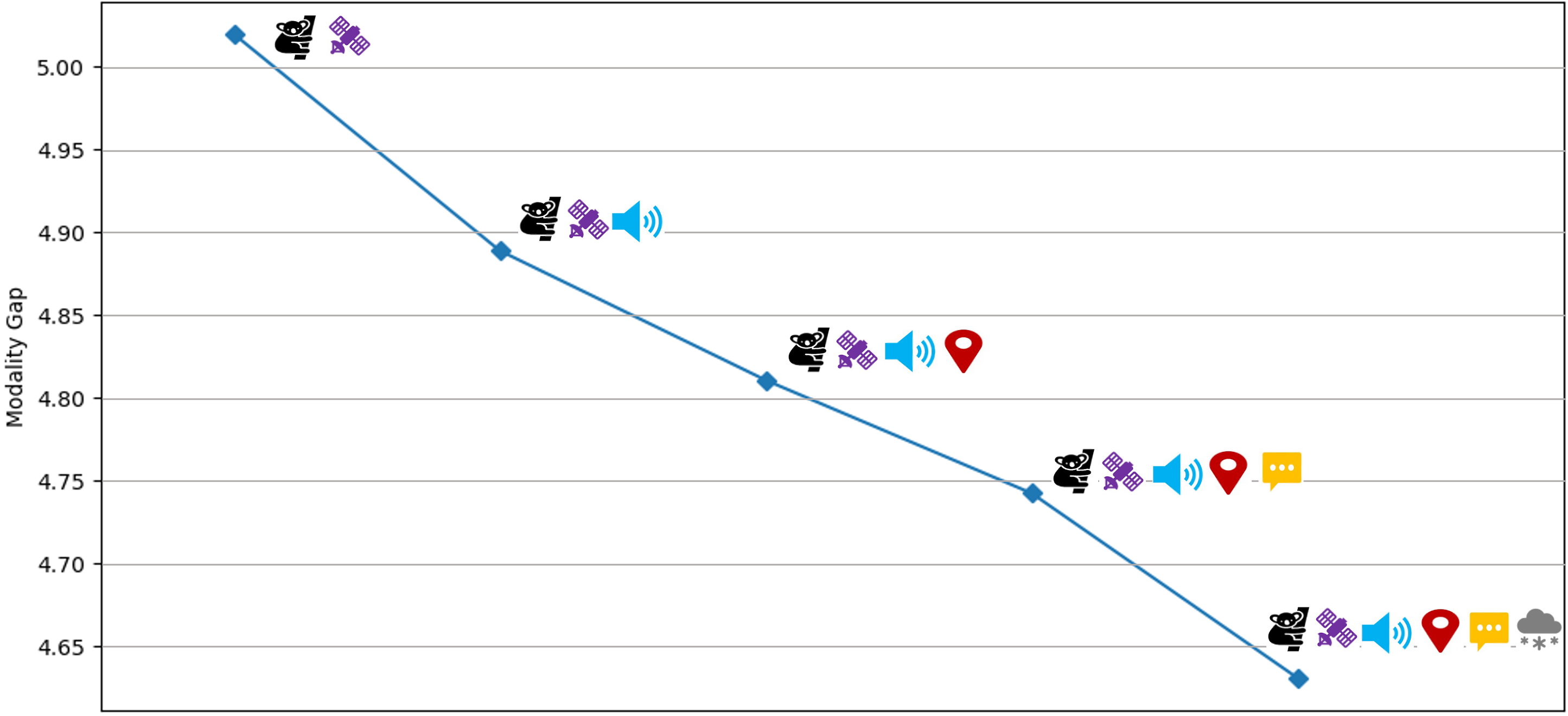}
  \caption*{\small(b) Progressively adding modalities}
  \end{minipage}
\caption{\textbf{Effect on modality gap between two modalities in presence of other modalities}. Quantitative evaluation of the modality gap between ground-level image and satellite image when other modalities are provided as input. It is noticed that the modality gap reduces as more modalities are added.}
\label{fig:quant_modality_gap}
\end{figure}

\textbf{Species Diversity and $||\sigma||_1$.} Intuitively, we anticipate that the $||\sigma||_1$ values of geographic locations reflect the species diversity of those locations. This is because if a specific location or habitat harbors multiple species, the task of predicting the other modalities such as text or ground-level imagery becomes challenging. To this end, we generate a 250x500 species diversity map over the USA using iNaturalist observations and compute Shannon Diversity index at each grid cell. We then compute the $||\sigma||_1$ value at each of those cell and finally calculate the correlation between the biodiversity and $||\sigma||_1$ map. We find a spearman correlation of 0.401 with p-value=0.0, indicating significant positive correlation. Figure~\ref{fig:biod_interval_shannon} depicts a positive correlation between $||\sigma||_1$ and Shannon Diversity index at various percentile intervals. Please see appendix for additional details and visualizations.

\subsection{What happens to the modality gap?}
To the best of our knowledge, we are the first work to analyze modality gap in a model with more than two modalities. In Figure~\ref{fig:quant_modality_gap}, we quantitatively evaluate the effect on modality gap when additional modalities are present. Specifically, we evaluate the modality gap between ground-level and satellite image in the hidden representation space. Following the procedure outlined in \citet{liang2022mind}, we calculate the modality gap by taking the distance between centroid of each modality’s embeddings. We find that the modality gap reduces when additional modalities are introduced. This phenomenon is consistent across all modalities. We investigate the modality gap in the input representation and the hidden representation space of our model in Figure~\ref{fig:mod_gap_pic} using UMAP~\cite{mcinnes2018umap}. We notice that the modality gap reduces as the representations pass through the encoder. For further analysis, please refer to the appendix.
\section{Ablations}
\label{abla}
Here we conduct ablation on dataset size for pretraining ProM3E and various architectural choices. Note that all ablations are done for the MVAE component of our model. The modality-specific encoders from TaxaBind are kept frozen and utilized as is. Additional ablations are in the appendix.

\begin{table*}[!t]
  \centering
  \resizebox{\linewidth}{!}{
  \begin{tabular}{lcccccccc}
    \toprule
    Task &Dataset& Modality & TaxaBind~\cite{sastry2025taxabind}&10\% & 20\% &50\%&75\%&100\%\\
    \midrule
    Image Classification&TaxaBench-8k~&\imageChar\arrowChar\textChar&34.45&36.51&37.31&37.42&38.34&39.45\\
    Image Classification&TaxaBench-8k~&\imageChar\plusChar\satChar\arrowChar\textChar&37.54&41.50&42.31&44.01&44.53&47.05\\
    \midrule
    Retrieval&TaxaBench-8k~&\satChar\arrowChar\locChar&8.43&10.68&10.78&10.82&15.35&17.87\\
    Retrieval&TaxaBench-8k~&\locChar\arrowChar\satChar&9.62&9.62&9.44&9.77&11.70&13.18\\
    \midrule
    Loc. Classification&EcoRegions~&\locChar&73.75&80.26&80.67&80.59&81.13&81.35\\
    Loc. Classification&Biome&\locChar&71.73&79.54&80.71&81.10&81.85&82.30\\
    \midrule
    Audio Classification&BirdCLEF-2023~&\audioChar&42.19&52.06&51.88&53.32&52.48&52.30\\
    Audio Classification&BirdCLEF-2023&\audioChar\plusChar\locChar&46.97&59.96&59.30&60.44&59.96&59.06\\
    \midrule
    \midrule
    Average&&&40.58&46.27&46.55&47.18&48.17&\cellcolor[gray]{0.8}\textbf{49.06}\\
    \bottomrule
  \end{tabular}
  }
  \caption{\textbf{Data Scaling}. We show that our model can be trained with much less all paired dataset and the performance across various dataset sizes and tasks remain consistent. For instance training with 10\% of the dataset (7,913 samples) only results in a performance drop of $\sim$3\% on average.}
  \label{tab:abla_data_scale}
\end{table*}

\begin{table*}[!t]
\begin{minipage}[t]{0.24\linewidth}\centering
\resizebox{0.83\linewidth}{!}{
\begin{tabular}{lcc}
    \toprule
    dim & cls & lin\\
    \midrule
     256&36.28&74.24\\
     512&38.26&78.00\\
     1024&\cellcolor[gray]{0.8}\textbf{39.45}&\cellcolor[gray]{0.8}81.35\\
     2048&39.23&\textbf{82.52}\\
     \bottomrule
  \end{tabular}}
  \caption*{\small(a) Encoder dimension}
\end{minipage}
\begin{minipage}[t]{0.24\linewidth}\centering
\resizebox{0.86\linewidth}{!}{
\begin{tabular}{lcc}
    \toprule
    depth & cls & lin\\
    \midrule
     1&\cellcolor[gray]{0.8}\textbf{39.45}&\cellcolor[gray]{0.8}\textbf{81.35}\\
     2&38.59&80.68\\
     3&38.47&80.73\\
     4&38.36&80.82\\
     \bottomrule
  \end{tabular}}
  \caption*{\small(b) Encoder depth}
\end{minipage}
\begin{minipage}[t]{0.24\linewidth}\centering
\resizebox{\linewidth}{!}{
\begin{tabular}{lcc}
    \toprule
    \small \# Registers & cls & lin\\
    \midrule
     0&39.21&78.51\\
     1&39.20&79.35\\
     2&38.61&80.21\\
     4&\cellcolor[gray]{0.8}\textbf{39.45}&\cellcolor[gray]{0.8}\textbf{81.35}\\
     \bottomrule
  \end{tabular}}
  \caption*{\small(c) \# Register tokens}
\end{minipage}
\begin{minipage}[t]{0.24\linewidth}\centering
\resizebox{\linewidth}{!}{
\begin{tabular}{lcc}
    \toprule
    \small Loss & cls & lin\\
    \midrule
     MSE&33.92&80.27\\
     Contrastive&\cellcolor[gray]{0.8}\textbf{39.45}&\cellcolor[gray]{0.8}\textbf{81.35}\\
     \bottomrule
  \end{tabular}}
  \caption*{\small(d) Loss function}
\end{minipage}
\caption{\textbf{Ablations.} We perform various ablations related to our architecture and loss function. Here, \textit{cls} denotes species image classification on TaxaBench-8k and \textit{lin} denotes linear probing on EcoRegions classification.}
\label{arch_abla}
\end{table*}

\textbf{Scaling ProM3E.} We investigate whether our model can scale well in low data regimes since acquiring multiple modalities is time consuming and expensive. ProM3E is based on the idea of aligning representation before fusing them in order to reduce the amount of paired data required for training. In Table~\ref{tab:abla_data_scale}, we report the performance of our model trained with varying amounts of training data. We evaluate the performance of the models on various tasks and report their average performance. We notice that ProM3E scales very well and performs well in low data regimes. For instance, ProM3E trained with only 7,931 (10\%) samples shows a performance drop of about 3\% on average. Training with the full MultiNat dataset demonstrates the best performance. In all settings, our model outperforms TaxaBind.

\textbf{Architecture and Loss.} We perform several ablations to determine the most optimal architecture and loss function as shown in Table~\ref{arch_abla}. We use species image classification on TaxaBench-8k and linear probing on EcoRegions classification to benchmark each model. From our experiments, we find a single encoder layer and higher embedding dimensions to work the best. We also find that including large number of register tokens improves downstream performance. As suspected, our contrastive objective outperforms vanilla VAE's MSE objective. We find that using MSE loss leads to representation collapse and prevents the model from learning intra-modal distributions.
\section{Conclusions}
In this paper, we introduced ProM3E, a probabilistic masked multimodal embedding model for ecology. Our model learns a joint probability distribution of arbitrary input modalities and reconstructs the embeddings of unavailable modalities. The probabilistic nature of our model allows us to capture the uncertainty of modalities. This is especially useful to quantify the uncertainty of geographic locations which we discovered to be correlated with the species diversity at those locations. The representations generated from our model show excellent performance in downstream tasks beating the state-of-the-art. Our future lines of work will focus on integrating additional modalities such as camera trap imagery to further enhance species understanding and mapping.
\section*{Acknowledgements}
This research used the TGI RAILs advanced compute and data resource which is supported by the National Science Foundation (award OAC-2232860) and the Taylor Geospatial Institute.
{
    \small
    \bibliographystyle{ieeenat_fullname}
    \bibliography{main}
}
\clearpage
\setcounter{page}{1}
\twocolumn[{%
\renewcommand\twocolumn[1][]{#1}%
\maketitlesupplementary
\raggedright
\section{Mapping and Visualization}
\subsection{ICA Visualization of Geo-Embeddings}
\begin{center}
    \centering
    \captionsetup{type=figure}
    \begin{minipage}[t]{0.328\linewidth}
  \centering
  \includegraphics[width=0.72\linewidth]
  {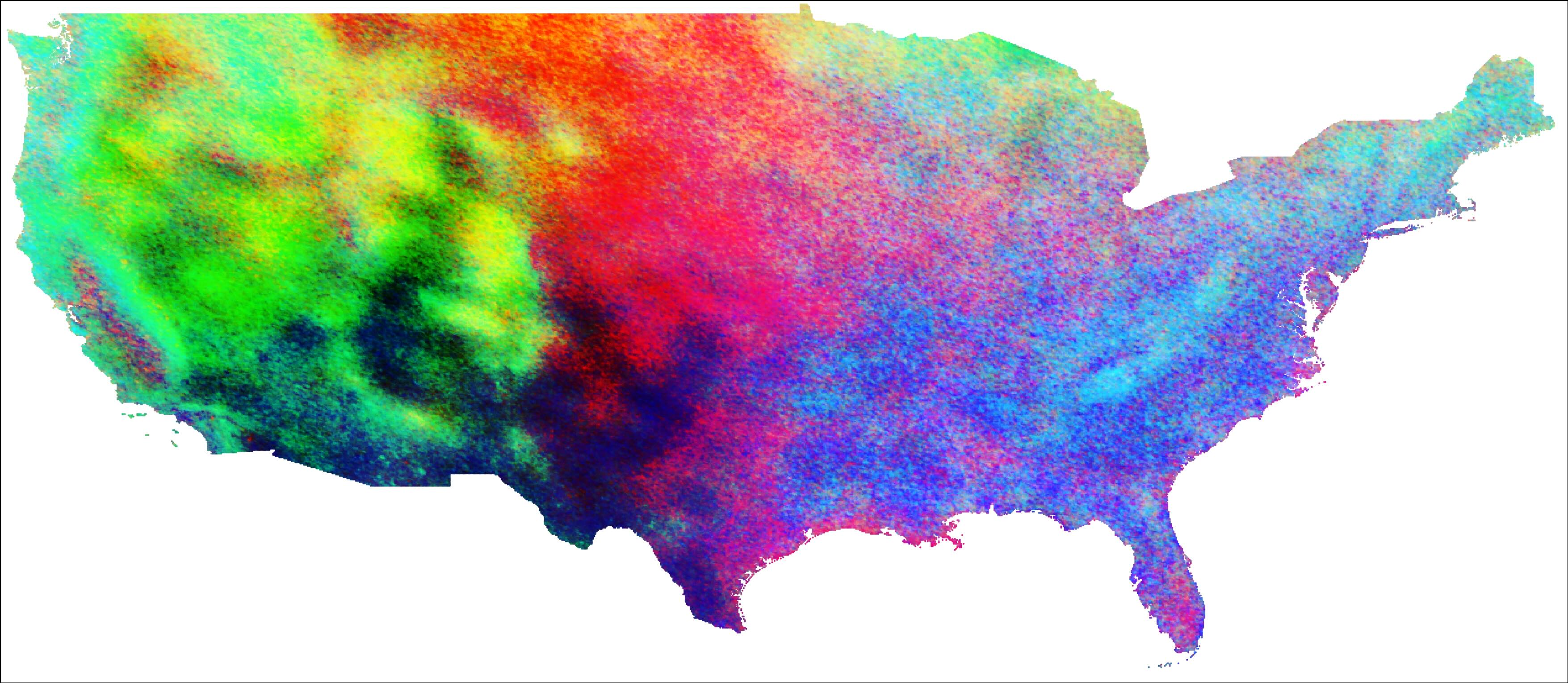}
  \caption*{\small(a) TaxaBind}
  \end{minipage}
  \begin{minipage}[t]{0.328\linewidth}
  \centering
  \includegraphics[width=0.72\linewidth]{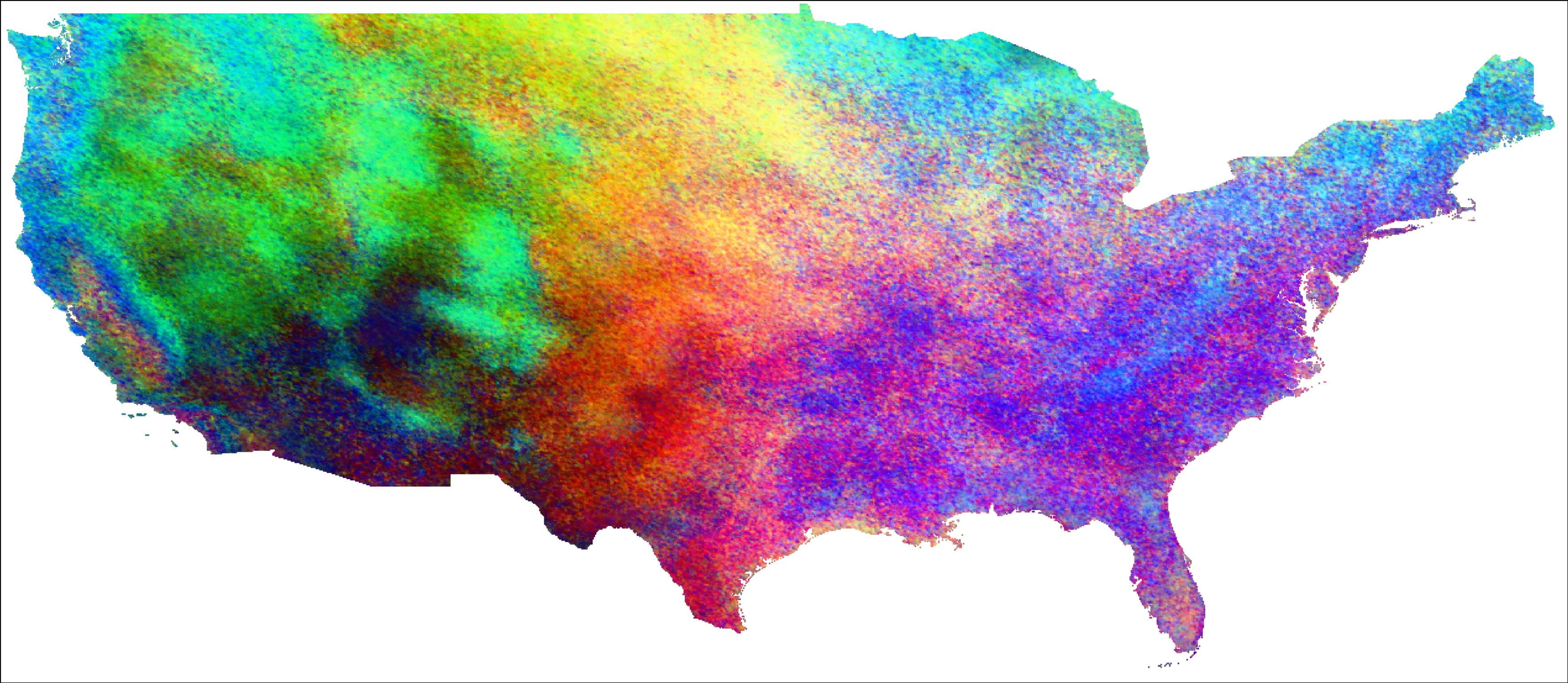}
  \caption*{\small(b) ProM3E (Location token)}
  \end{minipage}
  \begin{minipage}[t]{0.328\linewidth}
  \centering
  \includegraphics[width=0.72\linewidth]{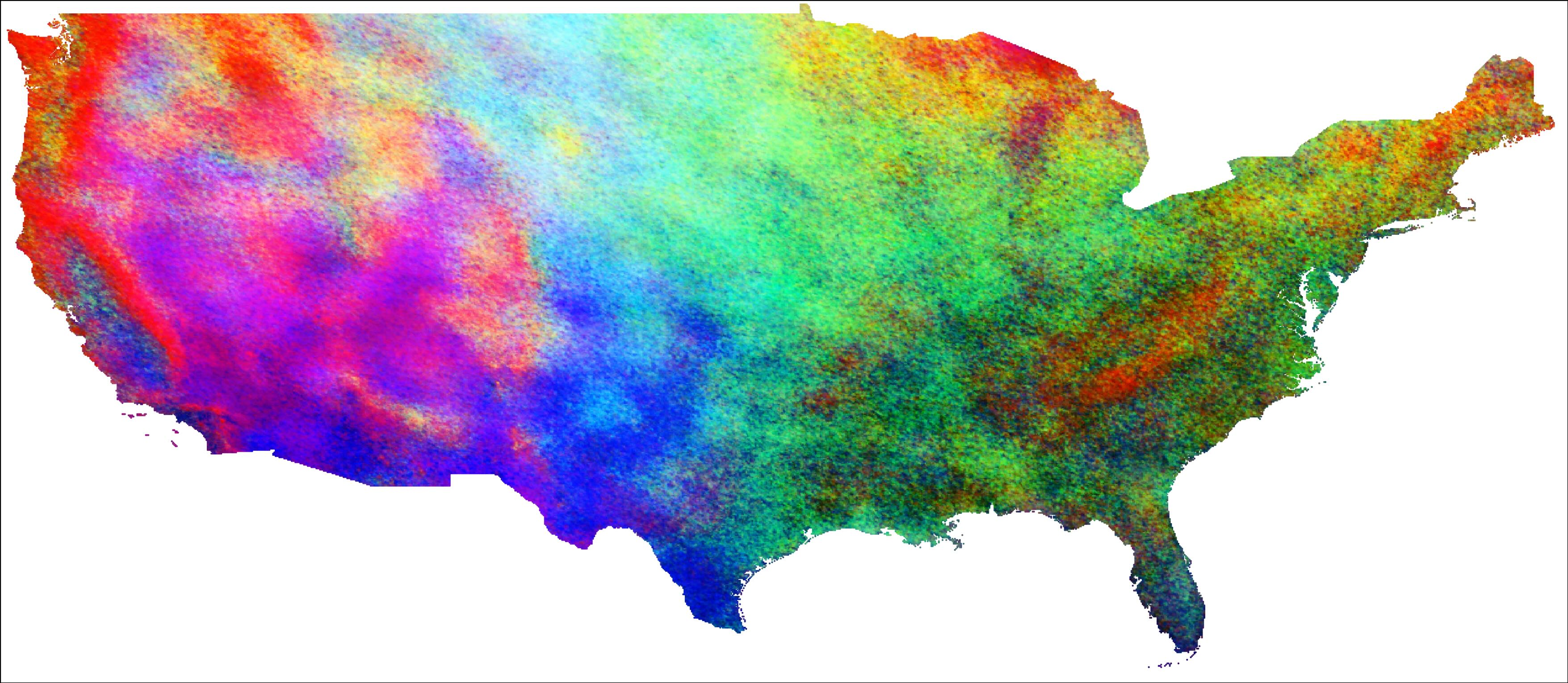}
  \caption*{\small(c) ProM3E (Register token \#1)}
  \end{minipage}
  \begin{minipage}[t]{0.328\linewidth}
  \centering
  \includegraphics[width=0.72\linewidth]
  {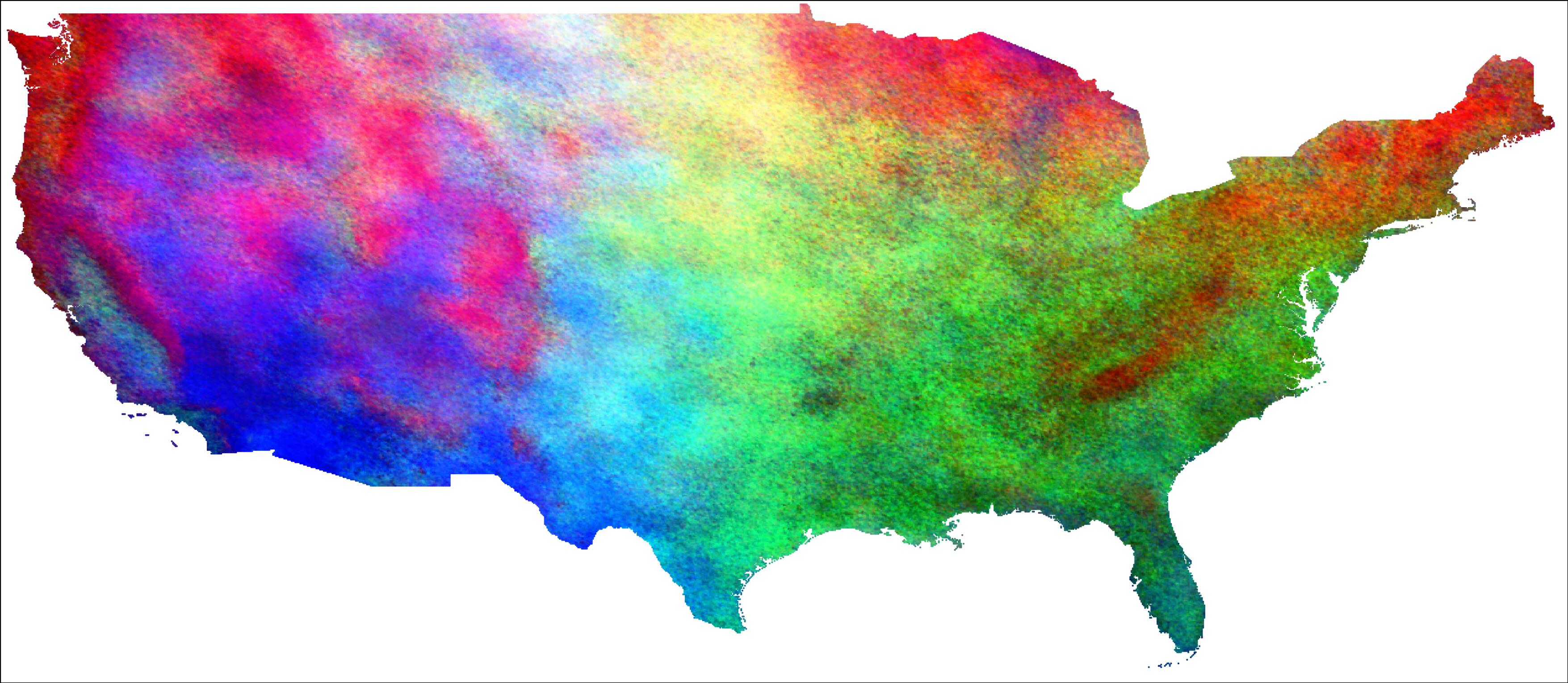}
  \caption*{\small(d) ProM3E (Register token \#2)}
  \end{minipage}
  \begin{minipage}[t]{0.328\linewidth}
  \centering
  \includegraphics[width=0.72\linewidth]{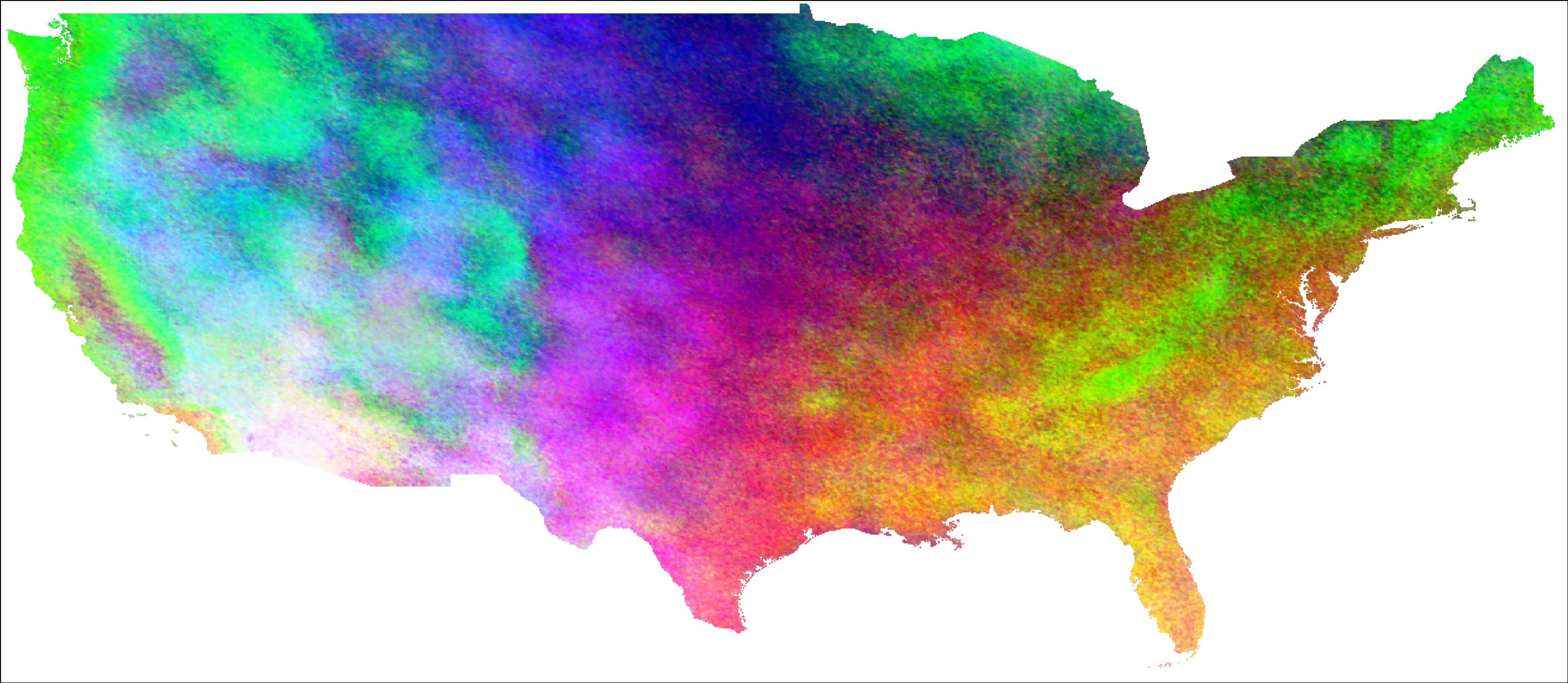}
  \caption*{\small(e) ProM3E (Register token \#3)}
  \end{minipage}
  \begin{minipage}[t]{0.328\linewidth}
  \centering
  \includegraphics[width=0.72\linewidth]{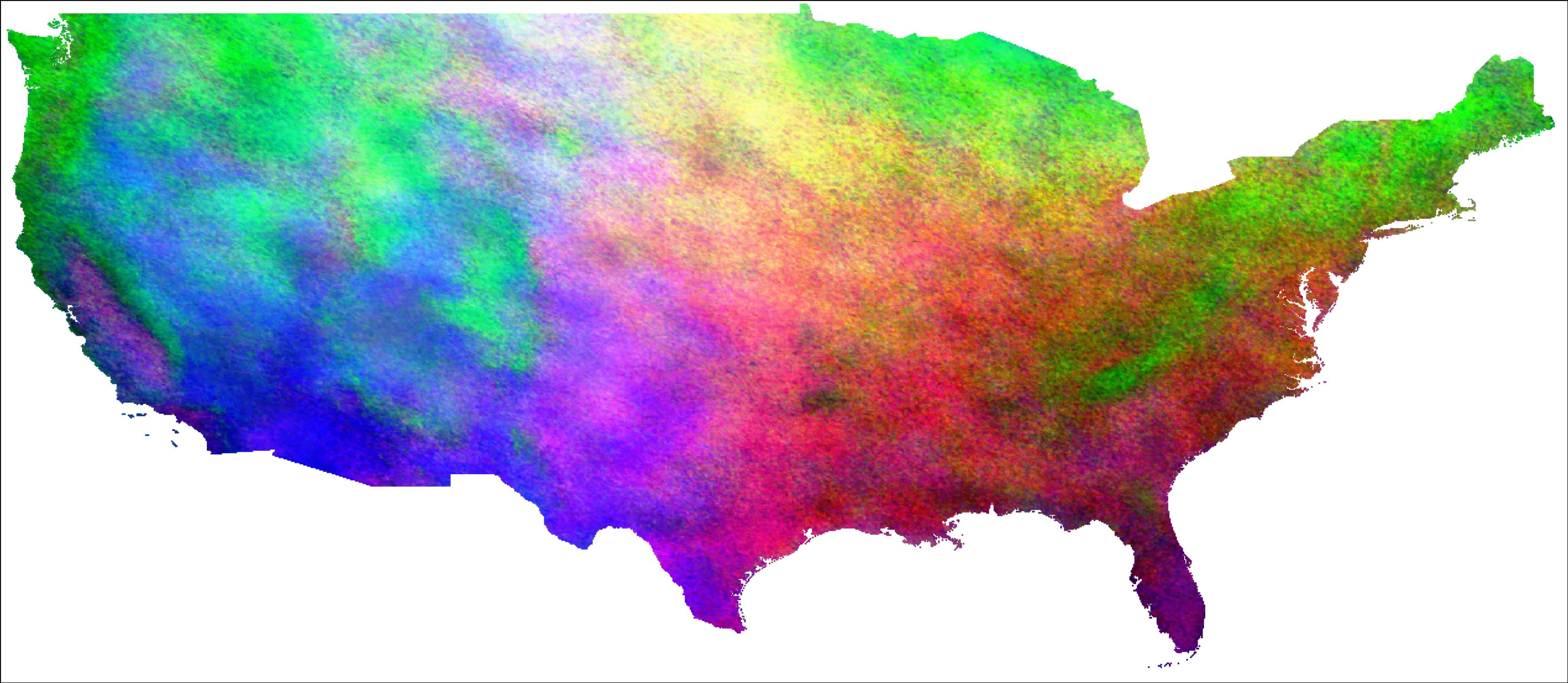}
  \caption*{\small(f) ProM3E (Register token \#4)}
  \end{minipage}
    
    \captionof{figure}{\textbf{ICA Plot of Location Embeddings}. We visually compare embeddings obtained from various tokens in the hidden representation of our model with the representation from TaxaBind. We notice that each register token captures different information.}

    \begin{minipage}[t]{0.328\linewidth}
  \centering
  \includegraphics[width=0.72\linewidth]
  {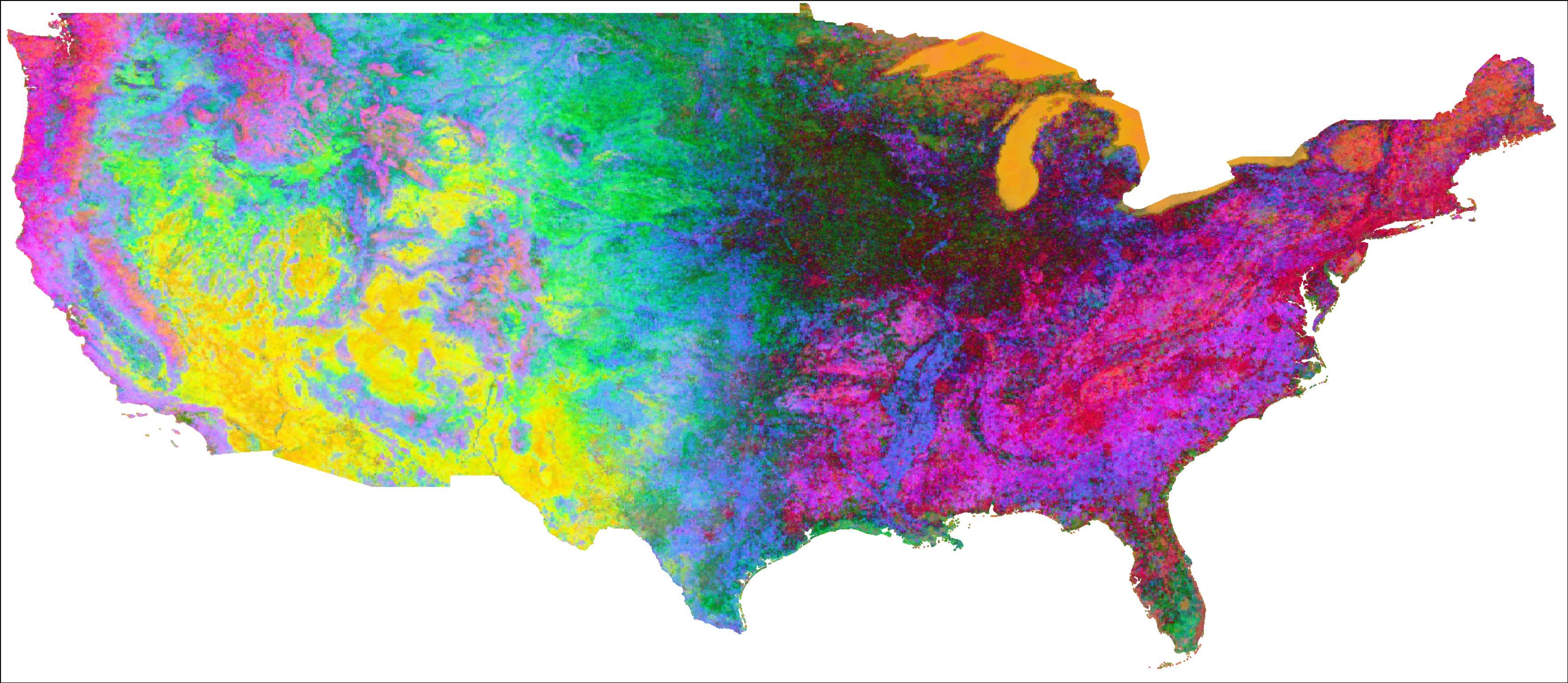}
  \caption*{\small(a) TaxaBind}
  \end{minipage}
  \begin{minipage}[t]{0.328\linewidth}
  \centering
  \includegraphics[width=0.72\linewidth]{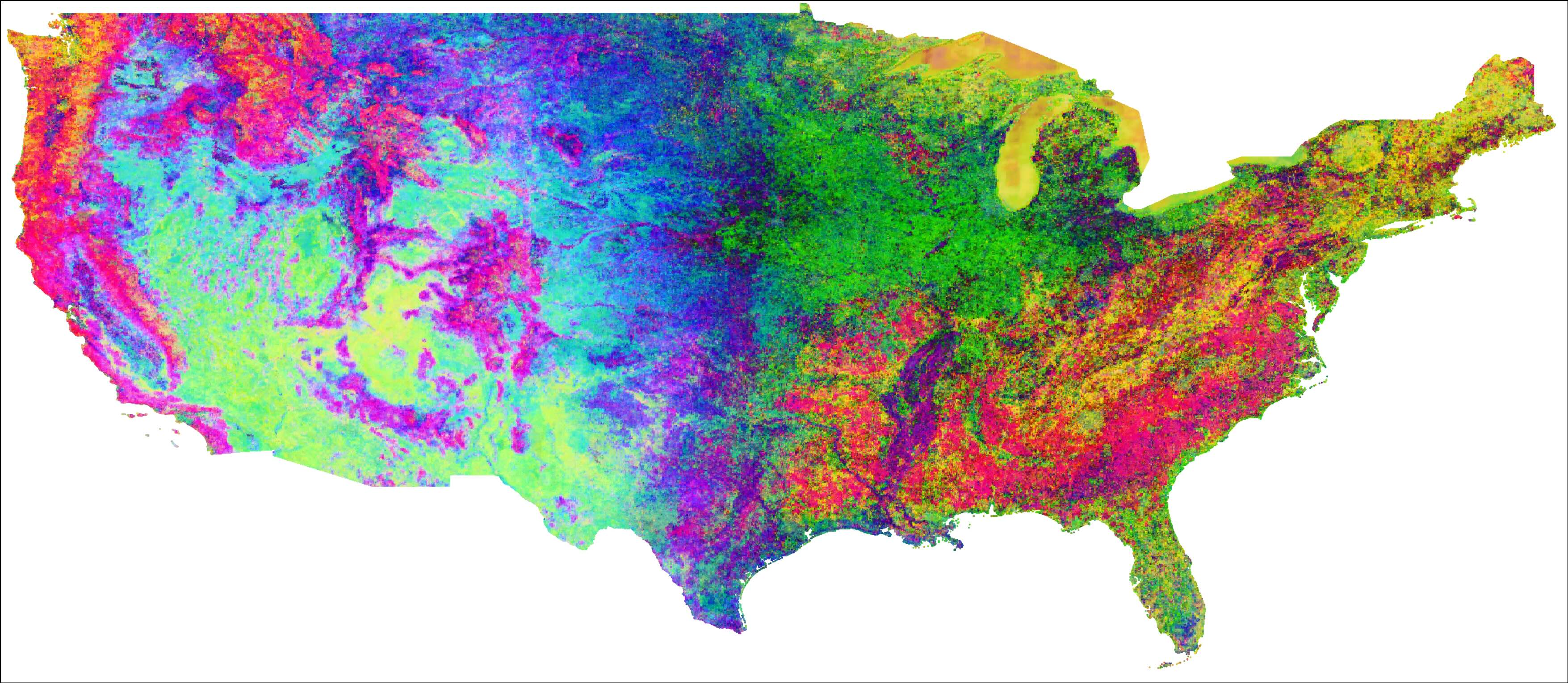}
  \caption*{\small(b) ProM3E (Sat. token)}
  \end{minipage}
  \begin{minipage}[t]{0.328\linewidth}
  \centering
  \includegraphics[width=0.72\linewidth]{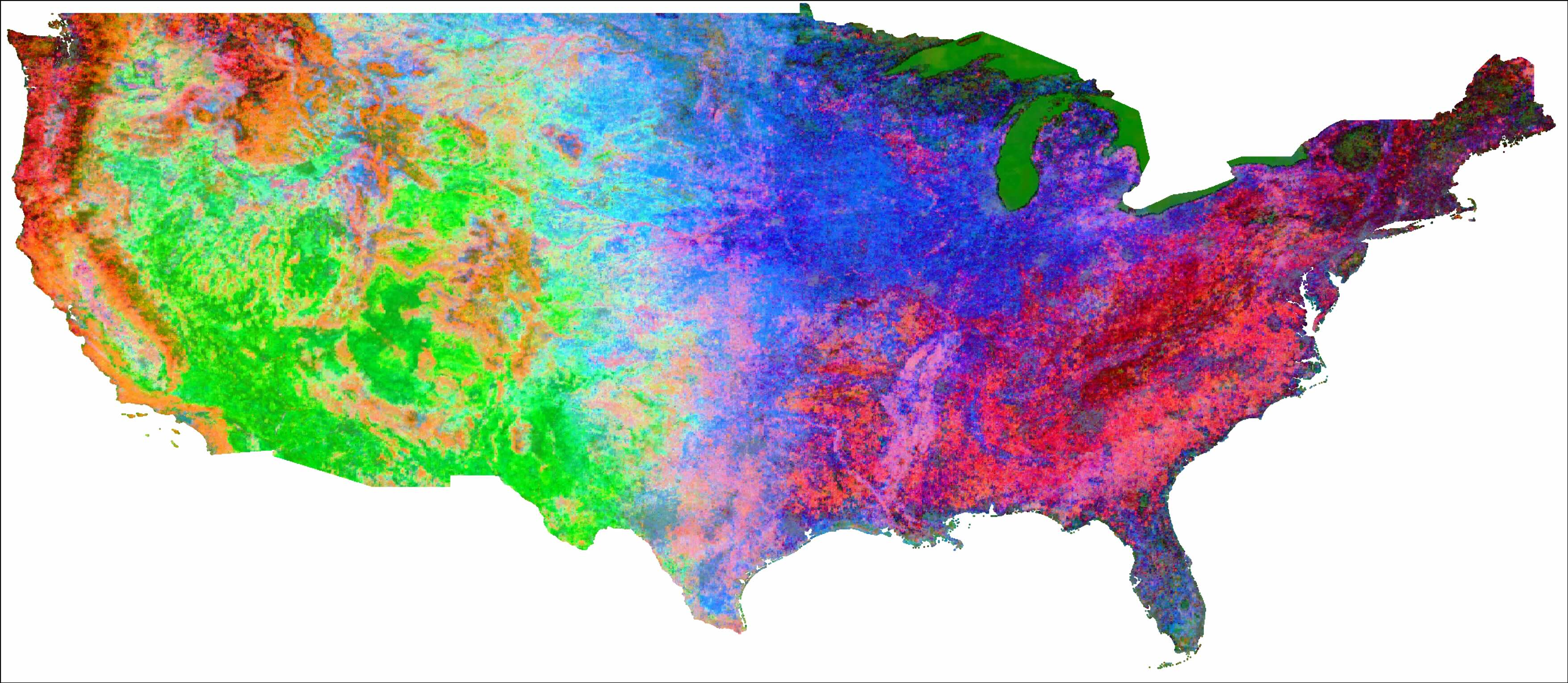}
  \caption*{\small(c) ProM3E (Register token \#1)}
  \end{minipage}
  \begin{minipage}[t]{0.328\linewidth}
  \centering
  \includegraphics[width=0.72\linewidth]
  {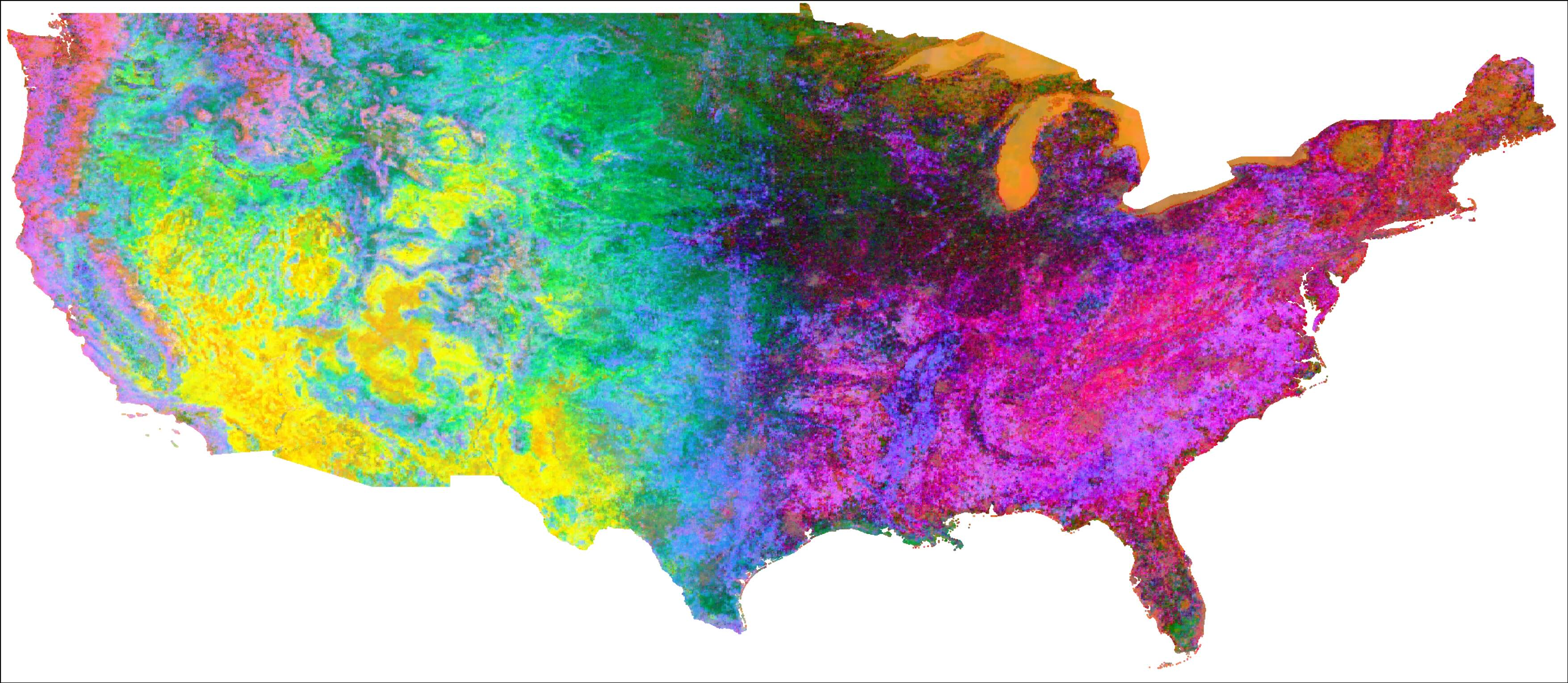}
  \caption*{\small(d) ProM3E (Register token \#2)}
  \end{minipage}
  \begin{minipage}[t]{0.328\linewidth}
  \centering
  \includegraphics[width=0.72\linewidth]{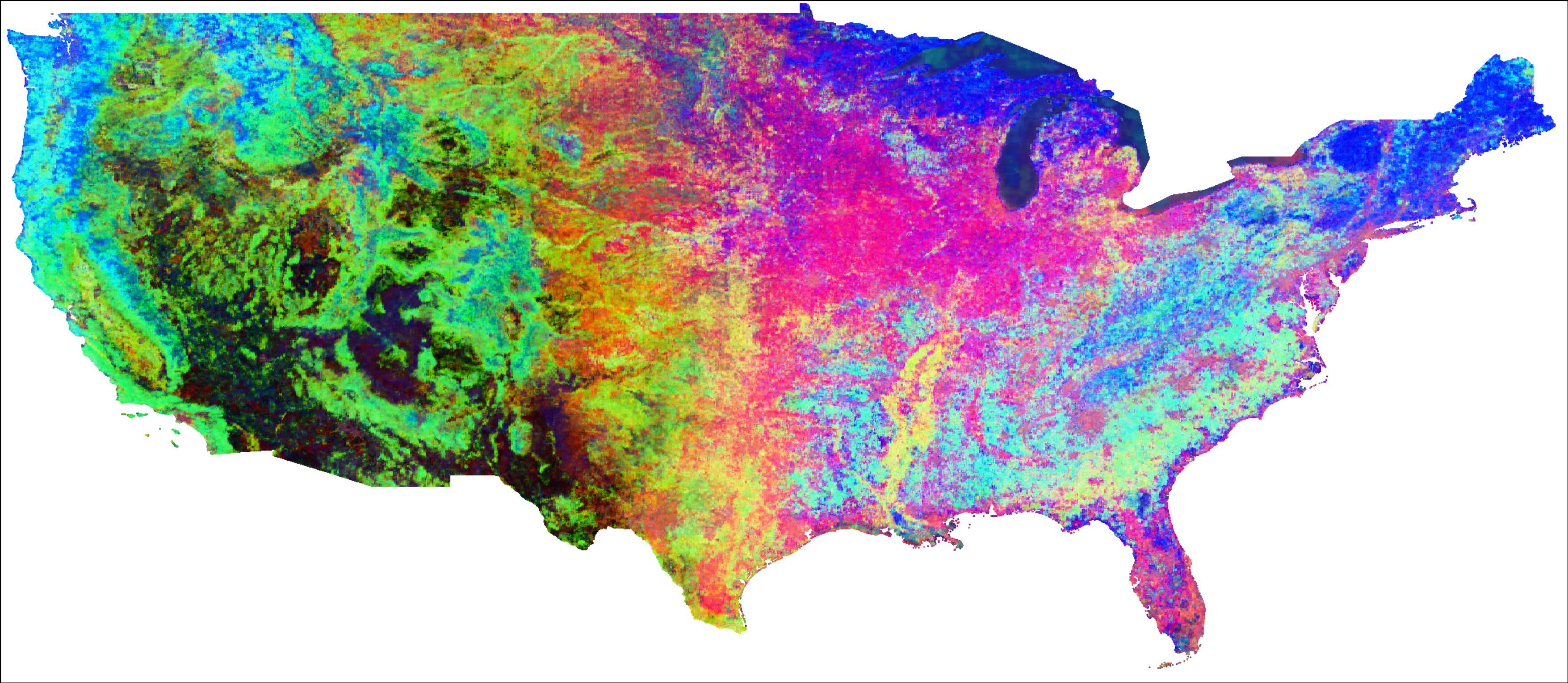}
  \caption*{\small(e) ProM3E (Register token \#3)}
  \end{minipage}
  \begin{minipage}[t]{0.328\linewidth}
  \centering
  \includegraphics[width=0.72\linewidth]{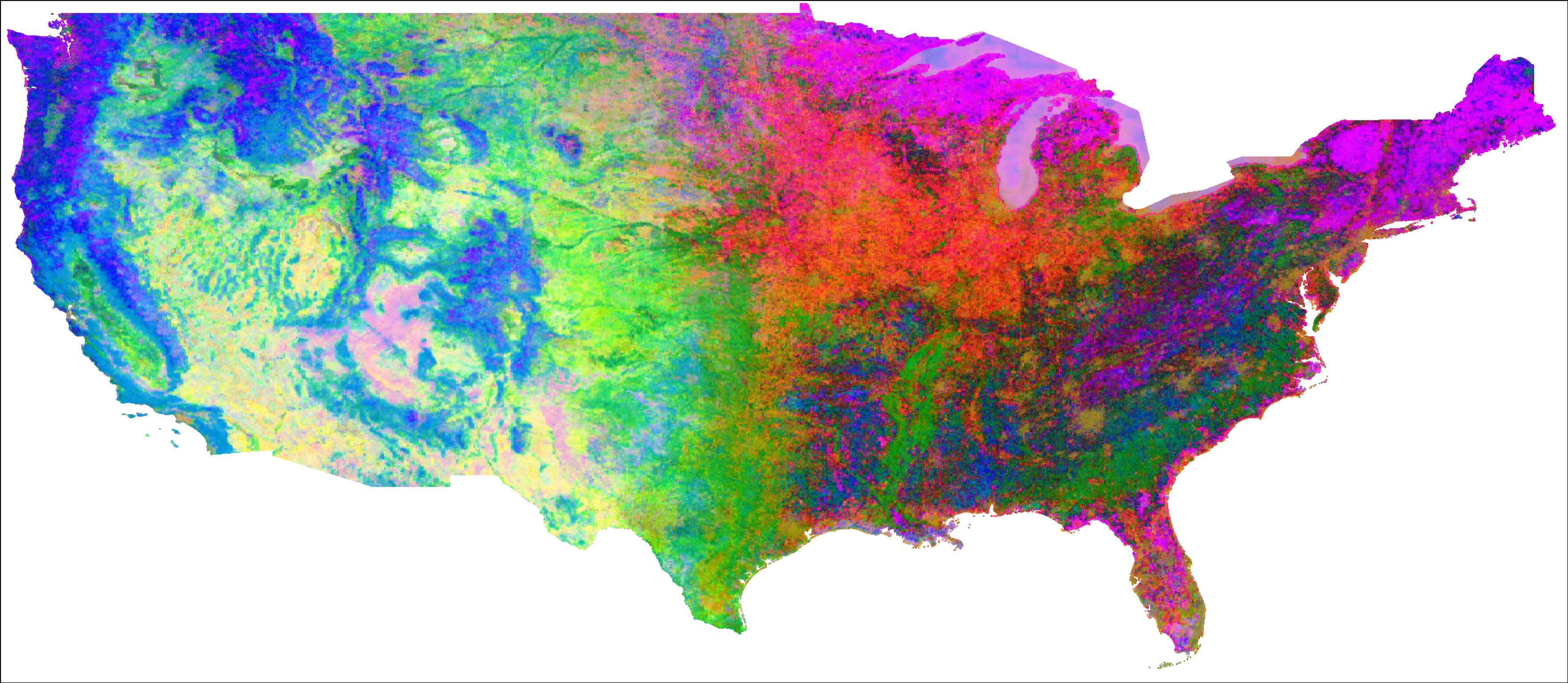}
  \caption*{\small(f) ProM3E (Register token \#4)}
  \end{minipage}
\captionof{figure}{\textbf{ICA Plot of Satellite Image Embeddings}. Similary, we compare satellite image embeddings with TaxaBind and notice register tokens capture diverse information.}

\begin{minipage}[t]{0.328\linewidth}
  \centering
  \includegraphics[width=0.72\linewidth]
  {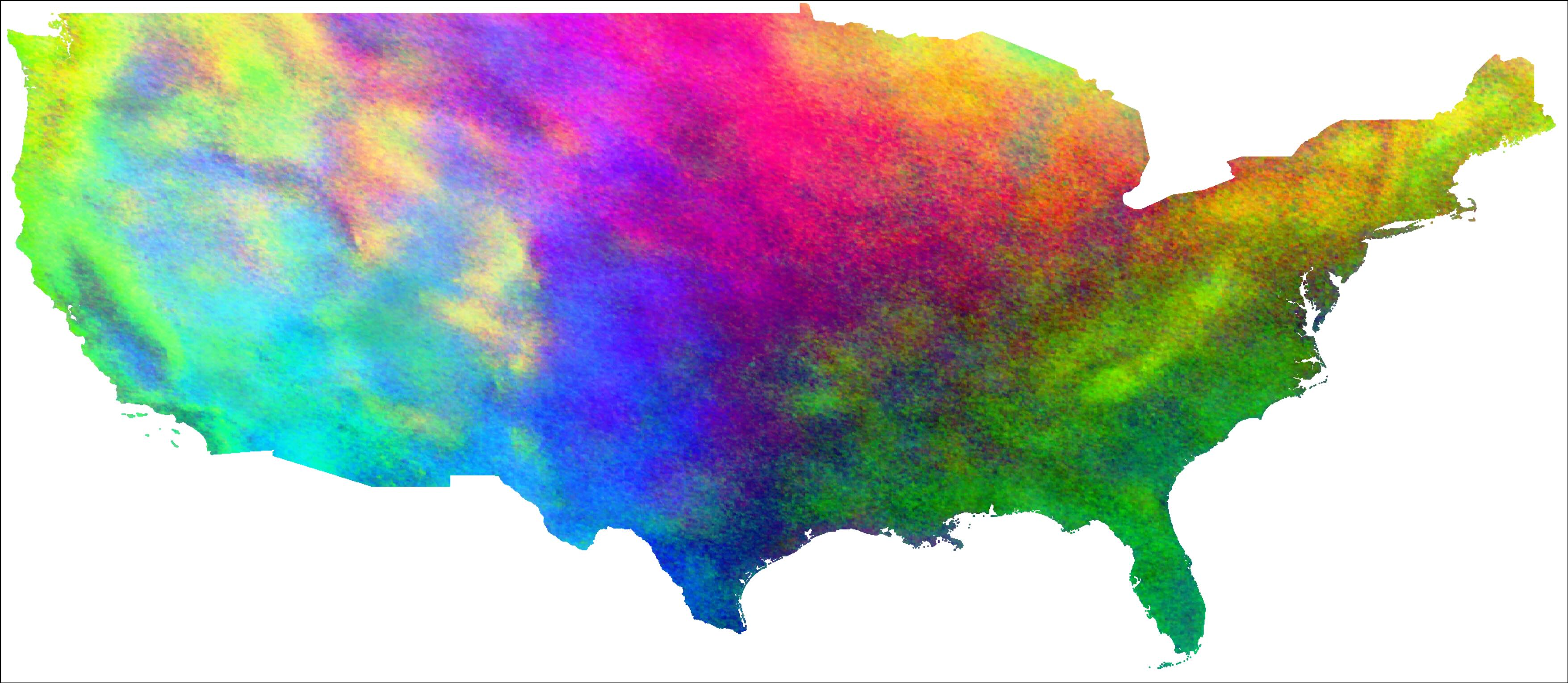}
  \caption*{\small(a) Location $\mu$ token}
  \end{minipage}
  \begin{minipage}[t]{0.328\linewidth}
  \centering
  \includegraphics[width=0.72\linewidth]{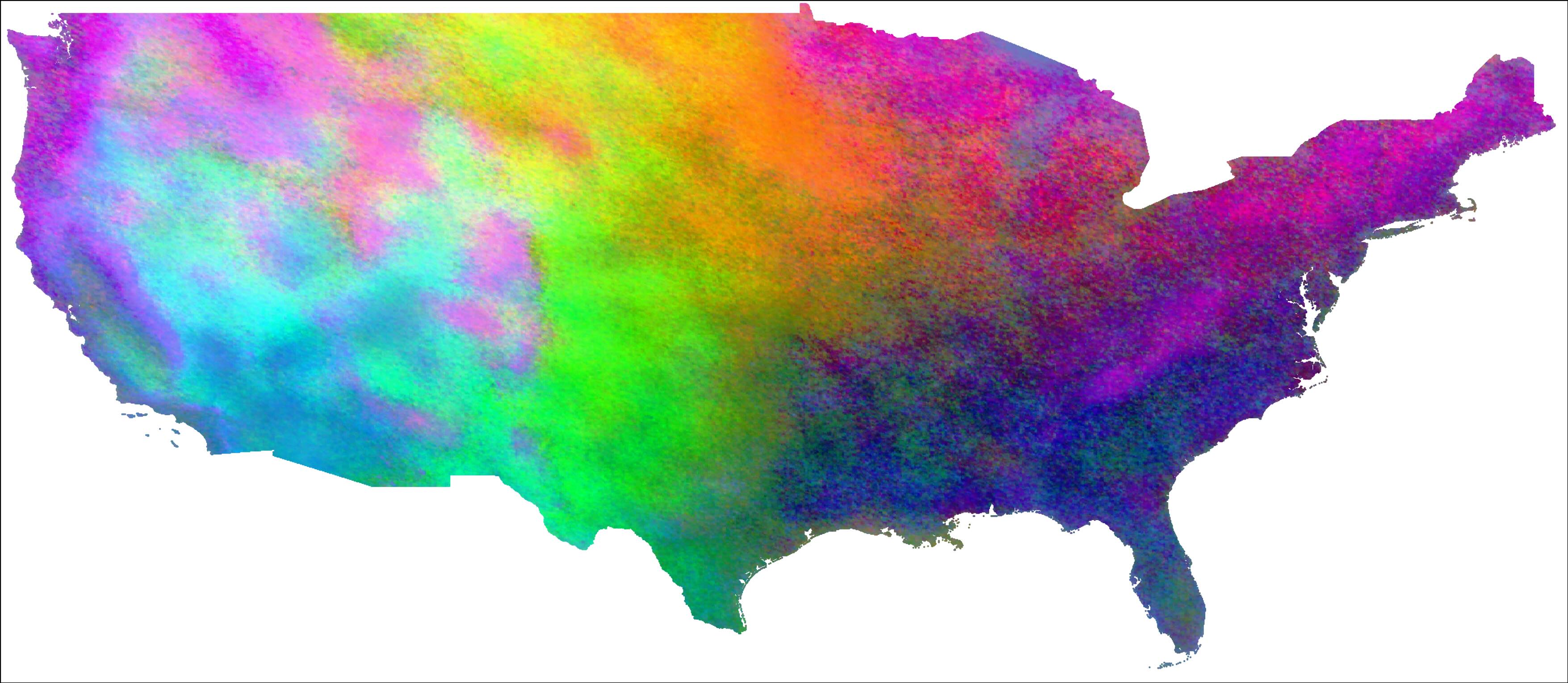}
  \caption*{\small(b) Location $\sigma$ token}
  \end{minipage}\\
  \begin{minipage}[t]{0.328\linewidth}
  \centering
  \includegraphics[width=0.72\linewidth]
  {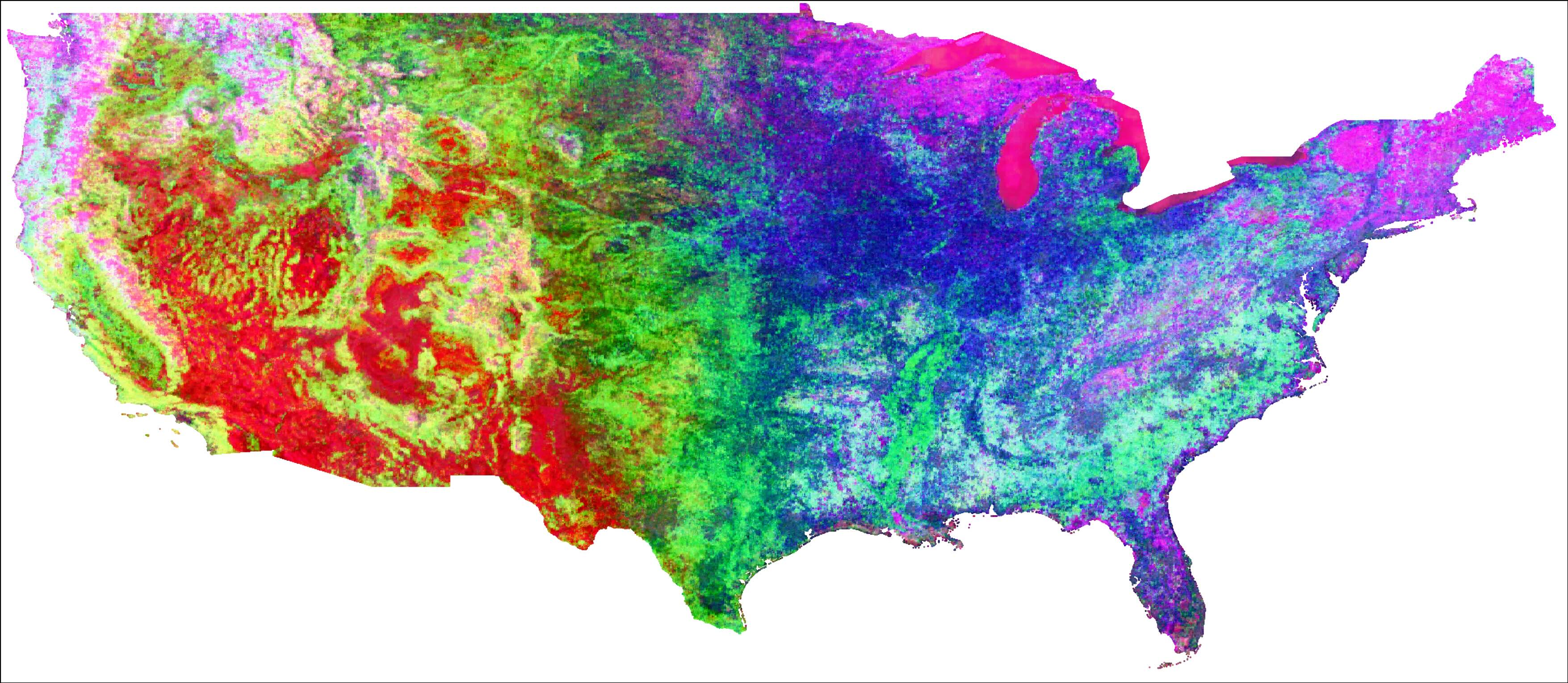}
  \caption*{\small(c) Satellite $\mu$ token}
  \end{minipage}
  \begin{minipage}[t]{0.328\linewidth}
  \centering
  \includegraphics[width=0.72\linewidth]{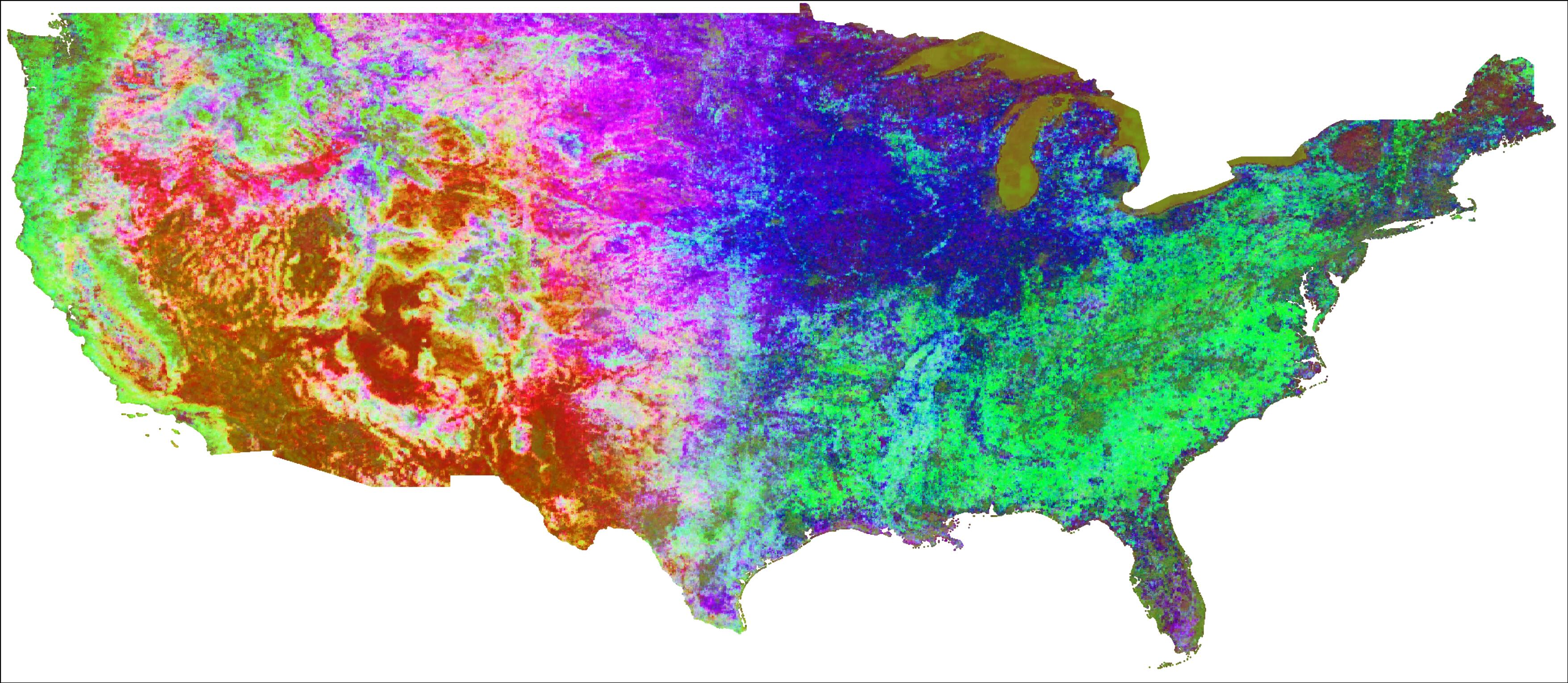}
  \caption*{\small(d) Satellite $\sigma$ token}
  \end{minipage}
\captionof{figure}{\textbf{ICA Plot of \textit{[$\mu$]} and \textit{[$\sigma$]} Tokens}. We plot the representations obtained from \textit{[$\mu$]} and \textit{[$\sigma$]} tokens for geographic location and satellite images across the USA.}
\end{center}%
}]

\subsection{Species Distribution Mapping}
\begin{figure*}
\begin{minipage}[t]{0.48\linewidth}
  \centering
  \includegraphics[width=0.88\linewidth]
  {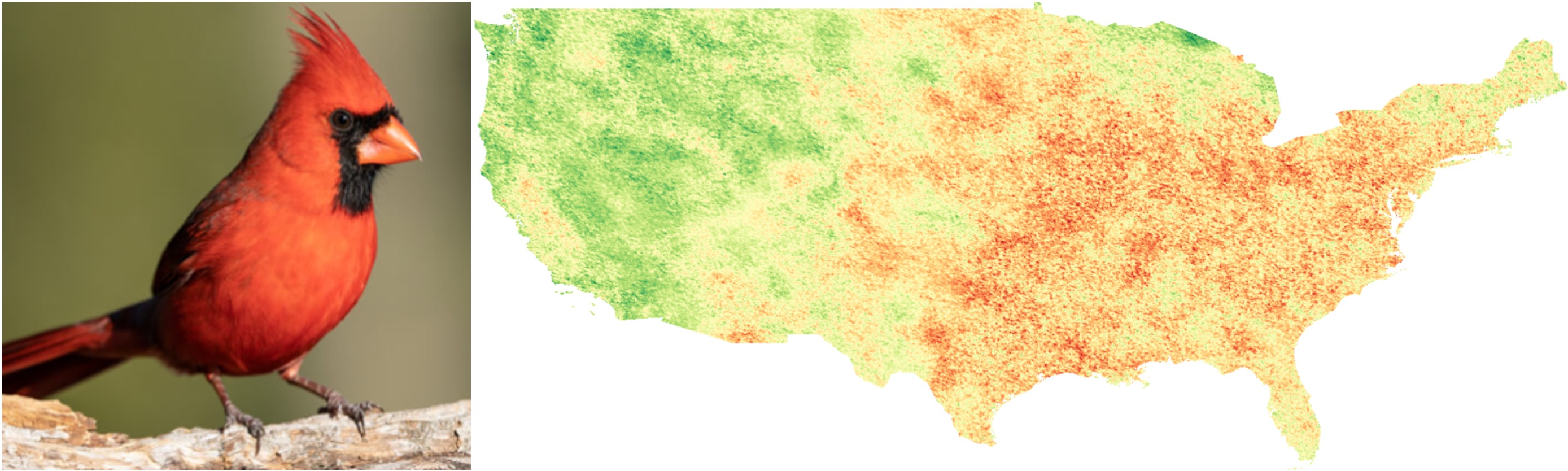}
  \caption*{\small(a) Northern Cardinal}
  \end{minipage}
  \begin{minipage}[t]{0.48\linewidth}
  \centering
  \includegraphics[width=0.88\linewidth]{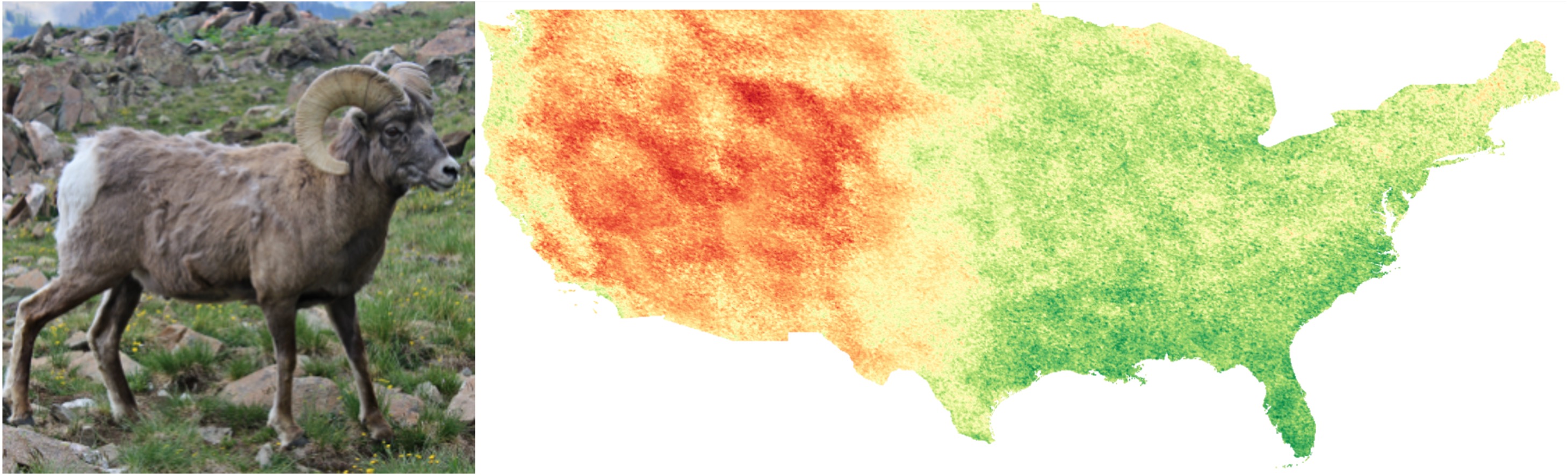}
  \caption*{\small(b) Bighorn Sheep}
  \end{minipage}\\
  \begin{minipage}[t]{0.48\linewidth}
  \centering
  \includegraphics[width=0.88\linewidth]{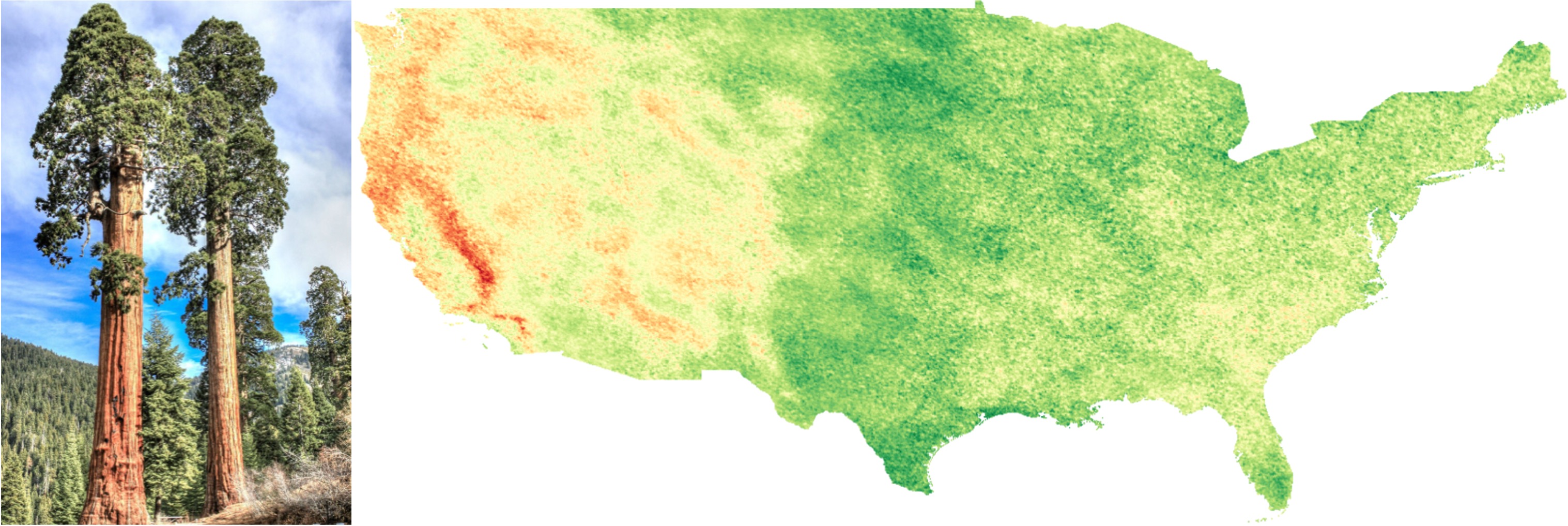}
  \caption*{\small(c) Giant Sequoia Trees}
  \end{minipage}
  \begin{minipage}[t]{0.48\linewidth}
  \centering
  \includegraphics[width=0.88\linewidth]{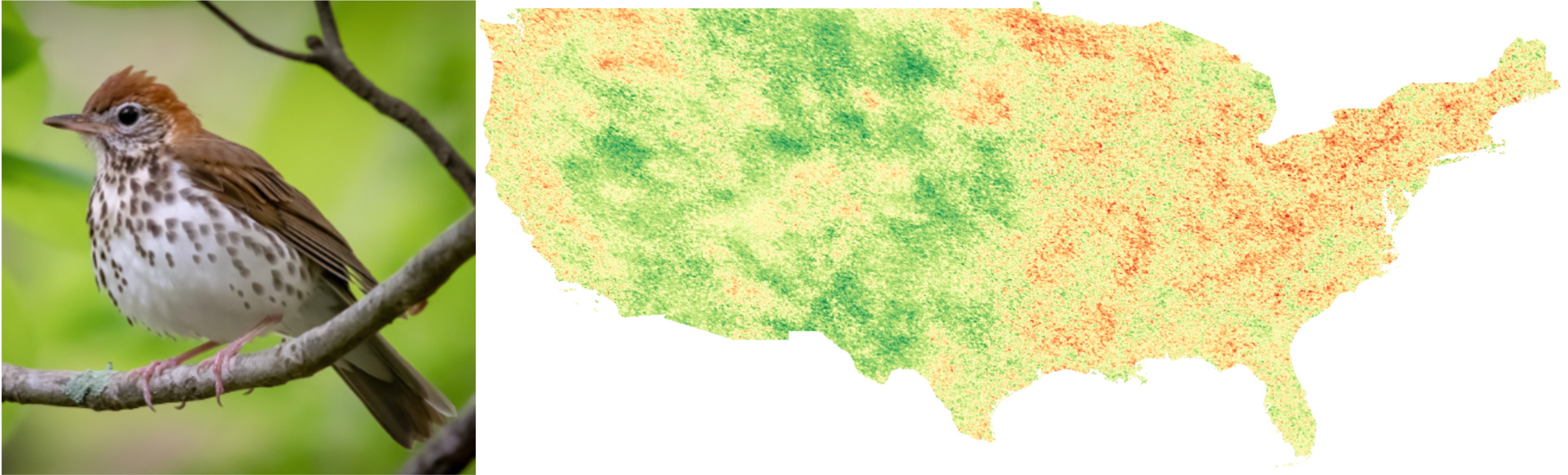}
  \caption*{\small(d) Wood Thrush}
  \end{minipage}\\
  \begin{minipage}[t]{0.48\linewidth}
  \centering
  \includegraphics[width=0.88\linewidth]{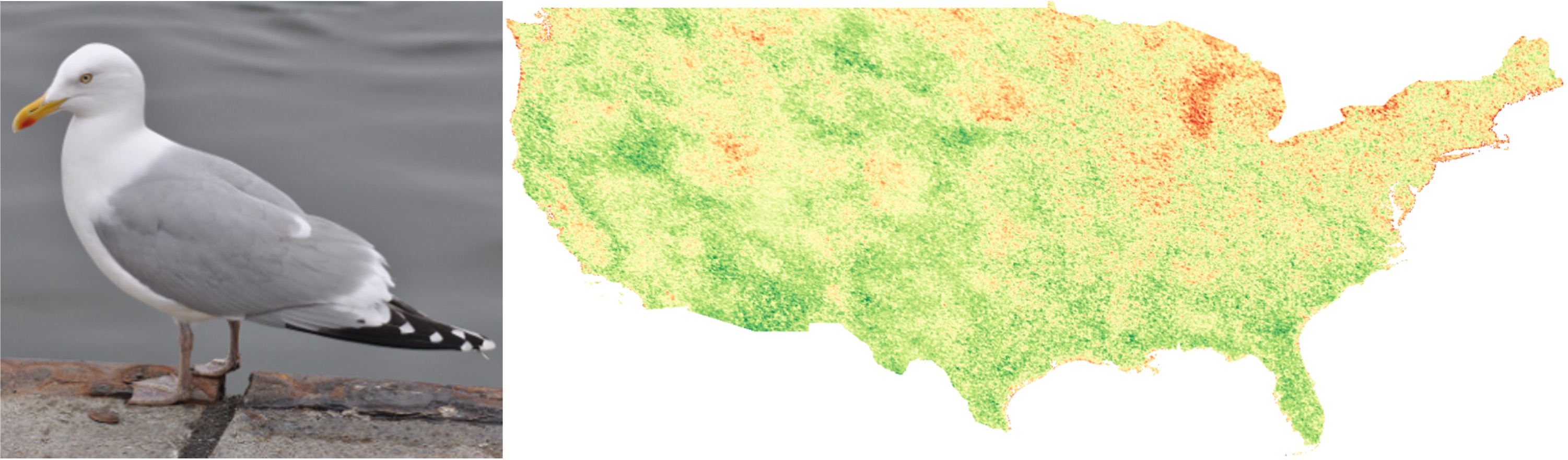}
  \caption*{\small(e) European Herring Gull}
  \end{minipage}
  \begin{minipage}[t]{0.48\linewidth}
  \centering
  \includegraphics[width=0.88\linewidth]{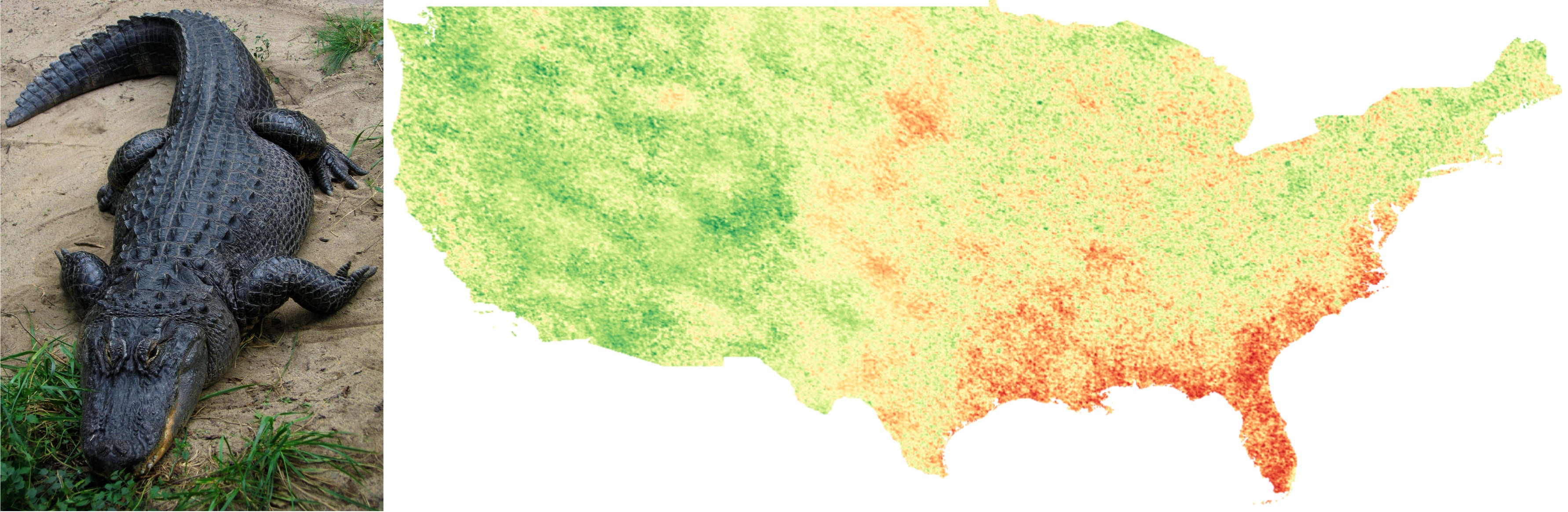}
  \caption*{\small(f) American Alligator}
  \end{minipage}\\
  \begin{minipage}[t]{0.48\linewidth}
  \centering
  \includegraphics[width=0.88\linewidth]{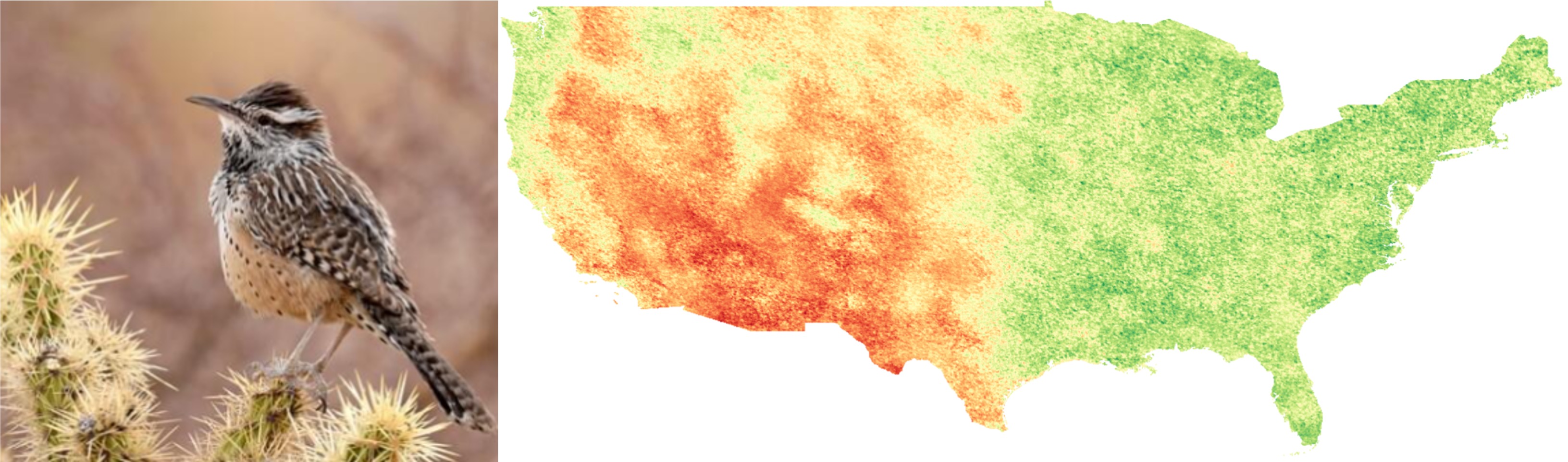}
  \caption*{\small(g) Cactus Wren}
  \end{minipage}
  \begin{minipage}[t]{0.48\linewidth}
  \centering
  \includegraphics[width=0.88\linewidth]{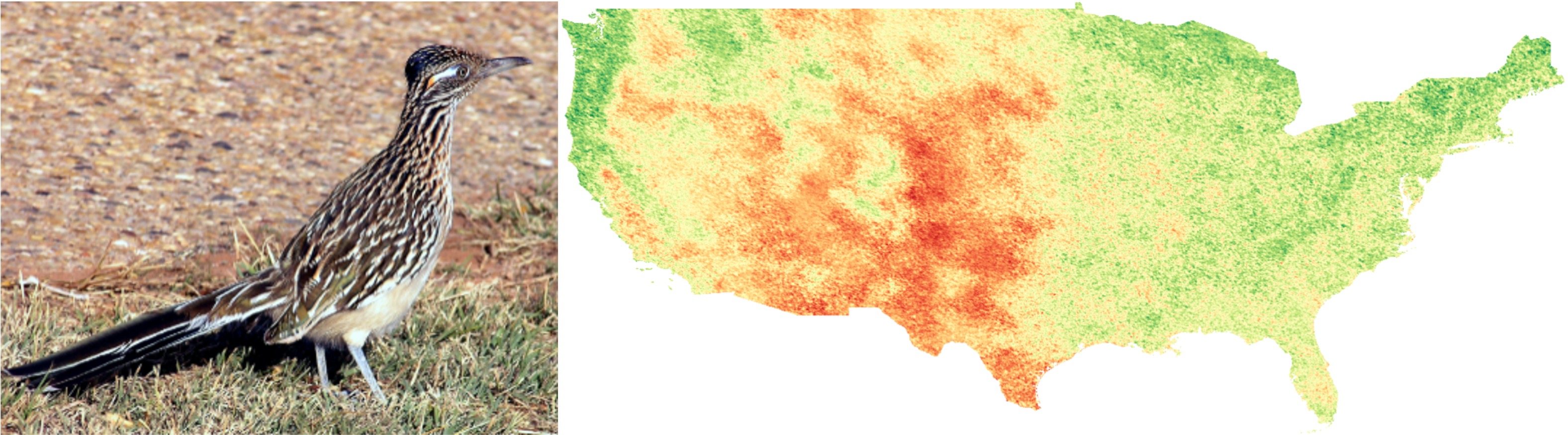}
  \caption*{\small(h) Greater Roadrunner}
  \end{minipage}\\
  \begin{minipage}[t]{0.48\linewidth}
  \centering
  \includegraphics[width=0.88\linewidth]{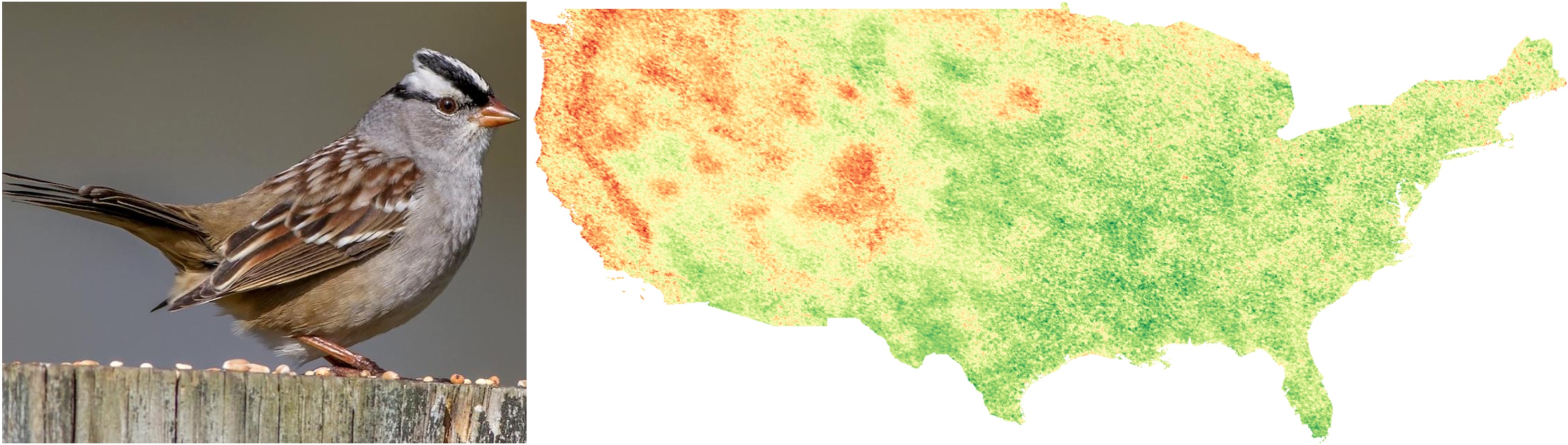}
  \caption*{\small(i) White Crowned Sparrow}
  \end{minipage}
  \begin{minipage}[t]{0.48\linewidth}
  \centering
  \includegraphics[width=0.88\linewidth]{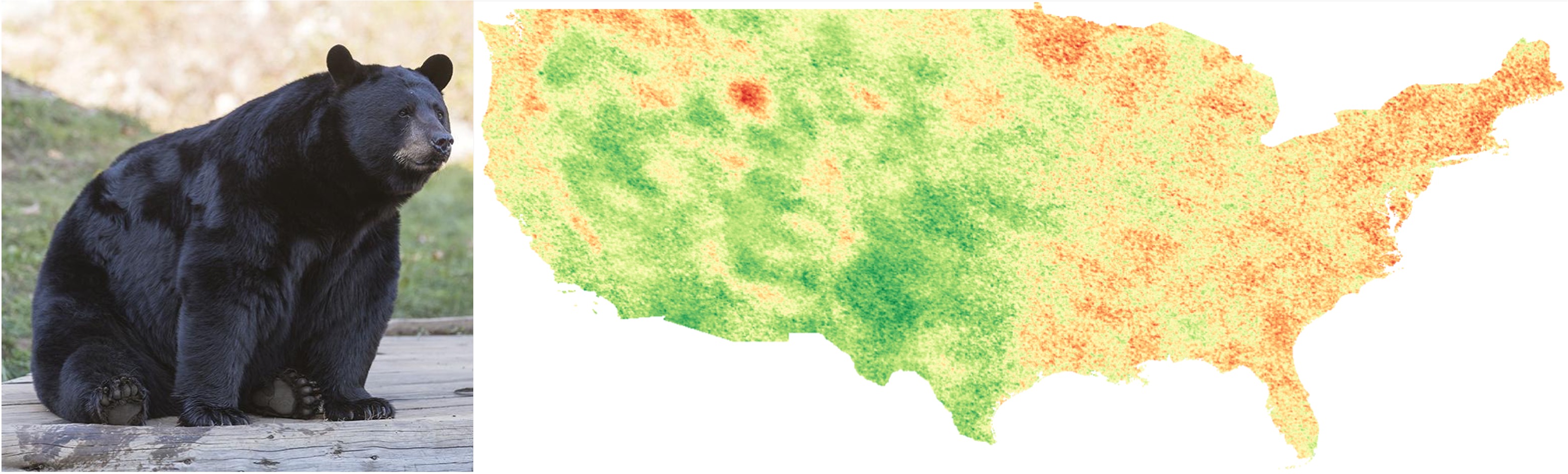}
  \caption*{\small(j) American Black Bear}
  \end{minipage}\\
  \begin{minipage}[t]{0.48\linewidth}
  \centering
  \includegraphics[width=0.88\linewidth]{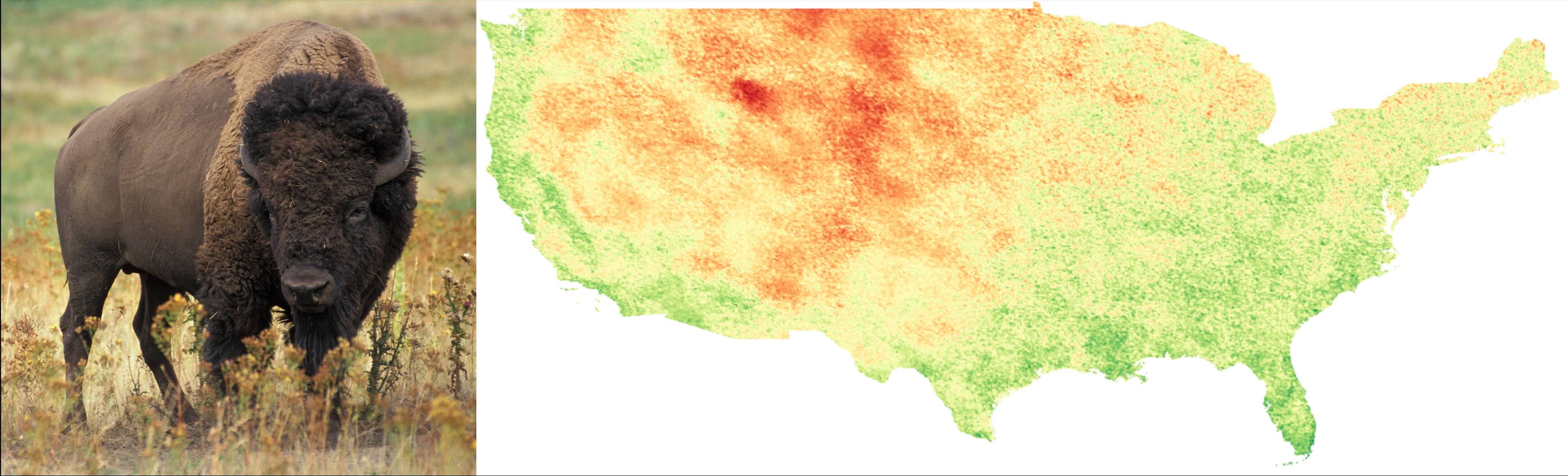}
  \caption*{\small(k) American Bison}
  \end{minipage}
  \begin{minipage}[t]{0.48\linewidth}
  \centering
  \includegraphics[width=0.88\linewidth]{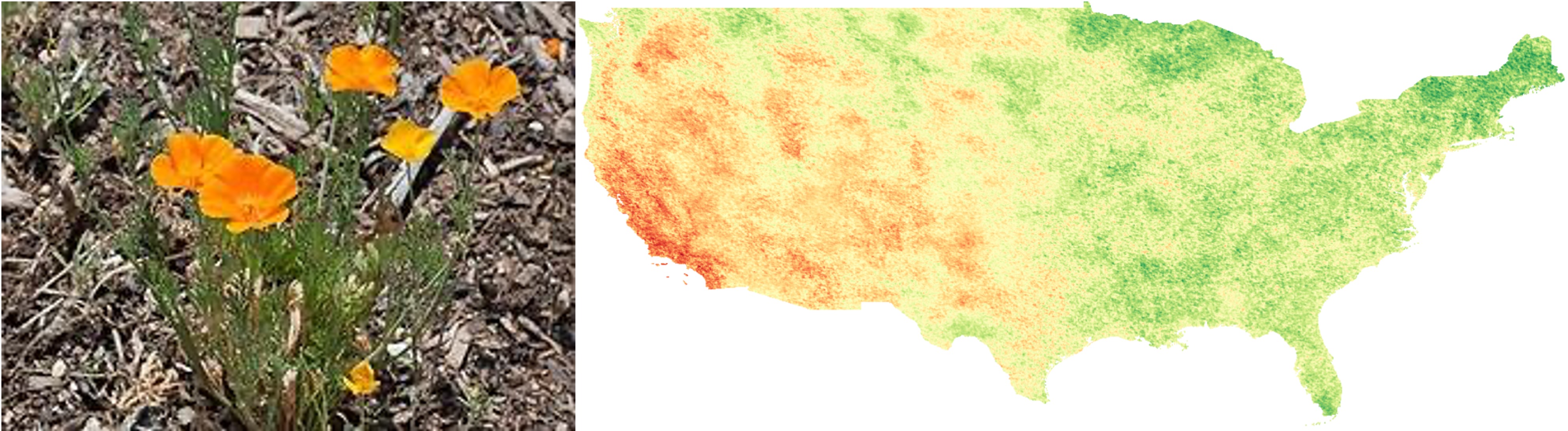}
  \caption*{\small(l) California Poppy}
  \end{minipage}\\
  \begin{minipage}[t]{0.48\linewidth}
  \centering
  \includegraphics[width=0.88\linewidth]{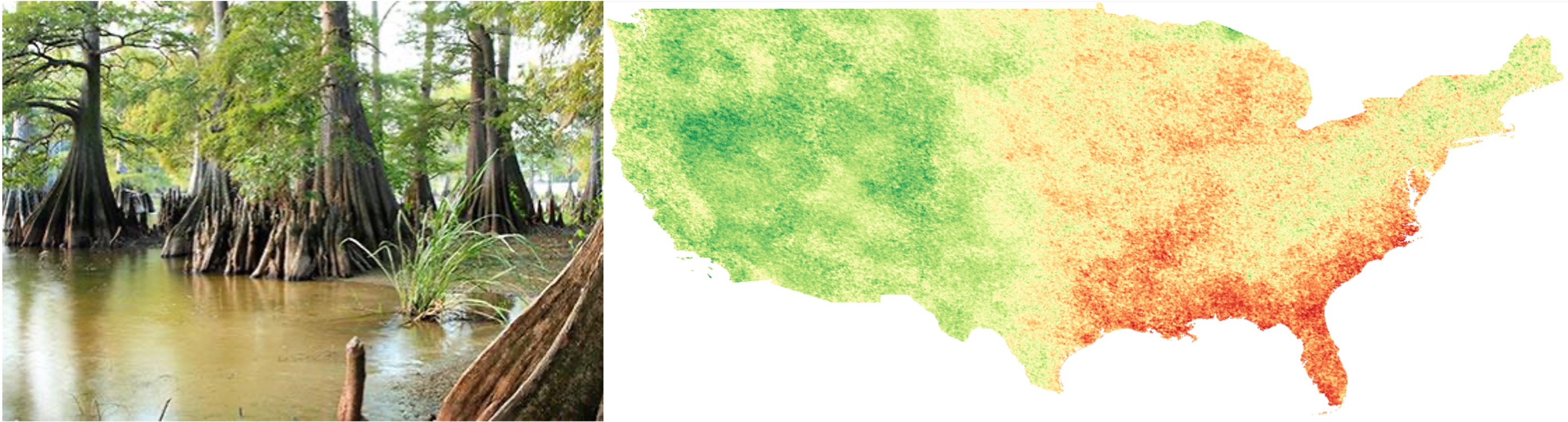}
  \caption*{\small(m) Bald Cypress}
  \end{minipage}
  \begin{minipage}[t]{0.48\linewidth}
  \centering
  \includegraphics[width=0.88\linewidth]{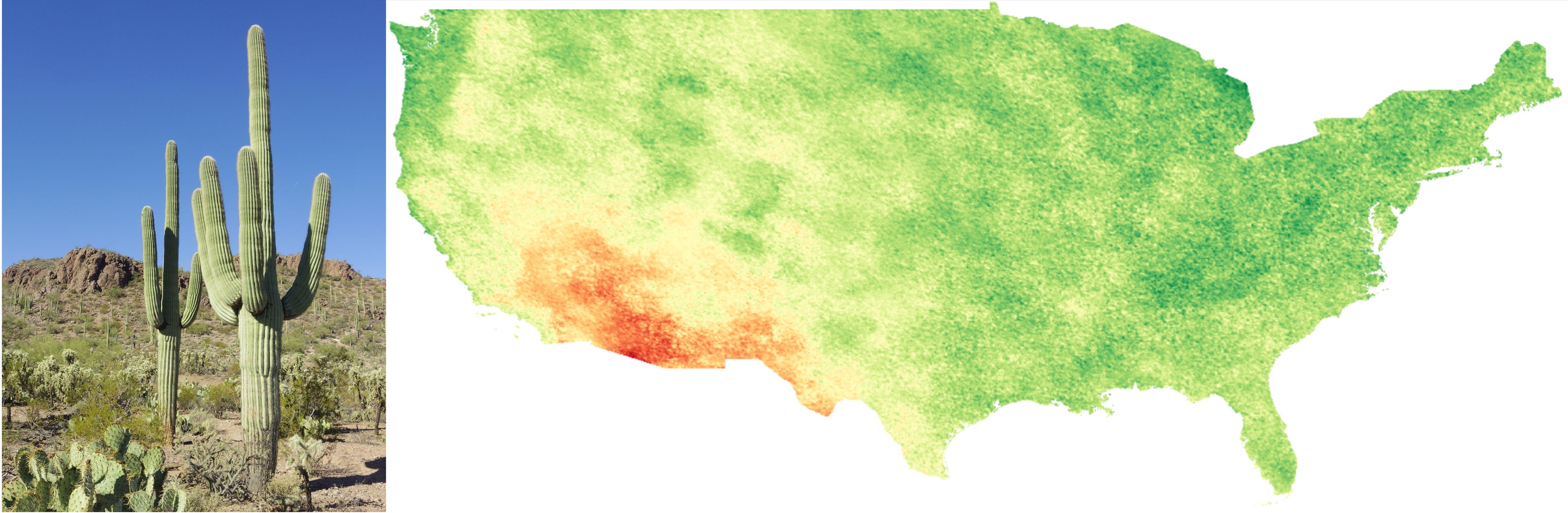}
  \caption*{\small(n) Saguaro Cactus}
  \end{minipage}
  \caption{\textbf{High Resolution Species Distribution Mapping using Ground-Level Imagery.} We create species distribution maps by computing the similarity between query ground-level image and geo-locations sampled uniformly across USA.}
  \label{sdm:hr}
\end{figure*}

In Figure~\ref{sdm:hr}, we show species distribution maps generated using our model given a query ground-level image depicting a species. We create a dense grid of satellite imagery over the USA and compute ProM3E embeddings for each location on the grid. We then compute cosine similarity of the query image and the embeddings of each location.

\begin{figure*}[!ht]
\centering
  \begin{minipage}[t]{0.328\linewidth}
  \centering
  \includegraphics[width=\linewidth]
  {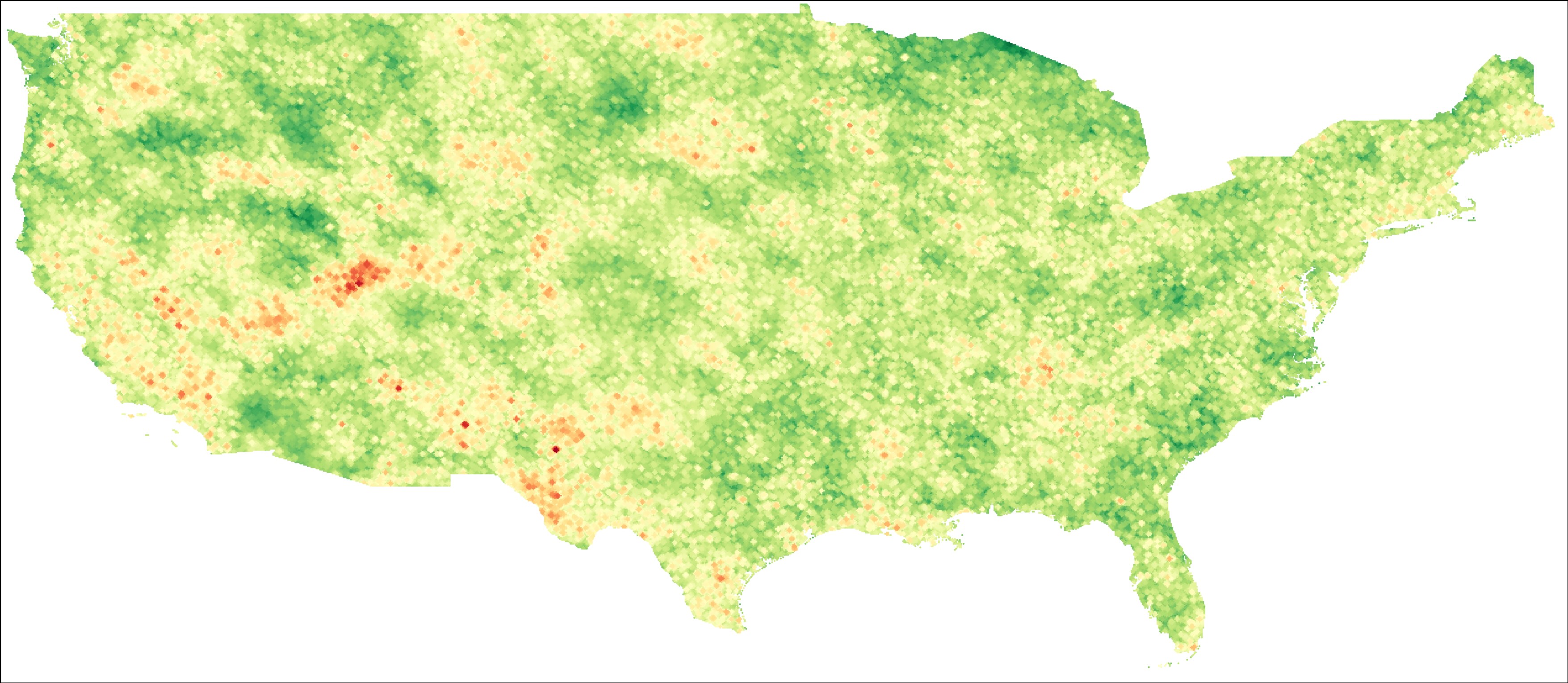}
  \caption*{\small(a) $||\sigma||_1$}
  \end{minipage}
  \begin{minipage}[t]{0.328\linewidth}
  \centering
  \includegraphics[width=\linewidth]{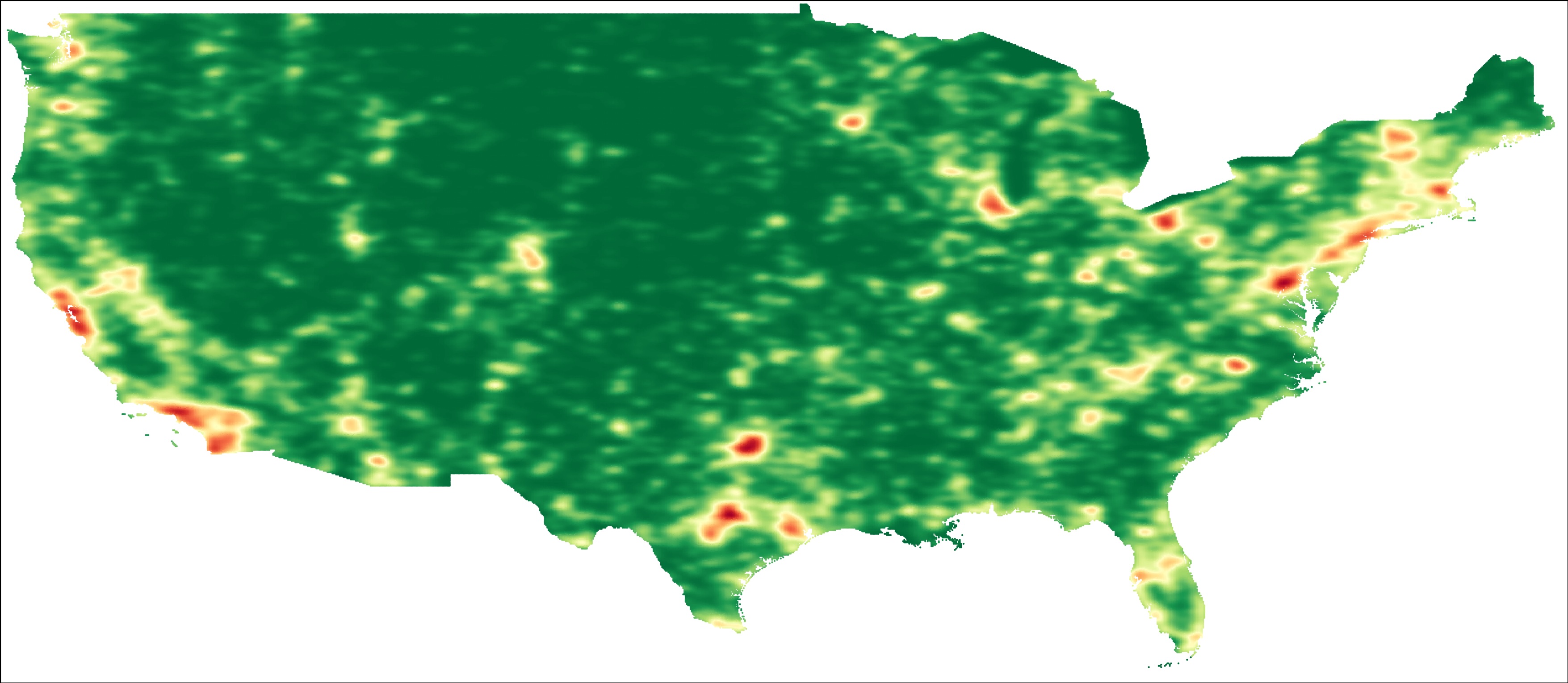}
  \caption*{\small(b) Shannon Diversity Index}
  \end{minipage}
  \begin{minipage}[t]{0.328\linewidth}
  \centering
  \includegraphics[width=\linewidth]{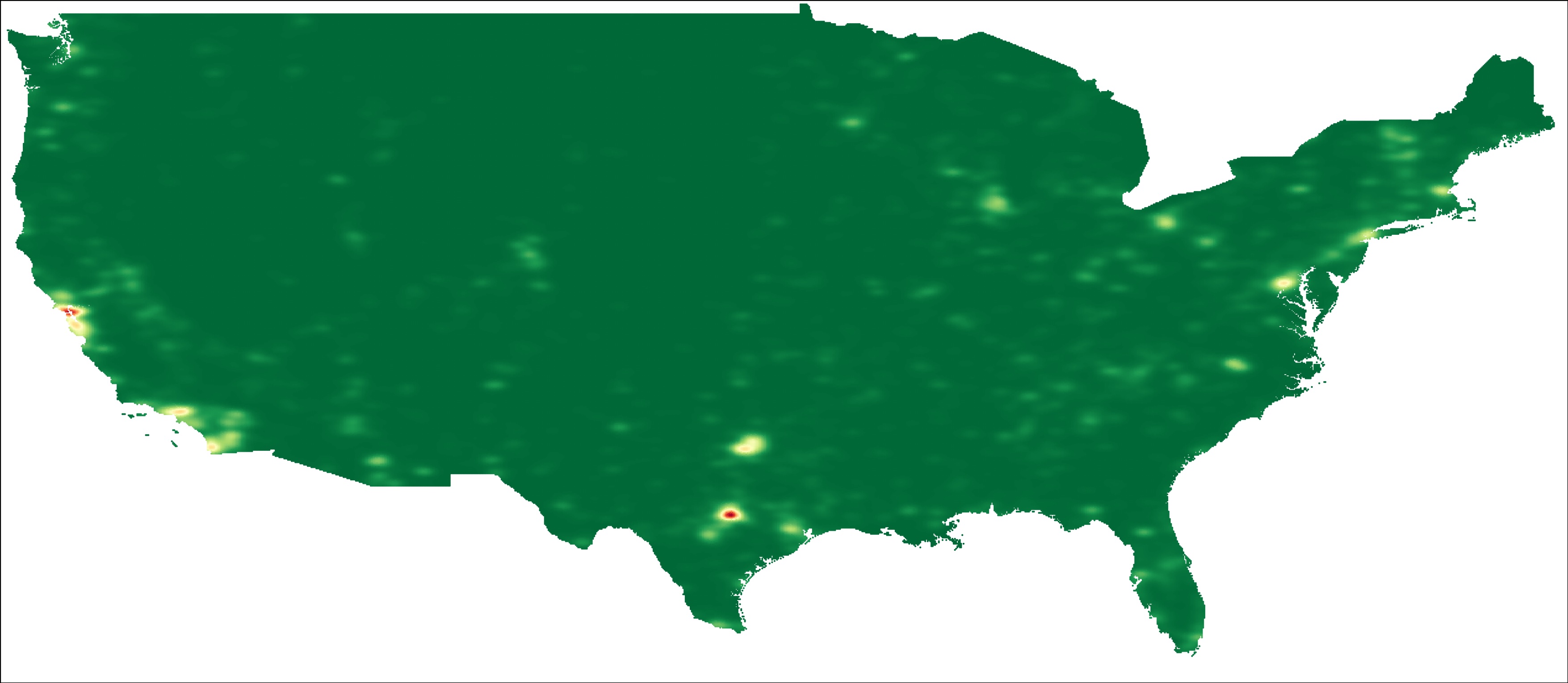}
  \caption*{\small(c) Species Richness}
  \end{minipage}
\caption{\textbf{Species Biodiversity Maps}. We plot $||\sigma||_1$ values predicted by our model and compare it with shannon diversity index and species richness maps derived from iNaturalist observations. The maps are plotted using a rectangular grid of 250x500 points over USA.}
\label{fig:biod_sigma}
\end{figure*}

\subsection{Generating an iNaturalist Species Diversity Map}
\label{biod_map}
We create species diversity and richness maps of the USA using iNaturalist observations. To create the maps, we first filtered observations to include only those within the contiguous United States (excluding Alaska and Hawaii). We then employed a spatial analysis technique that divided the US territory into a 250 × 500 grid based on the geographic bounding coordinates of the USA. We then filtered out all cells falling outside the USA. For each grid cell, we identified and counted the number of unique species observed by mapping latitude and longitude coordinates to their corresponding grid indices. For calculating species diversity, we used the Shannon index, which computes the entropy in the species distribution. The species richness is calculated as the number of unique species present within each grid cell. To each of the maps, we applied a kernel density estimation (KDE) based Gaussian smoothing with a sigma parameter of 2.0, which smoothed the discrete data across neighboring cells. 

Additionally, we generate an uncertainty map of the USA by computing the $||\sigma||_1$ value at each grid cell. We then compare all the generated maps visually. We visualize the maps in Figure~\ref{fig:biod_sigma}. Remember, in section, we conducted a quantitative comparison between $||\sigma||_1$ and Shannon diversity index and found a significant positive correlation between them. The maps in the figure show similarities visually. This is in agreement with the quantitative analyses conducted in the previous sections. 

\section{Dataset Details}
\subsection{Training Datasets}
ProM3E has a flexible two-stage framework that can be trained independently. The first stage allows for training on large-scale image-paired datasets while the second stage requires an all paired dataset of all modalities for training. The first stage involves training modality-specific encoders using TaxaBind recipe. Below we present the details for the pretraining datasets used in stage one.

\textbf{TreeofLife-10M.} This dataset is composed of 10M pairs of species images and their corresponding taxonomic labels derived from open databases such as . It was introduced by \citet{stevens2024bioclip}, which was used to train BioCLIP. Here, we utilize BioCLIP's image-text frozen embedding space and project all other modalities to this space.

\textbf{iNaturalist-2021.} We use the iNaturalist-2021 dataset primarily to train for aligning geographic location and species images. This dataset consists of 2.7M images across 10k species categories captured around the globe. Each image is associated with metadata including geographic location, timestamp, etc.

\textbf{iSatNat.} \citet{sastry2025taxabind} curated a paired dataset of satellite and species images using the iNaturalist-2021 dataset. For each ground-level image, they download a 256x256 Sentinel-2 imagery. This dataset is used to align satellite image with species images. There are 2.55M samples for training, 134k samples for validation and 100k samples for testing.

\textbf{iSoundNat.} This dataset~\cite{sastry2025taxabind} consists of paired species images and audio downloaded from the iNaturalist platform. There are 74k samples for training, 4k samples for validation and 8k samples for testing.

\textbf{WorldClim-2.} Climatic variables derived from WorldClim-2 are used to align environmental covariates and species images. These are environmental covariates are curated for each species in the iNaturalis-2021 dataset. 
\begin{figure*}[!t]
\centering
  \begin{minipage}[t]{0.328\linewidth}
  \centering
  \includegraphics[width=\linewidth]
  {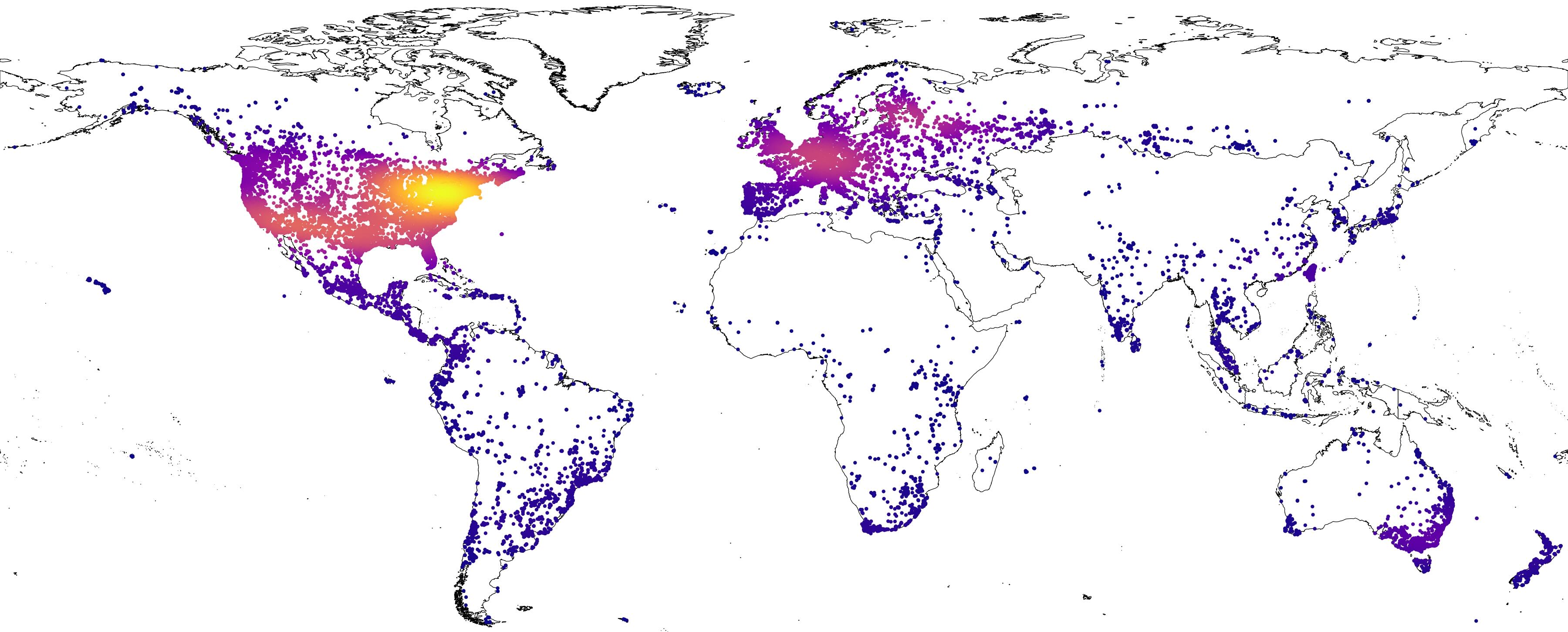}
  \caption*{\small(a) Train (MultiNat)}
  \end{minipage}
  \begin{minipage}[t]{0.328\linewidth}
  \centering
  \includegraphics[width=\linewidth]{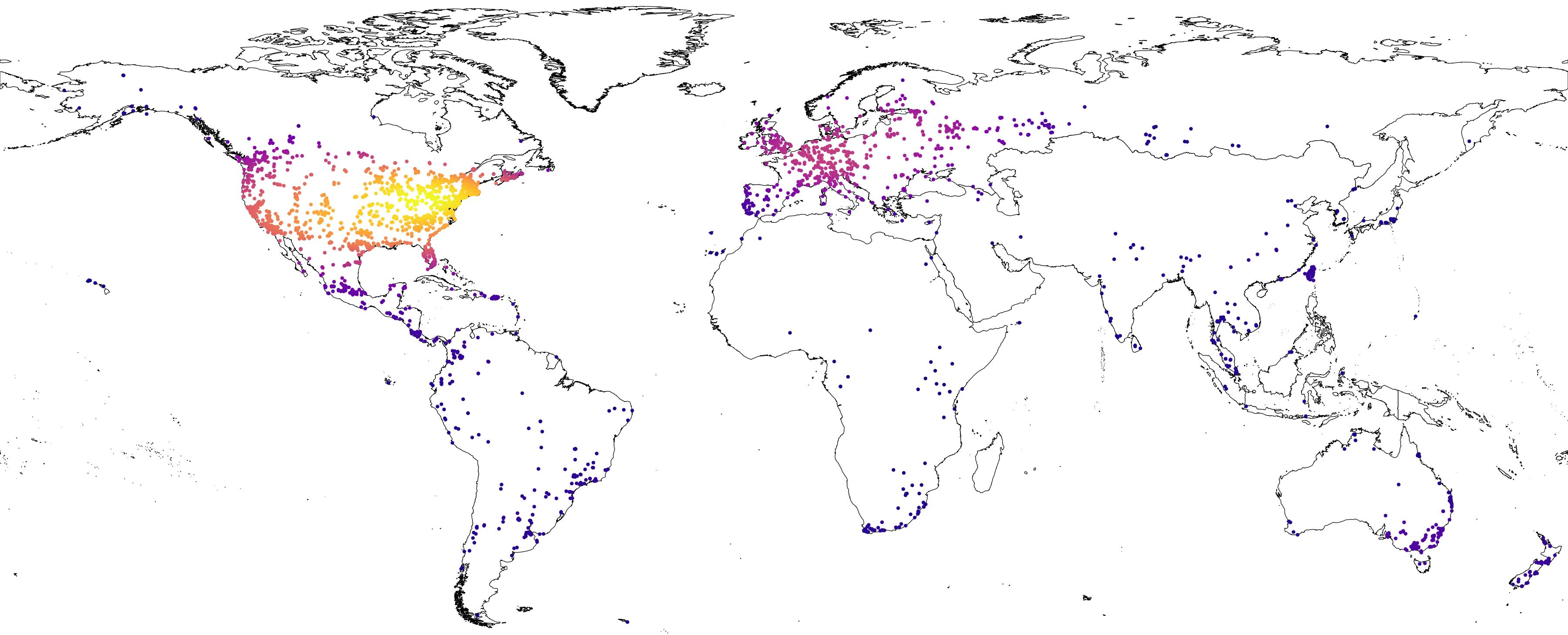}
  \caption*{\small(b) Validation (MultiNat)}
  \end{minipage}
  \begin{minipage}[t]{0.328\linewidth}
  \centering
  \includegraphics[width=\linewidth]
  {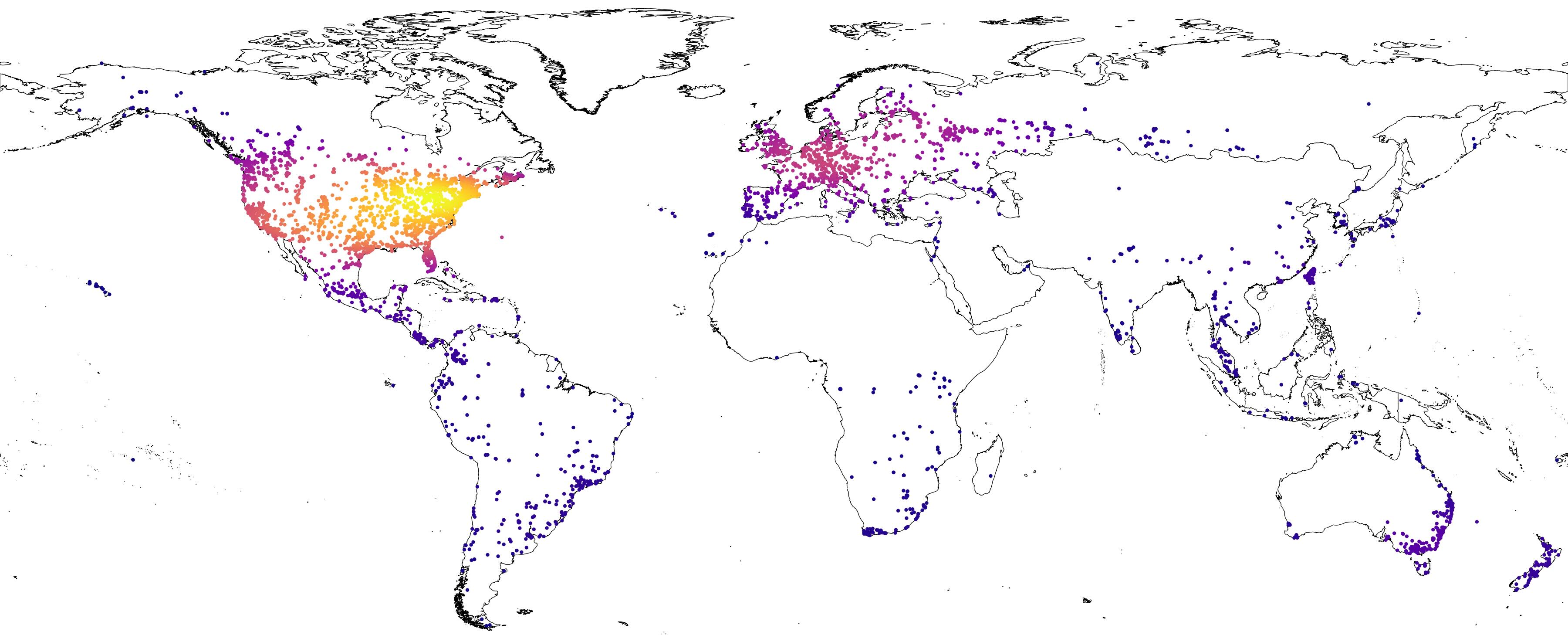}
  \caption*{\small(c) Test (Taxabench-8k)}
  \end{minipage}
\caption{\textbf{Spatial Distribution of Data}. The spatial distribution of our MultiNat dataset covering the globe.}
\label{fig:data_dist}
\end{figure*}

\subsection{Evaluation Datasets}
Below we provide details on the evaluation datasets used in the paper.

\noindent
\textbf{Taxabench-8k.} This dataset consists of 8813 observations from the iNaturalist platform including all modalities paired for each observation. This dataset is used primarily for evaluating the models for the task of cross-modal retrieval.

\noindent
\textbf{BirdClef series.} These datasets, released annually as part of the LifeClef~\cite{joly2024lifeclef} competition, contain geographically confined audio recordings of rare bird species. These datasets are used to identify bird species based on their soundscapes. We use the training, validation and testing split from TaxaBind for this task of bird species audio classification.

\noindent
\textbf{EcoRegions \& Biome.} We follow Range~\cite{dhakal2025range} and use their curated dataset for ecoregion and biome classification of given geographic locations. The dataset was curated by randomly sampling 100k geographic locations across the globe. Each geographic location was assigned a ecoregion label and a biome label. In total, there exist 846 ecoregions and 14 biomes.  
\section{Implementation Details}
Below we provide all the implementation details that were used to train our model.
\label{impl}
\begin{table*}[!h]
\centering    
\begin{tabular}{l|c}
Hyperparameter &  Value \\ \midrule 
batch size & 1024\\
max training epochs & 500\\
optimizer & AdamW\\
optimizer momentum & $\beta_1, \beta_2=0.9, 0.98$\\
learning rate & 1e-4\\
scheduler & cosine decay\\
weight decay & 0.01\\
gpu type & NVIDIA H-100\\
num gpus & 1\\
\end{tabular}
\caption{Configuration used for training.}
\label{table:train_config}

\end{table*}
\begin{table*}[!h]
\centering    
\begin{tabular}{l|c}
Configuration &  Value \\ \midrule 
VIB term weight ($\lambda$) & 0.001\\
base scale parameter ($\alpha$) & -5.0\\
base shift parameter ($\beta$) & 5.0\\
\end{tabular}
\caption{Base configuration for our loss function for MVAE.}
\label{table:loss_config}

\end{table*}
\begin{table*}[!h]
\centering    
\begin{tabular}{l|c}
Configuration &  Value \\ \midrule 
input embedding dim & 512\\
num projection layers & 2\\
encoder dim & 1024\\
feedforward activation function & GeLU\\
num encoder layers & 1\\
decoder dim & 1024\\
num decoder layers & 1\\
\end{tabular}
\caption{Base architecture configuration for MVAE.}
\label{table:arch_config}

\end{table*}
\begin{table*}[!h]
\centering    
\begin{tabular}{l|c}
Modality & Model Architecture \\ \midrule 
ground-level image & OpenCLIP~\cite{ilharco_gabriel_2021_5143773} (ViT-B/16)\\
satellite image & CLIP~\cite{radford2021learning} (ViT-B/16)\\
geographic location & GeoCLIP~\cite{vivanco2024geoclip}\\
environment covariates & SINR~\cite{cole2023spatial}\\
text & OpenCLIP~\cite{ilharco_gabriel_2021_5143773}\\
audio & CLAP~\cite{elizalde2023clap}\\
\end{tabular}
\caption{Architecture configuration for modality-specific encoders.}
\label{table:taxabind_config}
\end{table*}

\section{Additional Experiments \& Ablations}
\label{add_expr}
\subsection{Linear Probing}
\label{linpr}
\subsubsection{Embedding Generation Ablation}
In this experiment, we evaluate several design choices for curating embedding effective for linear probing tasks. The first choice is to utilized all the reconstructed embeddings. This is done by concatenating all embeddings at the output of our MVAE decoder. The rest of the choices involve using the hidden representations of our MVAE encoder. We could use the \textit{{[$\mu$]}}, \textit{[$m_i$]} or register tokens. Additionally, we can concatenate all the tokens from the hidden representations of our MVAE encoder. Table~\ref{tab:lin_prob_abla} presents results of all these choices. We perform linear probing on the BirdClef-2023, EcoRegions and Biome datasets. Except the experiment with \textit{{[$\mu$]}} token, all the choices outperform TaxaBind. Using all the hidden tokens results in the best performance. We find that register tokens are essential and result in a significant gain in performance.

\subsubsection{Probing Geo-Location Embeddings}
Our models can serve as general purpose ecological predictors over space. Generating insights about habitat and climatic conditions of various geographic locations around the world is crucial in understanding global ecological trends. In this experiment, we compare the performance of several pretrained geographic location encoders in predicting various ecological indicators over space. In Table~\ref{table:loc_exp}, we show the performance of our location encoder in predicting Biome, EcoRegion, Temperature and Elevation at a given geographic location. Climplicit is considered the absolute SOTA since it is specifically training on rich spatio-temporal climate data. We find that our model has the best performance beating TaxaBind, SINR and SatCLIP on all the tasks. We also conduct linear probing for the task of predicting several climatic variables in the ERA5 dataset. The results are presented in Table~\ref{table:era5}. We find that our model beats TaxaBind by a large margin and achives the second best performance on average after SINR. We believe that high-frequency geographic location features may not be necessary for these tasks. Climatic variables are typically low-frequency and often do not vary significantly across large regions. SINR is a simple feedforward-based model that outputs low-frequency geo-location embeddings. As a result, it achieves superior performance over other location encoding frameworks.
\begin{table*}[!ht]
  \centering
  \resizebox{\linewidth}{!}{
  \begin{tabular}{lccc|c|cccc}
    \toprule
    Task &Dataset& Modality & TaxaBind~\cite{sastry2025taxabind}&Recons.&\textit{{[$\mu$]}}&\textit{[$m_i$]}&Reg. Tokens&All\\
    \midrule
    Loc. Classification&EcoRegions~&\locChar&73.75&75.96&74.06&74.38&79.44&81.35\\
    Loc. Classification&Biome&\locChar&71.73&76.45&71.19&73.80&78.81&82.30\\
    \midrule
    Audio Classification&BirdCLEF-2023~&\audioChar&42.19&41.17&43.20&45.30&51.65&52.30\\
    Audio Classification&BirdCLEF-2023&\audioChar\plusChar\locChar&46.97&49.25&42.55&54.09&58.10&59.06\\
    \midrule
    \midrule
    Average&&&58.66&60.71&57.75&61.89&67.00&\cellcolor[gray]{0.8}\textbf{68.73}\\
    \bottomrule
  \end{tabular}
  }
  \caption{\textbf{Embedding Generation Ablation}. We investigate different choices for generating embeddings for linear probing. We find using all hidden tokens to achieve the best performance on average.}
  \label{tab:lin_prob_abla}
\end{table*}
\begin{table*}[!ht]
\centering    
\resizebox{\linewidth}{!}{
\begin{tabular}{l|c|cc|ccc}
\hline
                       & Modality & Biome & EcoRegions & Temperature & Elevation   \\ \midrule 
                       Direct   &\locChar                   & 29.1 & 0.6 & 0.381  & 0.025    \\
Cartesian\_3D     &\locChar                 & 30.2 & 1.8 & 0.362  & 0.030     \\
Wrap~\cite{Aodha_2019_ICCV}&\locChar                      & 34.4 & 1.1 & 0.861  & 0.085  \\
Theory~\cite{gaolearning} &\locChar                     & 33.5 & 1.0 & 0.849  & 0.093       \\
SphereM~\cite{mai2023sphere2vec} &\locChar                     & 36.4 & 27.3 & 0.629  & 0.139   \\
SphereM$^{+}$~\cite{mai2023sphere2vec}   &\locChar                   & 58.7 & 50.1 & 0.886  & 0.294   \\
SphereC~\cite{mai2023sphere2vec}   &\locChar                  & 36.3 & 52.9 & 0.461  & 0.185     \\
SphereC$^+$~\cite{mai2023sphere2vec} &\locChar & 53.2 & 61.6 & 0.842  & 0.260       \\

\midrule

CSP-INat~\cite{mai2023csp}   &\locChar                   & 61.1 & 57.1 & 0.717  & 0.388  \\
CSP-FMoW~\cite{mai2023csp}  &\locChar                  & 61.4 & 58.0 & 0.865 & 0.399  \\
SINR~\cite{cole2023spatial}  &\locChar                  & 67.9 & 54.9 & \textbf{0.942}  & 0.644  \\
GeoCLIP~\cite{vivanco2024geoclip}  &\locChar              & 70.2 & 71.6 & 0.916  &  0.604  \\
SatCLIP~\cite{klemmer2025satclip} &\locChar & 68.9 & 69.3 & 0.825       & 0.666  \\
TaxaBind~\cite{sastry2025taxabind}&\locChar  & 71.7 & 73.7 & 0.915 & 0.601  \\

\midrule

ProM3E \textbf{(ours)} &\locChar  & \textbf{82.3} & \textbf{81.3} & \underline{0.918} & \textbf{0.772}  \\ 
 & & \textcolor{Green}{+10.6}& \textcolor{Green}{+7.6} & \textcolor{Green}{+0.003} &\textcolor{Green}{+0.171}  \\ \midrule

Climplicit$^\dagger$~\cite{dollinger2025climplicit} \textcolor{Gray}{(Absolute SOTA)} & \locChar &83.3 &78.4&0.985&0.898\\

\hline
\end{tabular}
}
\caption{Comparison of various pretrained location encoders on predicting four ecological indicators. $^\dagger$Note that climplicit is pretrained on rich spatio-temporal climate data.}
\label{table:loc_exp}
\end{table*}
\begin{table*}[!ht]
\centering
\resizebox{\linewidth}{!}{
\begin{tabular}{l|cccccccc|c}
Models    & \multicolumn{1}{l}{temp\_mean} & \multicolumn{1}{l}{temp\_min} & \multicolumn{1}{l}{temp\_max} & \multicolumn{1}{l}{dew\_temp} & \multicolumn{1}{l}{precipitation} & \multicolumn{1}{l}{pressure} & \multicolumn{1}{l}{u\_wind} & \multicolumn{1}{l|}{v\_wind} & \multicolumn{1}{l}{Avg} \\ \hline
CSP       & 0.944                               & 0.933                              & 0.940                              & 0.918                         & 0.610                             & 0.427                                 & 0.499                       & 0.550                        & 0.727                       \\
CSP-INat & \underline{0.987}                               & 0.897                              & 0.886                              & 0.857                         & 0.534                             & 0.307                                 & 0.413                       & 0.386                        & 0.658                       \\
SINR      & \textbf{0.982}                         & \textbf{0.975}                        & \textbf{ 0.976}                        & \textbf{0.977}                   & \underline{0.758}                             & \underline{0.706}                                 & \textbf{0.726}                       & \textbf{0.694}                        & \textbf{0.849 }                      \\
GeoCLIP   & 0.960                               & 0.953                              & 0.948                              & 0.954                         & 0.591                             & 0.651                                 & 0.502                       & 0.529                        & 0.761                       \\
SatCLIP   & 0.904                               & 0.900                              & 0.887                              & 0.894                         & 0.497                             & 0.743                                 & 0.488                       & 0.455                        & 0.721                       \\
TaxaBind   & 0.965                               &  0.954                           & 0.955                              & 0.957                     & 0.637                             & 0.662         & 0.525                       & 0.560   &0.777                    \\ \midrule

ProM3E \textbf{(Ours)}    & 0.978                      & \underline{0.971}                     & \underline{0.970}                     & \underline{0.972}                & \textbf{0.730}                    & \textbf{0.758}                        & \underline{0.630}              & \underline{0.638}   & \underline{0.830} \\ 
&\textcolor{Green}{+0.013}&\textcolor{Green}{+0.017}&\textcolor{Green}{+0.015}&\textcolor{Green}{+0.015}&\textcolor{Green}{+0.093}&\textcolor{Green}{+0.096}&\textcolor{Green}{+0.105}&\textcolor{Green}{+0.078}&\textcolor{Green}{+0.058}\\
\midrule
\end{tabular}
}
\caption{We show the linear probe results on real-world climate data from ERA5. Our model consistently beats TaxaBind and achieves the second best performance on average after SINR.}
\label{table:era5}
\end{table*}
\begin{table*}[!ht]
\centering
\begin{tabular}{l|c|ccc|ccc}
\hline
                       Method & Modality & \multicolumn{3}{c|}{Biome} & \multicolumn{3}{c}{EcoRegions}    \\ 
                    &&Top-1&Top-5&Top-10&Top-1&Top-5&Top-10 \\ \midrule
TaxaBind~\cite{sastry2025taxabind}&\imageChar  &45.07&83.00&93.03&8.05& 29.16 & 43.10  \\

ProM3E \textbf{(ours)} &\imageChar  &\textbf{57.75}&\textbf{89.51}&\textbf{93.70}& \textbf{25.37} & \textbf{54.06} & \textbf{61.72}\\ 
\midrule
\midrule
TaxaBind~\cite{sastry2025taxabind}&\satChar  & 65.37 & 90.75 &93.56&35.33&65.43&\textbf{74.93}\\

ProM3E \textbf{(ours)} &\satChar  & \textbf{76.57} & \textbf{93.64}&\textbf{93.96}&\textbf{54.79}&\textbf{69.63}&70.18\\ 
\midrule
\midrule
TaxaBind~\cite{sastry2025taxabind}&\envChar  & 60.10 & 90.49 &93.36&26.19&55.13&60.90\\

ProM3E \textbf{(ours)} &\envChar  & \textbf{83.05} & \textbf{93.90}&\textbf{94.00}&\textbf{64.95}&\textbf{73.76}&\textbf{73.79}\\
\hline
\end{tabular}
\caption{\textbf{Habitat Classification.} We perform Biome and EcoRegion classification given species images as input. This is a challenging task and requires robust alignment of species images with geographic location and satellite images. We also test other inputs such as satellite images and environmental covariates.}
\label{table:main}
\end{table*}
\subsubsection{Habitat Classification}
In this experiment, our aim is to classify the habitat of species represented using a given ground-level image. To achieve this, we use the iNat-2021 dataset that includes over 2.7M images of species with corresponding geographic location information. For each sample, we extract the Biome and EcoRegion label. We then obtain the image embeddings for each sample using our model and train a single layer linear classification model to predict the Biome/EcoRegion label given the image embedding. We note that this is a single positive multi label (SPML) problem. For training, we use the assume negative loss which is a common loss used in SPML problems. We evaluate the trained model on the testing split of iNat-2021 dataset.

\subsection{Cross-Modal Retrieval}
\subsubsection{Embedding Generation Ablation}
In this section, we investigate an optimal procedure to generate embeddings for effective cross-modal retrieval. There are several design choices one could use. We compare these design choices in Table~\ref{tab:abla_cm_embedding}. We find that using the representations from the hidden \textit{[$m_i$]} leads to poor performance. We suspect that the representations useful for reconstruction are not necessarily useful for retrieval. We compare the reconstructed embeddings alone for retrieval and find that its performance is better than simply using the TaxaBind representations. We get the best performance using our proposed hybrid approach. 
\begin{table*}[!ht]
  \centering
  \begin{tabular}{lccc|cc}
    \toprule
    Task &Dataset& Modality & TaxaBind~\cite{sastry2025taxabind}&Recons.&Hybrid\\
    \midrule
    Image Classification&TaxaBench-8k~&\imageChar\arrowChar\textChar&34.45&33.23&39.45\\
    Image Classification&TaxaBench-8k~&\imageChar\plusChar\satChar\arrowChar\textChar&37.54&42.34&47.05\\
    \midrule
    Retrieval&TaxaBench-8k~&\satChar\arrowChar\locChar&8.43&17.19&17.87\\
    Retrieval&TaxaBench-8k&\locChar\arrowChar\satChar&9.62&12.64&13.18\\
    \midrule
    \midrule
    Average&&&58.66&60.71&\cellcolor[gray]{0.8}\textbf{68.73}\\
    \bottomrule
  \end{tabular}
  \caption{\textbf{Embedding Generation Ablation}. Here we investigate choices for embeddings useful for cross-modal retrieval.}
  \label{tab:abla_cm_embedding}
\end{table*}

\section{Uncertainty \& Modality Gap}
\label{mod_gap}

\begin{figure*}[!ht]
\centering
  \begin{minipage}[t]{0.19\linewidth}
  \centering
  \includegraphics[width=\linewidth]{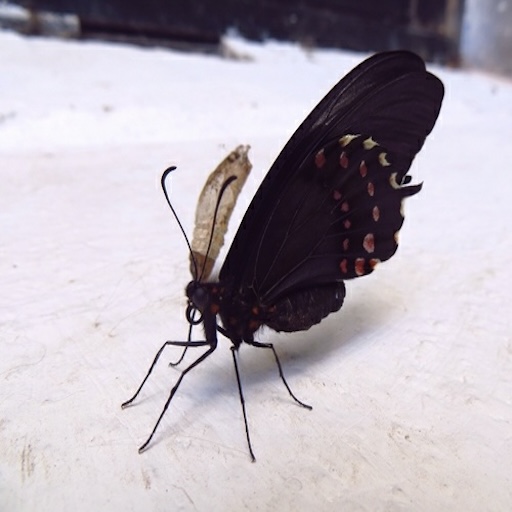}
  \captionsetup{font=footnotesize}
  \caption*{\small$||\sigma||_1=1798.89$}
  \end{minipage}
  \begin{minipage}[t]{0.19\linewidth}
  \centering
  \includegraphics[width=\linewidth]{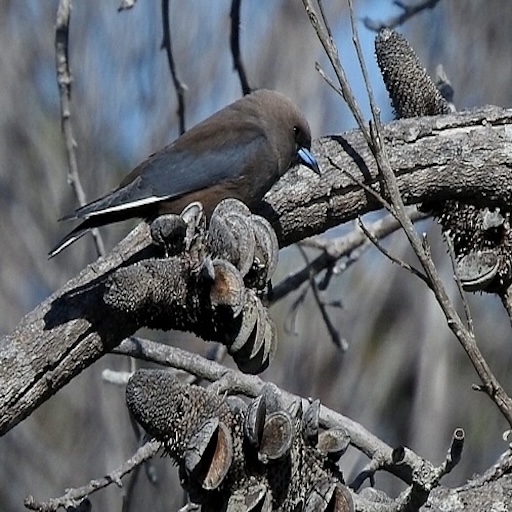}
  \captionsetup{font=footnotesize}
  \caption*{\small$||\sigma||_1=1598.42$}
  \end{minipage}
  \begin{minipage}[t]{0.19\linewidth}
  \centering
  \includegraphics[width=\linewidth]{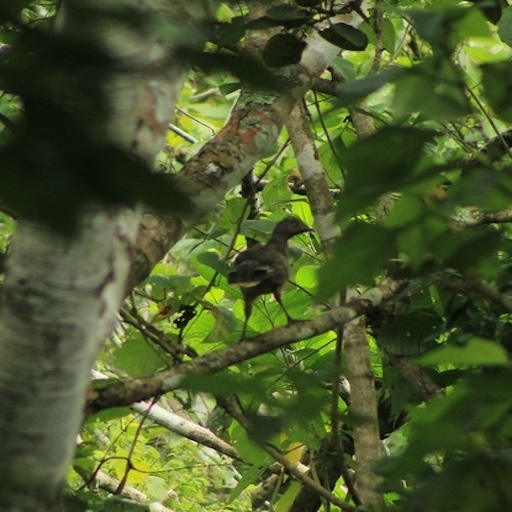}
  \captionsetup{font=footnotesize}
  \caption*{\small$||\sigma||_1=1568.46$}
  \end{minipage}
   \begin{minipage}[t]{0.19\linewidth}
  \centering
  \includegraphics[width=\linewidth]{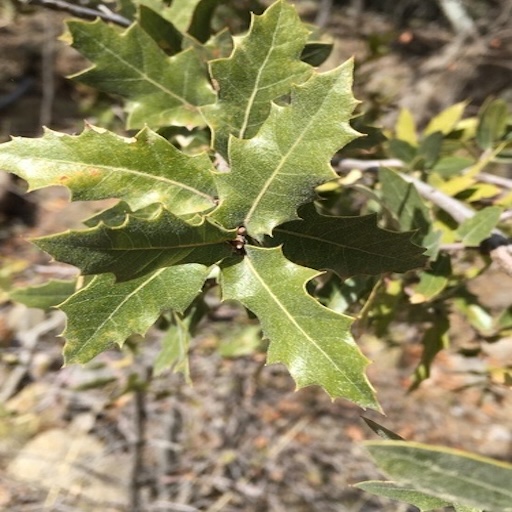}
  \captionsetup{font=footnotesize}
  \caption*{\small$||\sigma||_1=1565.41$}
  \end{minipage}
   \begin{minipage}[t]{0.19\linewidth}
  \centering
  \includegraphics[width=\linewidth]{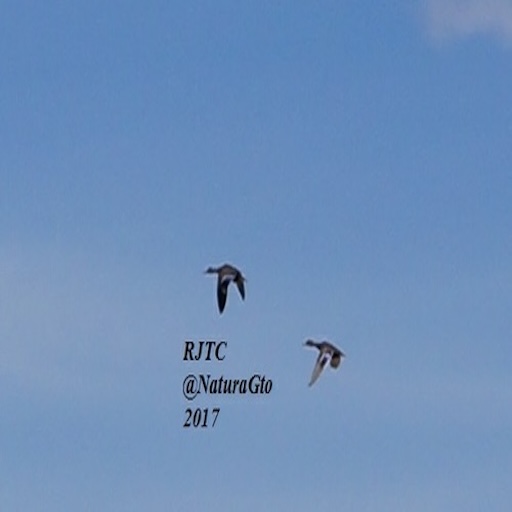}
  \captionsetup{font=footnotesize}
  \caption*{\small$||\sigma||_1=1558.27$}
  \end{minipage}
  \begin{minipage}[t]{0.19\linewidth}
  \centering
  \includegraphics[width=\linewidth]{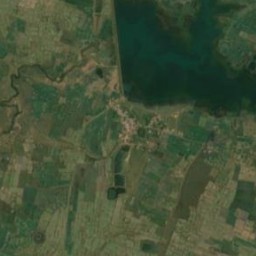}
  \captionsetup{font=footnotesize}
  \caption*{\small$||\sigma||_1=1503.92$}
  \end{minipage}
  \begin{minipage}[t]{0.19\linewidth}
  \centering
  \includegraphics[width=\linewidth]{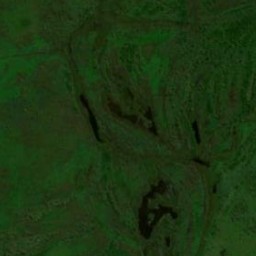}
  \captionsetup{font=footnotesize}
  \caption*{\small$||\sigma||_1=1500.65$}
  \end{minipage}
  \begin{minipage}[t]{0.19\linewidth}
  \centering
  \includegraphics[width=\linewidth]{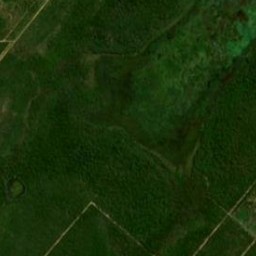}
  \captionsetup{font=footnotesize}
  \caption*{\small$||\sigma||_1=1493.47$}
  \end{minipage}
   \begin{minipage}[t]{0.19\linewidth}
  \centering
  \includegraphics[width=\linewidth]{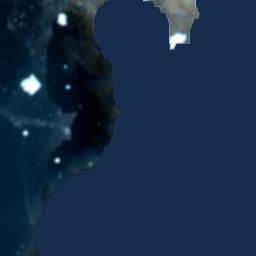}
  \captionsetup{font=footnotesize}
  \caption*{\small$||\sigma||_1=1493.26$}
  \end{minipage}
   \begin{minipage}[t]{0.19\linewidth}
  \centering
  \includegraphics[width=\linewidth]{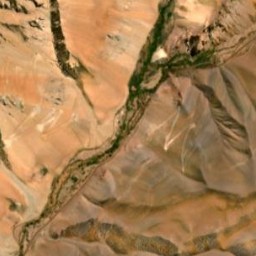}
  \captionsetup{font=footnotesize}
  \caption*{\small$||\sigma||_1=1484.82$}
  \end{minipage}
\caption{\textbf{Most Uncertain Images}. Most uncertain ground-level and satellite images.}
\label{fig:uncertain_imgs}
\end{figure*}

\begin{figure*}[!ht]
\centering
  \begin{minipage}[t]{0.19\linewidth}
  \centering
  \includegraphics[width=\linewidth]{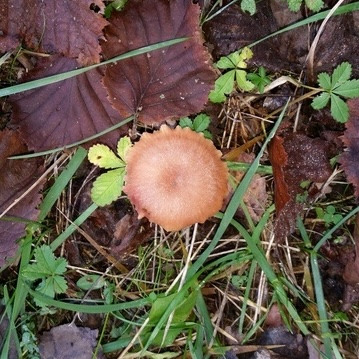}
  \captionsetup{font=footnotesize}
  \caption*{\small$||\sigma||_1=1266.91$}
  \end{minipage}
  \begin{minipage}[t]{0.19\linewidth}
  \centering
  \includegraphics[width=\linewidth]{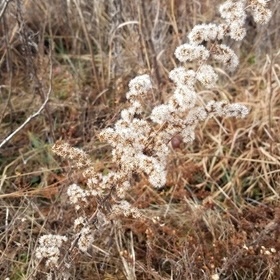}
  \captionsetup{font=footnotesize}
  \caption*{\small$||\sigma||_1=1267.42$}
  \end{minipage}
  \begin{minipage}[t]{0.19\linewidth}
  \centering
  \includegraphics[width=\linewidth]{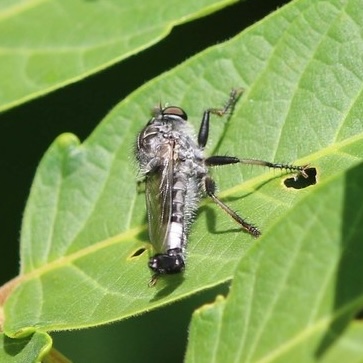}
  \captionsetup{font=footnotesize}
  \caption*{\small$||\sigma||_1=1268.29$}
  \end{minipage}
   \begin{minipage}[t]{0.19\linewidth}
  \centering
  \includegraphics[width=\linewidth]{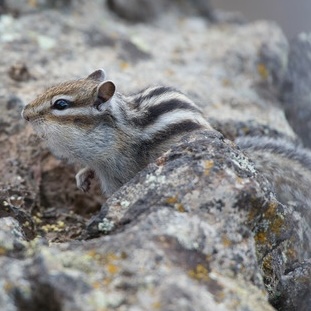}
  \captionsetup{font=footnotesize}
  \caption*{\small$||\sigma||_1=1269.04$}
  \end{minipage}
   \begin{minipage}[t]{0.19\linewidth}
  \centering
  \includegraphics[width=\linewidth]{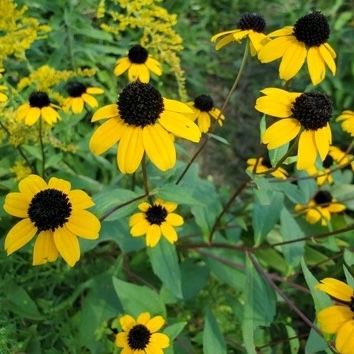}
  \captionsetup{font=footnotesize}
  \caption*{\small$||\sigma||_1=1269.05$}
  \end{minipage}
  \begin{minipage}[t]{0.19\linewidth}
  \centering
  \includegraphics[width=\linewidth]{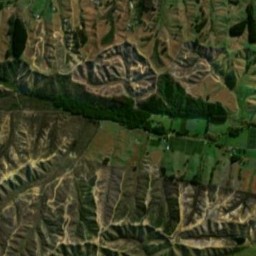}
  \captionsetup{font=footnotesize}
  \caption*{\small$||\sigma||_1=1282.97$}
  \end{minipage}
  \begin{minipage}[t]{0.19\linewidth}
  \centering
  \includegraphics[width=\linewidth]{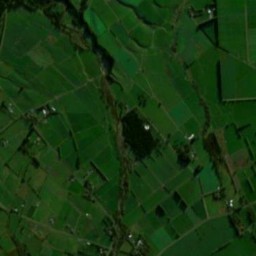}
  \captionsetup{font=footnotesize}
  \caption*{\small$||\sigma||_1=1285.12$}
  \end{minipage}
  \begin{minipage}[t]{0.19\linewidth}
  \centering
  \includegraphics[width=\linewidth]{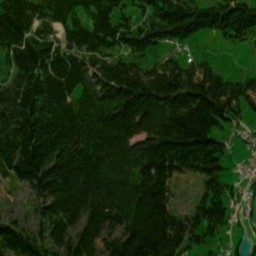}
  \captionsetup{font=footnotesize}
  \caption*{\small$||\sigma||_1=1287.17$}
  \end{minipage}
   \begin{minipage}[t]{0.19\linewidth}
  \centering
  \includegraphics[width=\linewidth]{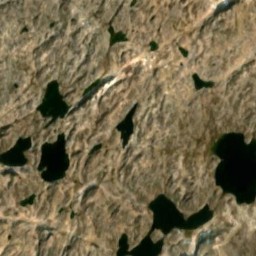}
  \captionsetup{font=footnotesize}
  \caption*{\small$||\sigma||_1=1289.06$}
  \end{minipage}
   \begin{minipage}[t]{0.19\linewidth}
  \centering
  \includegraphics[width=\linewidth]{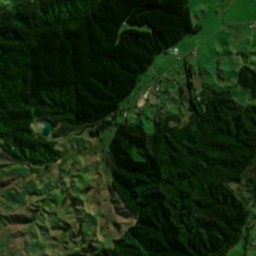}
  \captionsetup{font=footnotesize}
  \caption*{\small$||\sigma||_1=1289.63$}
  \end{minipage}
\caption{\textbf{Least Uncertain Images}. Least uncertain ground-level and satellite images.}
\label{fig:certain_imgs}
\end{figure*}
\begin{figure*}[!ht]
\centering
  \begin{minipage}[t]{0.265\linewidth}
  \centering
  \includegraphics[width=\linewidth]
  {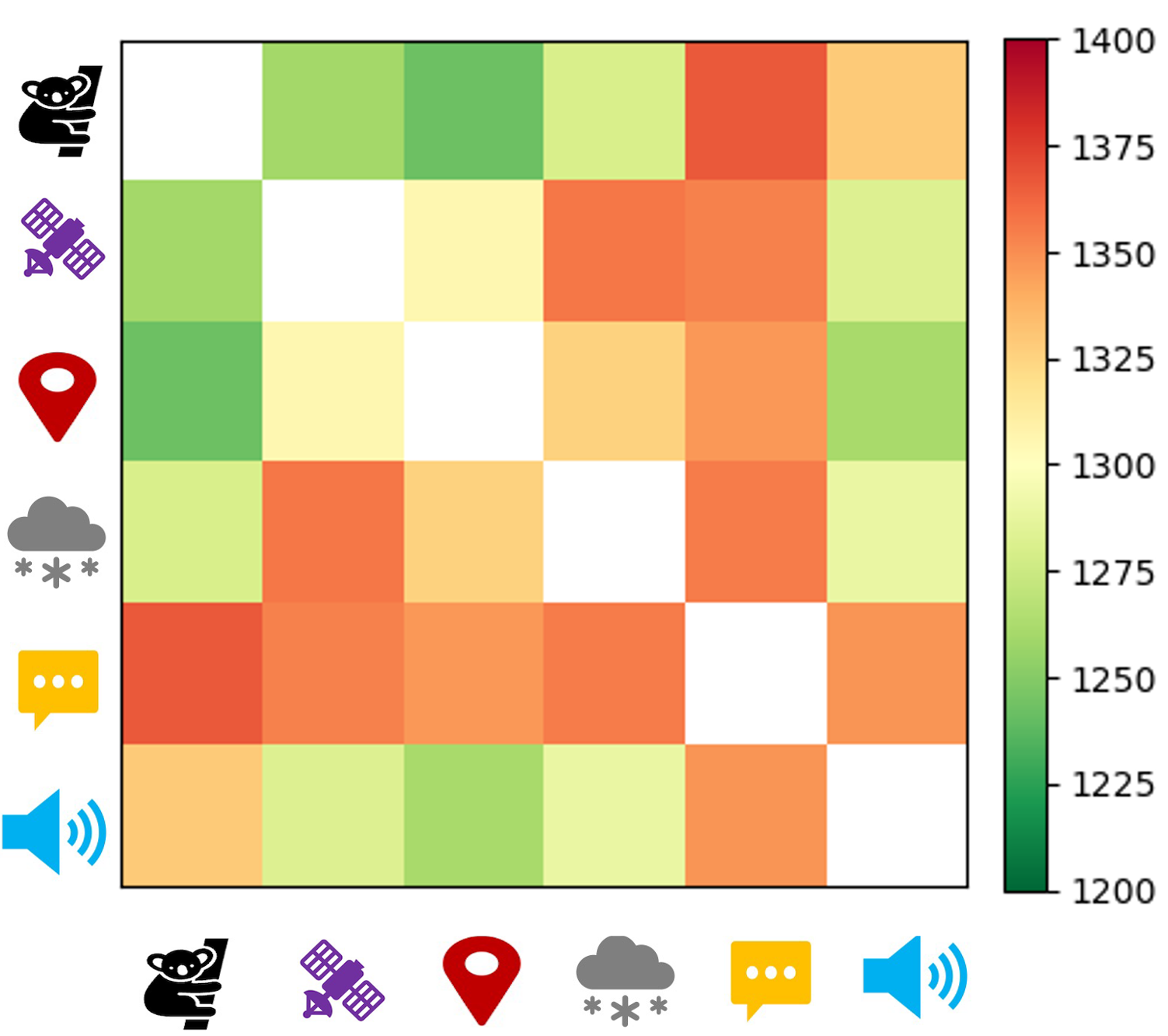}
  \caption*{\tiny(a) $||\sigma||_1$}
  \end{minipage}
  \begin{minipage}[t]{0.26\linewidth}
  \centering
  \includegraphics[width=\linewidth]{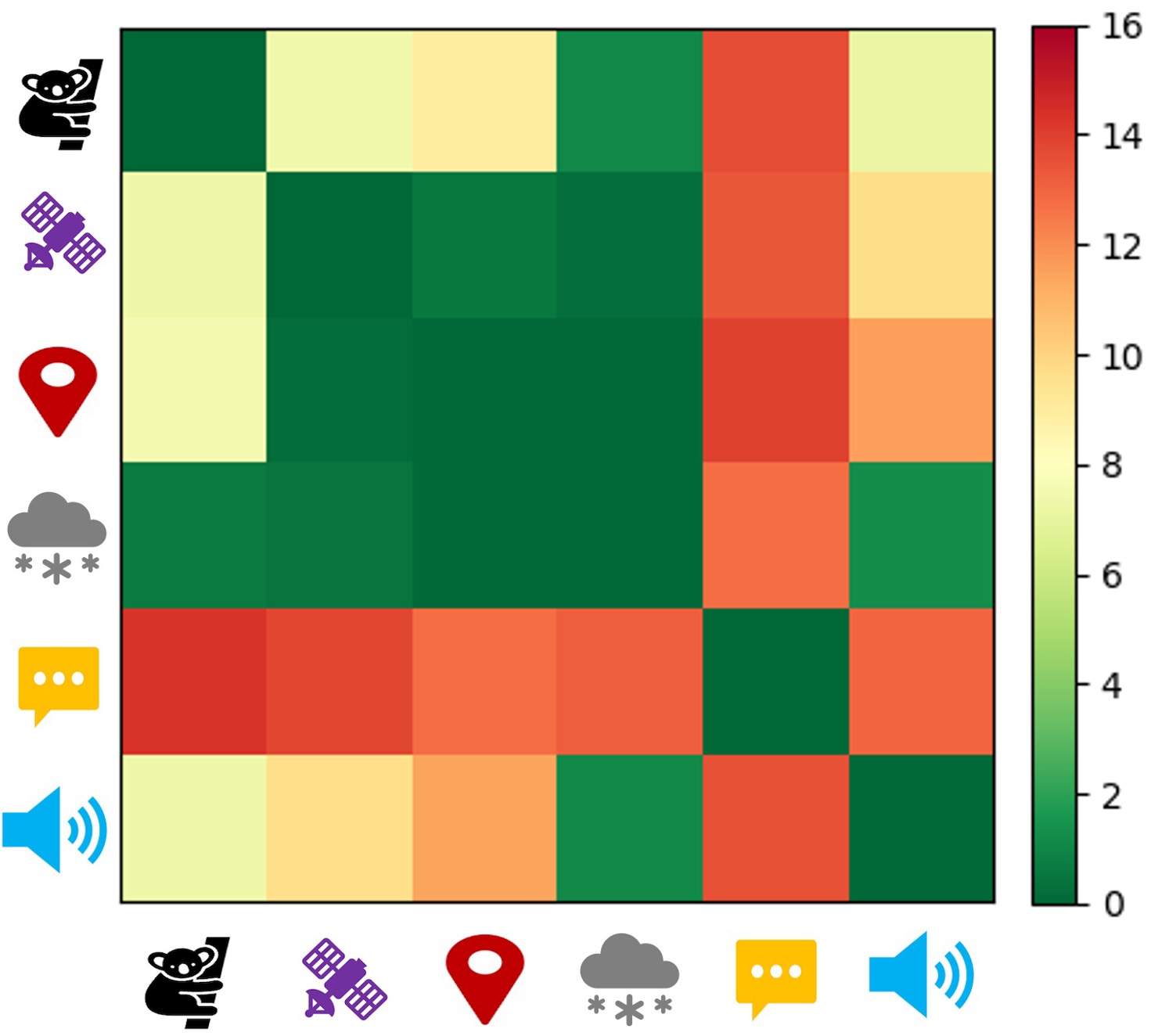}
  \caption*{\tiny(b) modality gap}
  \end{minipage}
\caption{\textbf{Pairwise $||\sigma||_1$ and Modality Gap}. We compare mean $||\sigma||_1$ values and modality gap when pairs of modality are provided as input. We find a spearman correlation of 0.32 between uncertainty and modality gap captured by our model.}
\label{fig:pair_mod_gap}
\end{figure*}

\begin{figure*}[!ht]
\centering
  \begin{minipage}[t]{0.328\linewidth}
  \centering
  \includegraphics[width=\linewidth]
  {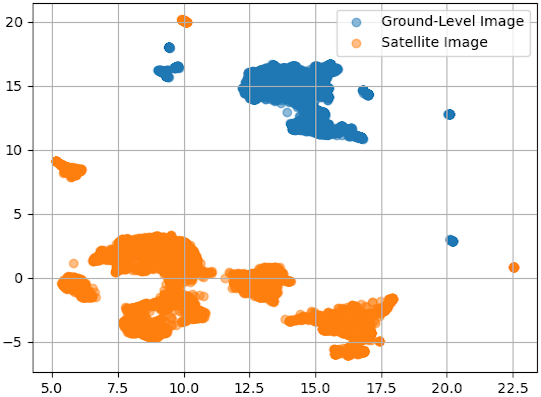}
  \caption*{\tiny(a) Input TaxaBind Representations}
  \end{minipage}
  \begin{minipage}[t]{0.328\linewidth}
  \centering
  \includegraphics[width=\linewidth]{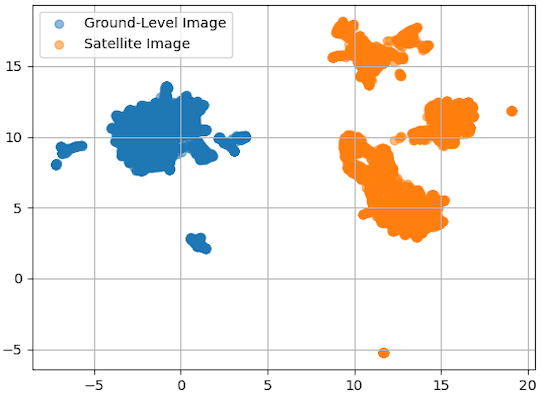}
  \caption*{\tiny(b) Input to ProM3E Encoder}
  \end{minipage}
  \begin{minipage}[t]{0.328\linewidth}
  \centering
  \includegraphics[width=\linewidth]{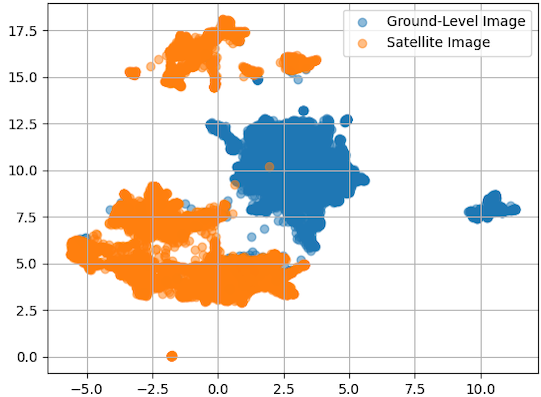}
  \caption*{\tiny(c) Hidden Representations}
  \end{minipage}\\
  \vspace{2mm}
  \begin{minipage}[t]{0.328\linewidth}
  \centering
  \includegraphics[width=\linewidth]{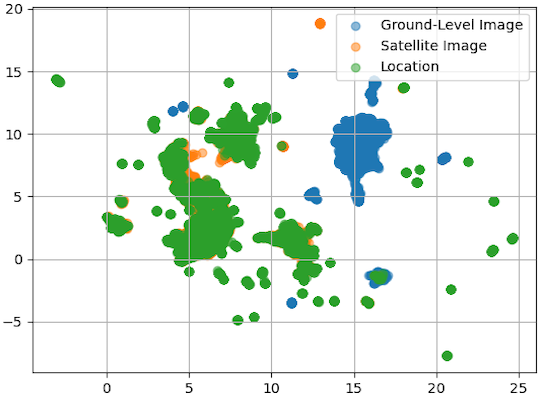}
  \caption*{\tiny(d) Input TaxaBind Representations}
  \end{minipage}
  \begin{minipage}[t]{0.328\linewidth}
  \centering
  \includegraphics[width=\linewidth]{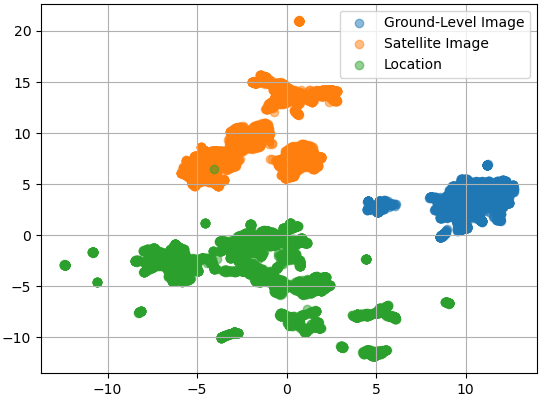}
  \caption*{\tiny(e) Input to ProM3E Encoder}
  \end{minipage}
  \begin{minipage}[t]{0.328\linewidth}
  \centering
  \includegraphics[width=\linewidth]{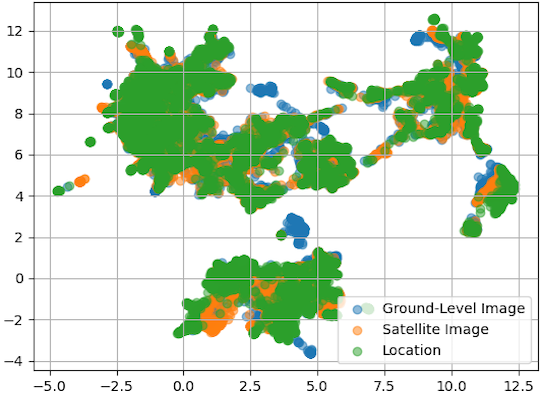}
  \caption*{\tiny(f) Hidden Representations}
  \end{minipage}
\caption{\textbf{Adding Modalities Reduces the Modality Gap}. UMAP visualization of embeddings describing the reduction in modality gap between two modalities in presence of a third modality. Top row presents ground-level and satellite image embeddings while the bottom row presents the embeddings when location is additionally provided as input.}
\label{fig:sdm}
\end{figure*}

\section{Broader Impact}
\subsection{Limitations}
We acknowledge that the datasets used for training and evaluation in this paper suffer from various biases including geographic, socio-economic and human biases. The aim of this paper is to demonstrate the benefits of fusing multiple modalities to improve performance of models on community accepted benchmark datasets. At present our model is limited to accept and process six modalities. However, given the simplicity of our approach, we believe it is trivial to incorporate additional modalities into the framework.

The species diversity and richness maps generated from iNaturalist observations might not accurately represent the Earth’s biodiversity. As noted above, crowdsourcing and citizen science often lead to biased observations, favoring densely populated regions and documenting a limited number of species. Our study aimed to investigate whether the uncertainty captured by our model at different geographic locations correlates with the diversity of species observations in those areas. We found a significant positive correlation between these two factors. This is a promising result which we believe can form basis for future research.

\subsection{Social Impact}
Our models can be effectively adapted to address several remote sensing and ecological challenges. This might mean fine-tuning on additional datasets to adapt our models for specific applications. Our models can serve as a starting point from which interesting applications can emerge. However, utmost care must be taken before deploying them in the real world as is. They might need additional validation before they can be utilized for real world applications. The inherent biases present in the training datasets could potentially lead to inaccurate predictions in certain cases. Consequently, the application of our models in real-world scenarios can benefit from domain expertise. Our model was trained only on openly available species observation data and does not necessarily include information about sensitive species. Yet, care must be taken when using our models for such species. 


\end{document}